\pdfoutput=1

\documentclass{article} %
\usepackage[T1]{fontenc}
\usepackage{preprint_style,times}

\usepackage{amsmath,amsfonts,bm}

\def\eqref#1{equation~\ref{#1}}

\def\1{\bm{1}}

\DeclareMathAlphabet{\mathsfit}{\encodingdefault}{\sfdefault}{m}{sl}
\SetMathAlphabet{\mathsfit}{bold}{\encodingdefault}{\sfdefault}{bx}{n}

\DeclareMathOperator*{\argmax}{arg\,max}

\usepackage{hyperref}
\usepackage{url}
\usepackage{xurl}      %

\usepackage{graphicx}    %
\usepackage{booktabs}    %
\usepackage{multirow}    %
\usepackage{amsmath}     %
\usepackage{amssymb}     %
\usepackage{enumitem}    %
\usepackage{tikz}        %
\usetikzlibrary{arrows.meta,positioning,calc,shapes.geometric}
\usepackage{tcolorbox}
\usepackage{capt-of}
\usepackage{placeins}     %

\title{Textual Planning with Explicit Latent Transitions}

\preprintfinalcopy

\title{Textual Planning\\ with Explicit Latent Transitions}

\makeatletter
\let\embedplan@maketitle\@maketitle
\def\@maketitle{\embedplan@maketitle\lhead{Preprint}}
\makeatother

\hypersetup{
  pdftitle={Textual Planning with Explicit Latent Transitions},
  pdfauthor={Eliezer Shlomi, Ido Levy, Eilam Shapira, Michael Katz, Guy Uziel, Segev Shlomov, Nir Mashkif, Roi Reichart, Sarah Keren},
  pdfsubject={EmbedPlan: a transition model for planning built on frozen text embeddings},
  pdfkeywords={EmbedPlan, LLM planning, transition model, world model, frozen text embeddings, embedding space, classical planning, PDDL, generalization},
}

\author{%
Eliezer Shlomi$^{1*}$\quad Ido Levy$^{2*}$\quad Eilam Shapira$^{1}$\quad Michael Katz$^{2}$\quad Guy Uziel$^{2}$\\
\textbf{Segev Shlomov$^{2}$\quad Nir Mashkif$^{2}$\quad Roi Reichart$^{1}$\quad Sarah Keren$^{1}$}\\
$^{1}$Technion -- Israel Institute of Technology\quad $^{2}$IBM\\
$^{*}$Equal contribution\quad \texttt{Eliezer@campus.technion.ac.il}, \texttt{Ido.Levy1@ibm.com}\\
Code: \href{https://github.com/embedplan/EmbedPlan}{github.com/embedplan/EmbedPlan}%
}

\begin{document}

\maketitle

\begin{abstract}
Planning requires a transition model that predicts how each action changes the current state. When a large language model (LLM) plays this role, every next state is generated token by token, which makes searching over many possible futures slow and expensive. Existing alternatives either still query an LLM at every step or require a symbolic model of the domain. We propose \textsc{EmbedPlan}, a transition model built on frozen text embeddings: it embeds natural language descriptions of the state and the action with a frozen LLM, predicts the embedding of the next state with a lightweight learned network, and returns the closest real state. Because this network can be trained on top of any encoder, \textsc{EmbedPlan} also provides a controlled way to compare text representations for learning transitions. We evaluate it on 9 classical planning domains, under six settings that hold out progressively more of the data, from transitions to entire domains, and against baselines ranging from predicting no change to learning symbolic action rules. On planning problems seen during training, \textsc{EmbedPlan} almost always ranks the true next state among its top five guesses, still does so for most queries even when every observed state is a candidate, and retains 92--99\% of its single-step accuracy when predicting several steps ahead from its own outputs. Given the same candidate states as GPT-5.4, it picks the true next state more often while taking about 0.17~ms per transition with cached embeddings. Accuracy is lower on unseen problems and near chance on unseen domains, and the controlled comparison traces this limit to the state representation rather than to the learned transition.
\end{abstract}

\section{Introduction}
\label{sec:introduction}

When a large language model (LLM) serves as the world model of a planner, each next state is generated token by token, with a full forward pass per token. This makes multi-step lookahead and rollout-based search, which robust long-horizon planning requires, prohibitively expensive in latency and cost. Without a compact transition function that predicts how actions transform states, planners cannot cheaply evaluate alternatives or backtrack from errors.

Prior work avoids this cost only in part. Prompting-based reasoning~\citep{wei2022cot, yao2023tot, yao2023react} still queries the LLM at every step. Compilation to the Planning Domain Definition Language (PDDL)~\citep{liu2023llmp, guan2023leveraging, oswald-et-al-icaps2024, tantakoun-etal-2025-llms, zuo-etal-2025-planetarium} requires a domain that can be expressed as a symbolic model. Latent world models~\citep{ha2018worldmodels, hafner2020dreamer, schrittwieser2020muzero} learn dynamics over perceptual or game observations rather than text. Whether frozen pre-trained language embeddings can host a learned transition function, turning each planning step into a cheap vector operation, has not been systematically tested.

\begin{figure}[t]
\centering
\resizebox{\textwidth}{!}{%
\begin{tikzpicture}[
    font=\footnotesize,
    card/.style={rectangle, rounded corners=3pt, draw=gray!55, fill=gray!6, align=left, inner sep=4pt, text width=2.45cm},
    frozenbox/.style={rectangle, rounded corners=3pt, draw=blue!45, fill=blue!10, line width=0.8pt, align=center, minimum height=1.9cm, minimum width=1.6cm},
    resultbox/.style={rectangle, rounded corners=3pt, draw=green!50!black, fill=green!8, line width=0.8pt, align=left, inner sep=4pt},
    arr/.style={-{Stealth[length=2.2mm]}, line width=0.8pt},
    small/.style={font=\scriptsize}
]
\node[card] (st) at (1.35,0.5) {\textbf{State $s$}\\[-1pt]{\scriptsize arm empty, C clear,\\ C on table, A on B}};
\node[card] (ac) at (1.35,-0.62) {\textbf{Action $a$}\\[-1pt]{\scriptsize\ttfamily pick-up(C)}};
\node[frozenbox] (E) at (4.05,0) {Frozen\\ LLM encoder\\ $E$};
\draw[arr] (st.east) -- (st.east -| E.west);
\draw[arr] (ac.east) -- (ac.east -| E.west);
\node[rectangle, rounded corners=4pt, draw=orange!40, fill=orange!3, minimum width=5.25cm, minimum height=2.2cm] (space) at (8.25,0) {};
\node[small, orange!60!black, anchor=west] at (5.72,-0.93) {latent space};
\foreach \x/\y in {7.25/0.02, 7.6/-0.35, 8.1/-0.6, 9.75/0.62, 10.35/0.45, 10.5/-0.45, 10.0/0.0}{\fill[gray!55] (\x,\y) circle (1.7pt);}
\fill[gray!55] (9.3,-0.93) circle (1.7pt); \node[small, gray!40!black, anchor=west] at (9.37,-0.93) {real states};
\fill[blue!65!black] (5.95,0.2) circle (2.3pt);
\node[small, blue!50!black, anchor=north] at (5.95,0.12) {$\mathbf{h}_s$};
\node[diamond, fill=violet!75!black, inner sep=0pt, minimum size=6.5pt] (ha) at (6.25,-0.33) {};
\node[small, violet!70!black, anchor=west] at (6.35,-0.33) {$\mathbf{h}_a$};
\draw[arr] (st.east -| E.east) -- node[pos=0.35, above, small, orange!70!black, inner sep=1.5pt] {$\pi_s$} (5.87,0.2);
\draw[arr] (ac.east -| E.east) -- node[pos=0.27, below, small, orange!70!black, inner sep=1.5pt] {$\pi_a$} (ha.west);
\node[star, star points=5, star point ratio=2.2, fill=orange!85!black, inner sep=0pt, minimum size=8.5pt] (pred) at (9.0,0.12) {};
\draw[-{Stealth[length=2mm]}, line width=1.2pt, orange!85!black] (6.07,0.23) to[bend left=16] coordinate[pos=0.22] (mid) node[above, small, orange!70!black, pos=0.5] {learned transition $T_\theta(\mathbf{h}_s,\mathbf{h}_a)$} (pred);
\draw[dotted, violet!70!black, line width=0.8pt] (ha.north) -- (mid);
\node[small, orange!70!black, anchor=north east] at (8.80,0.17) {$\hat{\mathbf{h}}_{s'}$};
\fill[green!50!black] (9.25,-0.12) circle (2.5pt);
\draw[dashed, green!45!black, line width=0.6pt] (9.12,0.0) circle (0.38);
\node[small, green!35!black, anchor=north] at (9.12,-0.4) {nearest state};
\node[resultbox, text width=2.2cm] (res) at (12.6,0) {\textbf{Next state $\hat{s}'$}\\[-1pt]{\scriptsize arm holding C,\\ A on B}};
\draw[arr] (space.east) -- (res.west);
\node[small, green!35!black, font=\scriptsize\itshape, anchor=north] at (res.south) {$\approx$\,0.17\,ms per transition};
\draw[-{Stealth[length=2mm]}, dashed, green!45!black, line width=0.7pt] ($(res.south)+(-0.95,-0.33)$) -- ++(0,-0.5) -| (-0.25,0.5) -- (st.west);
\end{tikzpicture}}
\caption{\textsc{EmbedPlan}. A frozen LLM encoder $E$ embeds the state $s$ and the action $a$ (Blocksworld, \texttt{pick-up(C)}), and learned heads $\pi_s,\pi_a$ project them into a latent space where the transition is computed: a small network $T_\theta$ predicts the next-state embedding $\hat{\mathbf{h}}_{s'}$, and the nearest real state is retrieved as the successor $\hat{s}'$. Fed back as the next input, it supports multi-step rollout. Search itself is external. Latency is per transition with state embeddings cached (Appendix~\ref{app:compute}).}
\label{fig:main_architecture}
\end{figure}

We introduce \textsc{EmbedPlan}, which learns explicit transition dynamics over frozen LLM embeddings of text-described planning domains (Figure~\ref{fig:main_architecture}). It encodes a natural language state description (e.g., ``Block A is on B, Block C is clear'') and an action (e.g., \texttt{pick-up(C)}) with a frozen LLM. A small learned head projects both into a low-dimensional space and predicts the next-state embedding there, and the nearest real state is retrieved as the successor. Two contrastive objectives train it: state prediction, which identifies the correct next state among candidates, and action disambiguation, which separates the effects of different actions on the same state. Semantic encoding thus stays with the frozen LLM, and only the cheap dynamics are learned. \textsc{EmbedPlan} is a transition component, not a complete planner: search composes many transitions, and we study the single transition that search queries repeatedly. Because the same head can be trained over any encoder, the framework also serves as a controlled probe of what a frozen text representation supports for learning dynamics.

We evaluate \textsc{EmbedPlan} on 9 classical planning domains from \textsc{ACPBench}~\citep{kokel2025acpbench}, with states rendered as natural language (nearly 3 million transitions), and 4 frozen encoders from MPNet to Llama-3.3-70B. Six protocols form a ladder of exposure (Table~\ref{tab:protocols}). Test transitions come from observed problems in Interpolation (held out at random) and Plan-Variant (along unseen optimal plans), from unseen problems of an observed domain in Extrapolation and Multi-Domain (one model for all nine domains), and from unseen domains in Cross-Domain (trained on one other domain) and Leave-One-Out (trained on the other eight). We report Hit@$k$, the fraction of queries whose true next state ranks in the top $k$ of 128 candidates (chance: $3.9\%$ at $k{=}5$).

Our experiments show that, within observed problems, \textsc{EmbedPlan} predicts transitions accurately and fast enough for search (Section~\ref{sec:within}). Its predictions are near-perfect ($99.7\%$ Hit@5), keep the true next state in the top five for most queries even when every observed state is a candidate, and stay accurate when each prediction becomes the input to the next transition, as when rolling out a plan. Given the same candidates, \textsc{EmbedPlan} also picks the true next state more often than GPT-5.4, at about $0.17$~ms per transition with cached state embeddings. This accuracy depends on exposure, however: Hit@5 falls to $54.6\%$ on unseen problems of an observed domain and to $6.6\%$, near chance, on unseen domains (Section~\ref{sec:boundary}). Varying only the state representation shows that this boundary lies in the representation, not in the learned transition (Section~\ref{sec:why}).

\textbf{Contributions.}
\begin{itemize}[leftmargin=1.2em, topsep=1pt, itemsep=1pt, parsep=0pt]
\item \textsc{EmbedPlan}, a transition model that runs on frozen LLM embeddings: a lightweight head predicts the next-state embedding, and nearest-neighbor retrieval returns a real state that can be fed back for multi-step rollout (Section~\ref{sec:setup}).
\item A controlled evaluation methodology for learned transition models over text: six protocols that hold out progressively more, from transitions to entire domains, across 9 domains and 4 encoders, with reference methods from no-change floors to symbolic action-model induction (Table~\ref{tab:protocols}).
\item A systematic study that maps where such transitions work and where they fail, and identifies the frozen state representation as the main limit on transfer (Section~\ref{sec:results}).
\end{itemize}

\section{Related work}
\label{sec:related_work}
Prompting methods frame planning as iterative generation, using chain-of-thought \citep{wei2022cot}, tree search \citep{yao2023tot}, or environment interaction \citep{yao2023react, shinn2023reflexion}. Other approaches treat LLMs as world models for Monte Carlo tree search (MCTS) \citep{hao2023rap} or as text-based world simulators \citep{wang2024textworldsim}, or compile text into PDDL for classical solvers \citep{liu2023llmp, guan2023leveraging}. However, the prompting and world-model approaches query an LLM at every decision, compilation requires a domain that can be expressed as a symbolic model, and LLM outputs in either role can be hallucinated or unexecutable \citep{kambhampati2024position, katz2024thought}. \textsc{EmbedPlan} addresses this compute bottleneck by efficiently computing next states via latent transitions in frozen embedding spaces.

Early work demonstrated that planners can operate in a learned latent space rather than a hand-crafted symbolic state space. \citet{asai2018latplan} introduced \textsc{LatPlan}, which learns discrete state encodings from visual inputs via autoencoders and applies classical planners in latent space. Model-based reinforcement learning extended this by learning dynamics models that predict latent states under actions \citep{ha2018worldmodels, hafner2019planet, hafner2020dreamer, schrittwieser2020muzero}, including transformer world models \citep{micheli2023iris}, latent tree search \citep{gieselmann2022latent}, and language-conditioned world models \citep{lin2024dynalang}. Closest to our design, C-SWM \citep{kipf2020cswm} learns latent transitions contrastively and evaluates them by ranking, and DINO-WM \citep{zhou2025dinowm} learns dynamics over frozen pretrained visual features. However, these methods typically model visual or game observations with small or low-dimensional action spaces, whereas \textsc{EmbedPlan} builds on a frozen text encoder and must generalize over action \emph{text} in combinatorial grounded action spaces (487 grounded actions on Ferry, 311 on Logistics).  A parallel line of work keeps the state in text: \citet{li2026bridgingagentworldgaptext} survey \emph{text world models} (transition functions over textual states) and organize them by state representation into natural language, structured formats, and executable code, a taxonomy in which a learned latent vector is absent. Belief-graph agents for text games \citep{adhikari2020gata} likewise keep an explicit symbolic state. \citet{huang2026stateconsistencybehaviorconsistency} further argue that single-step state-matching metrics understate how a text world model behaves over a trajectory, motivating our Plan-Variant and plan-level evaluations alongside single-step Hit@$k$. Structured approaches offer complementary inductive biases for cross-domain transfer, each presupposing a symbolic decomposition of the state: object-centric architectures that factor states into entities and relations~\citep{battaglia2018relational}, neural generalized planners over lifted PDDL structure~\citep{toyer2018asnets, shen2020stripshgn, stahlberg2022gnn}, action-model learners that induce symbolic transition models from traces~\citep{yang2007arms, cresswell2013locm, aineto2019fama, juba2021sam}, of which our lifted STRIPS reference is an instance, and symbolic compilation to PDDL~\citep{liu2023llmp, guan2023leveraging}. Our study quantifies the gap that these structured approaches must close and provides a diagnostic methodology to measure it.

Sentence embedding methods target semantic similarity, often using contrastive objectives \citep{reimers2019sbert, gao2021simcse, radford2021clip, oved2025snap}. Representation learning for control uses similar contrastive signals to induce state structure \citep{oord2018cpc, srinivas2020curl, schwarzer2021spr}, while JEPA-style methods predict held-out representations in an embedding space rather than reconstructing raw inputs \citep{assran2023self}. Retrieval-augmented models similarly rely on embedding space lookups for prediction \citep{khandelwalgeneralization} and control \citep{humphreys2022large}. However, these approaches largely focus on visual/RL settings or general-purpose similarity, and are not evaluated as action-conditioned transition operators for planning in frozen language embedding spaces. \textsc{EmbedPlan} bridges this gap by learning transition functions in a frozen LLM embedding space and systematically evaluating generalization across planning domains and problem configurations.

Classical planning benchmarks provide controlled testbeds \citep{mcdermott1998pddl}. \textsc{ACPBench} measures predictive transition accuracy and generalization across held-out problems \citep{kokel2025acpbench}. Compositional generalization benchmarks reveal that neural models often memorize training patterns \citep{lake2018scan, kim2020cogs}. We evaluate transition learning in embedding space with explicit within-distribution and out-of-distribution metrics, connecting failures to embeddings that cluster by surface form rather than by generalizable structure.

\section{Method and experimental setup}
\label{sec:setup}

\paragraph{Task formulation}
Given a state $s$ and an action $a$ described in natural language, we predict the next state $s'$ with one learned vector step followed by retrieval, so that every prediction is a real state. Let $E$ be a frozen LLM encoder mapping text to $\mathbb{R}^d$, and let $\pi_s, \pi_a: \mathbb{R}^d \to \mathbb{R}^{d'}$ be learned state and action projection heads with $d' = 128$. With $\mathbf{h}_s = \pi_s(E(s))$ and $\mathbf{h}_a = \pi_a(E(a))$, a lightweight transition network $T_\theta: \mathbb{R}^{d'} \times \mathbb{R}^{d'} \to \mathbb{R}^{d'}$ predicts the next-state embedding
\begin{equation}
    \hat{\mathbf{h}}_{s'} = T_\theta\big(\mathbf{h}_s,\, \mathbf{h}_a\big).
\end{equation}
At inference, the next state is the nearest neighbor in a candidate pool $\mathcal{C}$ of encoded states:
\begin{equation}
    \hat{s}' = \argmax_{c \in \mathcal{C}} \; \mathrm{sim}\big(\hat{\mathbf{h}}_{s'},\, \pi_s(E(c))\big),
\end{equation}
where $\mathrm{sim}(\cdot, \cdot)$ denotes cosine similarity. Only $\pi_s$, $\pi_a$, and $T_\theta$ (together, the head) are trained.

\begin{table}[t]
\caption{Evaluation protocols, grouped by exposure: what the test data share with training. Every query is ranked against 128 states, the true successor and 127 distractors (definitions in Appendix~\ref{app:protocols}).}
\label{tab:protocols}
\centering\small\setlength{\tabcolsep}{5pt}
\begin{tabular}{@{}llll@{}}
\toprule
\textbf{Exposure} & \textbf{Protocol} & \textbf{Trained on $\rightarrow$ tested on} & \textbf{Distractors from} \\
\midrule
Observed problems & Interpolation & $80\%$ $\rightarrow$ $20\%$ of transitions & the whole domain \\
                  & Plan-Variant  & some $\rightarrow$ other optimal plans & alternative-plan successors \\
Unseen problems   & Extrapolation & ${\sim}80\%$ $\rightarrow$ $20\%$ of problems & the query's problem \\
                  & Multi-Domain  & the same, all nine domains at once & the query's problem \\
Unseen domains    & Cross-Domain  & one domain $\rightarrow$ another & the target domain \\
                  & Leave-One-Out & eight domains $\rightarrow$ the ninth & the target domain \\
\bottomrule
\end{tabular}
\end{table}

\paragraph{Architecture}
The frozen encoder $E$ is MPNet (all-mpnet-base-v2), BGE-M3, Qwen2.5-7B, or Llama-3.3-70B (768 to 8{,}192 dimensions). For the two decoder-only models we mean-pool the final hidden states (Appendix~\ref{app:encoders}). Results use Llama-3.3-70B unless another encoder is named. Each state is encoded as a prompt that contains the problem and goal description followed by a \texttt{CURRENT STATE} block listing the facts (PDDL literals) that hold (Appendix~\ref{app:examples}). For $T_\theta$ we evaluate a residual MLP that concatenates its inputs and adds a skip connection, and a hypernetwork that generates action-specific parameters, both with fewer than 500K parameters. The two perform on par (Appendix~\ref{app:arch_encoder_compare}), so we report the MLP throughout (details in Appendix~\ref{app:architecture}).

\paragraph{Training objective}
We optimize $\mathcal{L} = \mathcal{L}_{\text{state}} + \lambda \mathcal{L}_{\text{action}}$ with $\lambda = 2$. The state loss is InfoNCE~\citep{oord2018cpc} between predicted and true next-state embeddings with in-batch negatives. The action loss requires the true action to yield a prediction closer to $s'$ than up to $K{=}50$ alternative actions applied to the same state (Appendix~\ref{app:training}). Hyperparameters and checkpoints are selected on a held-out development split, and results average 3 seeds.

\paragraph{Intended use}
Action selection and search remain external to \textsc{EmbedPlan}. It targets settings where states and actions are available as text but no symbolic model is, so that the alternative is querying an LLM at every expansion. Our classical domains have exact simulators, which makes them a controlled testbed: every prediction can be scored against ground truth. Retrieval needs a set of candidate successors that contains the true one, so we measure how accuracy degrades as that set grows to every state observed in a domain, and how the transition behaves when iterated on its own predictions. Training is per domain: a one-time embedding pass plus ${\sim}15$ minutes of training, which breaks even against autoregressive generation after ${\sim}4{,}300$ transitions with Llama-3.3-70B (Appendix~\ref{app:compute}), so it pays off for a domain that is planned in repeatedly.

\paragraph{Data}
We use nine classical PDDL domains from \textsc{ACPBench}~\citep{kokel2025acpbench} (Appendix~\ref{app:domains}): Blocksworld (a robot arm stacks blocks), Depot (hoists and trucks move crates), Ferry (a ferry carries one car at a time), Floortile (robots paint a tiled floor), Goldminer (a robot clears rock with a laser or bombs to reach gold), Grid (a robot carries keys through a grid of locked cells), Logistics (trucks and airplanes deliver packages within and across cities), Rovers (rovers sample, image, and transmit data), and Satellite (satellites calibrate instruments and take images). A \emph{domain} fixes the action schemas, and a \emph{problem} fixes the objects, the initial state, and the goal. Each action is a \emph{grounded} instance (e.g., \texttt{pick-up(C)}) of a \emph{lifted} action schema (\texttt{pick-up(?x)}). For each problem we extract state-action-next-state triplets $(s_t, a_t, s_{t+1})$, 2.97M transitions over 67 problems (259K unique states), from 13{,}256 (Depot) to 1{,}248{,}696 (Satellite) per domain, together with optimal plans for the plan-level protocols (Appendix~\ref{app:dataset}).

\paragraph{Metrics}
Hit@$k$ is the fraction of queries whose true next state ranks in the top $k$ of a candidate pool. Unless stated otherwise the pool holds 128 states, the true successor and 127 distractors drawn as in Table~\ref{tab:protocols}, so chance is $3.9\%$ for Hit@5 and $0.8\%$ for Hit@1. We emphasize Hit@5 because it is robust to the pool: when the pool grows to every observed state (Interpolation), Hit@1 falls by about $3\times$ while Hit@5 falls by at most $1.33\times$ (Appendix~\ref{app:pool_size}), so 128-way Hit@1 largely reflects the pool size. A top-5 list is also useful where a verifier can filter candidates. Where we claim usability without one, we report Hit@1. Hit@1 and Hit@10 appear in most appendix tables.

\paragraph{Reference methods}
To locate what limits transfer, we keep the head and pool and vary only the state representation (Section~\ref{sec:why}). Two non-learned floors need no training: Identity predicts the current state's embedding, and Offset adds the mean training displacement of the ground action or of its schema. The learned references feed the same head sparse lexical features of the same text (TF-IDF bigrams, bag-of-words, character 3--5 grams), a tabular one-hot code with one lookup row per state, or a bag of the state's symbolic facts (bag-of-literals). Lifted STRIPS induction instead infers each action schema's add and delete effects from those facts and serves as an oracle upper bound. Two null controls bound the protocol from below: the prompt without its \texttt{CURRENT STATE} block (context-only) and a fixed random Gaussian vector per state (Appendix~\ref{app:baselines}).

\section{Results}
\label{sec:results}

A learned transition over frozen text embeddings is near-perfect within observed problems, partial on unseen problems of an observed domain, and near chance on unseen domains. Section~\ref{sec:within} shows that within observed problems the component has the properties search needs, Section~\ref{sec:boundary} maps where generalization stops, and Section~\ref{sec:why} traces the limit to the frozen state representation.

\subsection{Within observed problems}
\label{sec:within}

\begin{table}[t]
\begin{minipage}[t]{0.40\textwidth}\centering
\caption{Top-5 retrieval degrades gracefully as the pool grows to every observed state, while Hit@1 falls about $3\times$. Interpolation, Llama-3.3-70B, \% (mean over 3 seeds, SE in Table~\ref{tab:app_pool_size}). ``Full'': every observed state of the domain.}
\label{tab:pool_main}
\small\setlength{\tabcolsep}{3pt}
\begin{tabular}{@{}rcccc@{}}
\toprule
& \multicolumn{2}{c}{\textbf{Ferry}} & \multicolumn{2}{c}{\textbf{Logistics}} \\
\cmidrule(lr){2-3}\cmidrule(lr){4-5}
\textbf{Pool} & \textbf{Hit@1} & \textbf{Hit@5} & \textbf{Hit@1} & \textbf{Hit@5} \\
\midrule
128         & 99.0 & 100.0 & 96.3 & 99.9 \\
512         & 97.2 & 99.9  & 88.2 & 99.4 \\
2{,}048     & 90.9 & 99.4  & 70.1 & 96.7 \\
8{,}192     & 76.6 & 98.0  & 47.2 & 83.1 \\
Full        & 35.2 & 86.9  & 32.6 & 74.9 \\
\midrule
\textit{Full size} & \multicolumn{2}{c}{46{,}205} & \multicolumn{2}{c}{13{,}373} \\
\bottomrule
\end{tabular}
\end{minipage}\hfill
\begin{minipage}[t]{0.57\textwidth}\centering
\caption{Trained on other transitions of the same problems, \textsc{EmbedPlan} matches or beats every LLM tested at a fraction of the compute. \emph{Generate}: Hit@5 of the true successor among the LLM's generated candidates (\textsc{EmbedPlan}'s retrieval Hit@5: first row of Table~\ref{tab:pool_main}). \emph{Rank}: top-1 accuracy over the same 128 candidates \textsc{EmbedPlan} ranks. Tokens and GFLOPs are per query, for \textsc{EmbedPlan} with cached state embeddings (Appendix~\ref{app:compute}). n/a: not available.}
\label{tab:llm_baselines}
\small\setlength{\tabcolsep}{3.5pt}
\begin{tabular}{@{}lcccc@{}}
\toprule
\textbf{Method} & \textbf{Ferry} & \textbf{Logistics} & \textbf{Tokens} & \textbf{GFLOPs} \\
\midrule
\multicolumn{5}{@{}l}{\textit{Generate $s'$, Hit@5 (\%)}} \\
Llama-3.1-8B  & 2  & 95  & 3{,}840 & $\sim$4.8k \\
Qwen3-30B     & 41 & 86  & 874  & $\sim$18.7k \\
Llama-3.1-70B & 45 & 100 & 781  & $\sim$35k \\
GPT-oss-120B  & 56 & 100 & 1{,}776 & $\sim$281k \\
GPT-5.4       & 56 & 100 & 829  & n/a \\
\midrule
\multicolumn{5}{@{}l}{\textit{Rank 128 candidates, top-1 (\%)}} \\
GPT-5.4           & 44 & 92 & n/a & n/a \\
\textsc{EmbedPlan} & \textbf{99.0} & \textbf{96.3} & n/a & \textbf{0.3} \\
\bottomrule
\end{tabular}
\end{minipage}
\end{table}

\begin{figure}[t]
\centering
\begin{minipage}[b]{0.48\textwidth}\centering
\raisebox{0.49cm}{\includegraphics[height=4.89cm,trim=19 20 19 19,clip]{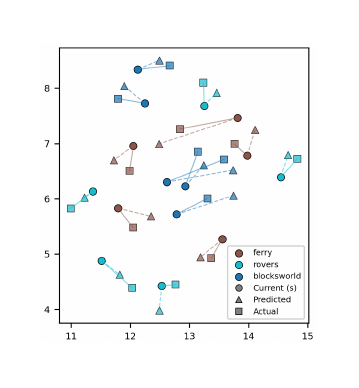}}
\end{minipage}\hfill
\begin{minipage}[b]{0.48\textwidth}\centering
\includegraphics[height=5.3cm,trim=5 5 6 5,clip]{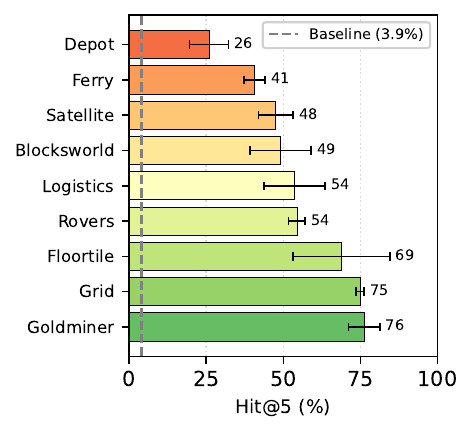}
\end{minipage}\\[2pt]
\begin{minipage}[t]{0.48\textwidth}
\caption{Predictions land next to their targets (illustrative). PCA of the latent space for sampled transitions from three domains (Interpolation, BGE-M3): states $s$ ($\circ$), predicted embeddings $\hat{\mathbf{h}}_{s'}$ ($\triangle$), true next states $s'$ ($\square$). Each state is linked to its true successor (solid) and to its prediction (dashed).}
\label{fig:pca}
\end{minipage}\hfill
\begin{minipage}[t]{0.48\textwidth}
\caption{Extrapolation varies widely by domain (Llama-3.3-70B). Hit@5 (\%) on unseen problems. Dashed line: chance (5 of 128 = 3.9\%). Error bars: $\pm$~SE.}
\label{fig:per_domain}
\end{minipage}
\end{figure}

\paragraph{Accuracy}
For every encoder but the smallest, MPNet (Table~\ref{tab:encoder}), Interpolation Hit@5 is at least $99.5\%$ ($99.7\%$ with Llama-3.3-70B, mean Hit@1 $92.1\%$ in Table~\ref{tab:single_step}), and predictions land next to their targets (Figure~\ref{fig:pca}). MPNet scores near zero on Rovers and Satellite even along plans of observed problems (Table~\ref{tab:plan_eval_encoders}), likely because MPNet reads at most 384 tokens and these domains' long prompts end with the state block.
Interpolation is the easy regime: test transitions come from training problems, a transition that recurs across plans can fall in both splits, and distractors come from the whole domain, so even predicting no change (Identity) reaches $73.7\%$ Hit@5 on the three domains of Table~\ref{tab:baselines_extrap}. Only learned transitions reach the ceiling, and Hit@1 separates them from the floor ($95.4\%$ against $46.9\%$, Table~\ref{tab:app_interp}).

\paragraph{Candidate-pool scaling}
Top-5 retrieval holds up when the 128-state pool grows to every observed state of the domain (Table~\ref{tab:pool_main}). Hit@1 falls about $3\times$, to $35.2\%$ on Ferry ($46{,}205$ states) and $32.6\%$ on Logistics ($13{,}373$), whereas Hit@5 falls from $100.0\%$ to $86.9\%$ and from $99.9\%$ to $74.9\%$, and Hit@10 stays at $95.0\%$ and $88.0\%$ (Appendix~\ref{app:pool_size}). The true successor thus stays in a short list against the full pool, though not reliably at rank one.

\paragraph{Comparison to LLMs}
Within observed problems, \textsc{EmbedPlan} matches or beats every LLM we test (Table~\ref{tab:llm_baselines}). We compare on Ferry and Logistics, two domains with different state-tracking structure, in two tasks. In \emph{generation}, the LLM receives $s$ and $a$ and writes $s'$: all five do well on Logistics, but none matches the true Ferry state more than $56\%$ of the time. Because generation is harder than retrieval, we also give GPT-5.4 \emph{the identical ranking task} over the same 128 candidates: it selects the true successor for $44\%$ of Ferry and $92\%$ of Logistics queries, against $99.0\%$ and $96.3\%$ for \textsc{EmbedPlan} (on Logistics the margin is small next to the spread across our runs, $91.2$--$96.3\%$). The comparison is matched in task, not in information: \textsc{EmbedPlan} has observed other transitions of the same problems, whereas the LLM is zero-shot, and on unseen problems \textsc{EmbedPlan}'s Ferry Hit@1 falls to $12.0\%$ (Table~\ref{tab:full_metrics}). What one-time training buys within observed problems is accurate successor ranking $101\times$ faster end to end than generation through an LLM API ($1.9$~s per transition), and $11{,}215\times$ faster with cached state embeddings (Appendix~\ref{app:compute}).

\begin{table}[t]
\caption{Replacing each state prediction with the nearest real state before the next step (\emph{snapped}) keeps multi-step rollout within $92$--$99\%$ of the accuracy obtained with the true state at each step. Step-wise Hit@1 (\%), Interpolation, 1{,}000 candidate states, mean $\pm$ SE over seeds (Blocksworld: single run). Retention: snapped over true-state accuracy.}
\label{tab:closed_loop}
\centering\small
\begin{tabular}{@{}lccc@{}}
\toprule
\textbf{Regime} & \textbf{Ferry} & \textbf{Logistics} & \textbf{Blocksworld} \\
\midrule
True state at each step          & $95.3 \pm 0.5$ & $74.9 \pm 2.2$ & $80.3$ \\
Own prediction, snapped          & $94.2 \pm 0.7$ & $69.0 \pm 2.1$ & $78.4$ \\
Own prediction, not snapped      & $68.3 \pm 2.0$ & $40.0 \pm 1.9$ & $59.6$ \\
\midrule
Retention (snapped)              & $99\%$ & $92\%$ & $98\%$ \\
\bottomrule
\end{tabular}
\end{table}

\paragraph{Closed-loop rollout}
Supplying the true state at every step (teacher forcing) hides error accumulation, so we also feed the model's own output back as the next input (Table~\ref{tab:closed_loop}), in two regimes. \emph{With snapping}, each state prediction is replaced by the nearest real state the model retrieves, right or wrong, and that state becomes the next input, so retrieval errors propagate. \emph{Without snapping}, the raw predicted embedding is fed back and no candidate states are consulted at intermediate steps. With snapping, rollout retains $92$--$99\%$ of teacher-forced step accuracy on all three domains. Because the regimes coincide at depth~1 by construction, the aggregate dilutes the contrast: from depth~2 on, snapped rollout on Ferry retains $98\%$ of teacher-forced Hit@1 ($94.1\%$ vs.\ $96.0\%$), against $49\%$ ($46.7\%$) without snapping, whose latent drifts off the manifold of real state embeddings (cosine to the nearest real state $0.83$ after one step and ${\sim}0.55$ thereafter, Appendix~\ref{app:closed_loop}). Per-step snapping is therefore what keeps multi-step rollout accurate. The test trajectories are short (mean $2.2$ steps on Ferry), so this establishes stability over short horizons only.

\paragraph{Action disambiguation}
To test whether the model separates the effects of different actions, we apply every action applicable in $s$ and rank the actions by how close their predictions come to the true $s'$. Acc@$k$ is the fraction of queries whose true action ranks in the top $k$. Mean Acc@5 is $85.3\%$ within observed problems (Acc@1 $32.0\%$, Table~\ref{tab:single_step}) and $16.2\%$ on unseen problems, where training without the action loss reaches only $4.8\%$ (Appendix~\ref{app:ablation_action_loss}).

\subsection{Where generalization stops}
\label{sec:boundary}

Accuracy falls with each step down the exposure ladder (Table~\ref{tab:hierarchy}). Observing a domain lifts Hit@5 from $6.6\%$ (Cross-Domain) to $54.6\%$ on its unseen problems, and observing the problems themselves lifts it to $99.7\%$. The second gap also includes the harder, same-problem distractors of Extrapolation, so it upper-bounds the cost of unseen problems alone.

\begin{table}[t]
\begin{minipage}[t]{0.45\textwidth}\centering
\caption{Hit@5 stays near chance until the target domain has been observed. Llama-3.3-70B, mean $\pm$ SE across domains. Rows are ordered by what the test data share with training, and $\Delta$ is the gain over chance in percentage points (pp). Plan-Variant faces the hardest distractors (Table~\ref{tab:protocols}).}
\label{tab:hierarchy}
\small\setlength{\tabcolsep}{3.5pt}
\begin{tabular}{@{}llcc@{}}
\toprule
\textbf{Protocol} & \textbf{Observed} & \textbf{Hit@5} & \textbf{$\Delta$} \\
\midrule
Interpolation & Problems & $99.7 \pm 0.1$ & $+96$ \\
Plan-Variant  & Problems & $51.2 \pm 5.5$ & $+47$ \\
Extrapolation & Domain   & $54.6 \pm 5.5$ & $+51$ \\
Multi-Domain  & Domain   & $37.2 \pm 3.8$ & $+33$ \\
Leave-One-Out & 8 others & $9.2 \pm 1.2$  & $+5.3$ \\
Cross-Domain  & 1 other  & $6.6 \pm 0.5$  & $+2.7$ \\
\midrule
Chance        &          & $3.9$          &  \\
\bottomrule
\end{tabular}
\end{minipage}\hfill
\begin{minipage}[t]{0.52\textwidth}\centering
\caption{Larger encoders extrapolate better, but none closes the gap. Hit@5 (\%), mean $\pm$ SE across the nine domains. Gap in pp, with a paired $t$-test across domains: $^{**}p<0.01$, $^{***}p<0.001$.}
\label{tab:encoder}
\small\setlength{\tabcolsep}{2.5pt}
\begin{tabular}{@{}lrccc@{}}
\toprule
\textbf{Encoder} & \textbf{Params} & \textbf{Interp.} & \textbf{Extrap.} & \textbf{Gap} \\
\midrule
MPNet         & 110M & $70.0 \pm 15$  & $26.8 \pm 6.3$ & $43^{**}$ \\
BGE-M3        & 568M & $99.6 \pm 0.2$ & $36.3 \pm 5.0$ & $63^{***}$ \\
Qwen2.5-7B    & 7B   & $99.5 \pm 0.2$ & $47.7 \pm 4.9$ & $52^{***}$ \\
Llama-3.3-70B & 70B  & $99.7 \pm 0.1$ & $54.6 \pm 5.5$ & $45^{***}$ \\
\bottomrule
\end{tabular}
\end{minipage}
\end{table}

\paragraph{Unseen problems}
On unseen problems of an observed domain, Hit@5 is $54.6\%$ ($14\times$ chance) and well above the no-change floor at Hit@1 ($21.4\%$ against $3.3\%$ for Identity on the three reference domains of Table~\ref{tab:baselines_extrap}). Scaling the encoder helps with diminishing returns (from $26.8\%$ for MPNet to $54.6\%$ for Llama-3.3-70B, Table~\ref{tab:encoder}), and no encoder closes the gap. Difficulty depends more on the domain than on the encoder ($26$--$76\%$, Figure~\ref{fig:per_domain} and Table~\ref{tab:per_domain_full}): grid-based domains with local, predictable effects (Goldminer, Grid) extrapolate best, and Depot, which requires compositional multi-object reasoning, extrapolates worst. When the true successor misses the top five, the top-ranked state typically differs from it in one to three facts (median~2, Appendix~\ref{app:error_analysis}).

\paragraph{Along whole plans}
We also score every step of held-out optimal plans with teacher forcing. Under Plan-Variant the plans are unseen but their problems are observed, and each step is ranked against successors along alternative optimal plans, the hardest distractors we use: they are reachable and often differ from the target in a single fact. Mean per-step Hit@5 is $51.2\%$, and $22.6\%$ of plans are correct at every step. Plans of unseen problems fare far worse, at $10.1\%$ per step and $3.3\%$ of plans (Table~\ref{tab:plan_eval_llama}), with the same collapse for every encoder (Table~\ref{tab:plan_eval_encoders}).

\paragraph{Unseen domains}
Transfer across domains fails almost completely. Cross-Domain reaches $6.6\%$, only $2.7$~pp above chance (the untrained model scores $4.2\%$, Appendix~\ref{app:untrained}). The full $9\times9$ matrix (Table~\ref{tab:cross_domain_matrix}) shows one notable success, Ferry$\rightarrow$Logistics ($22.3\%$), two domains that both transport objects between locations. Leave-One-Out fares little better at $9.2\%$ despite eight training domains (Table~\ref{tab:loo_results}). All nine domains share discrete states, typed objects, and precondition--effect schemas, yet the learned transition captures essentially none of this commonality. We see two plausible causes, which our experiments do not separate. First, frozen embeddings organize states by surface form, which differs across domains, rather than by structural role (Appendix~\ref{app:pca}): nothing tells the model that \texttt{pick-up(BlockA)} and \texttt{board(Car1)} follow the same precondition--effect pattern. Second, the learned action semantics are domain-specific: the action disambiguation loss improves generalization within a domain (Appendix~\ref{app:ablation_action_loss}) by learning fine distinctions among one domain's actions, which transfer poorly to another's. Training one model on all nine domains is a partial remedy: it reaches $37.2\%$, which is $17$~pp below the single-domain models (Table~\ref{tab:multi_domain}), keeping about two-thirds of their accuracy with one model instead of nine.

\subsection{What limits transfer}
\label{sec:why}

\begin{table}[t]
\caption{With the head, data, and pool fixed, the state representation decides transfer to unseen problems. Extrapolation, Hit@5 (\%) per domain (mean $\pm$ SE over 3 seeds) and three-domain means. The learned rows train the identical head on a different state representation (lifted STRIPS induction instead induces symbolic effects), and the first group requires the symbolic facts of each state. On these templated states an action edits only a few facts, which favors sparse and symbolic representations (Section~\ref{sec:why}). Bold: best without symbolic facts.}
\label{tab:baselines_extrap}
\centering\small\setlength{\tabcolsep}{5pt}
\begin{tabular}{@{}lccccc@{}}
\toprule
& \multicolumn{4}{c}{\textbf{Hit@5}} & \textbf{Hit@1} \\
\cmidrule(lr){2-5}\cmidrule(l){6-6}
\textbf{Method} & \textbf{Ferry} & \textbf{Logistics} & \textbf{Goldminer} & \textbf{Mean} & \textbf{Mean} \\
\midrule
\multicolumn{6}{@{}l}{\textit{Given the symbolic facts of each state}} \\
Lifted STRIPS induction (oracle) & $100.0 \pm 0.0$ & $100.0 \pm 0.0$ & $99.3 \pm 0.3$ & $99.8$ & $99.8$ \\
Bag-of-literals  & $51.9 \pm 5.7$ & $81.4 \pm 0.7$ & $95.7 \pm 2.0$ & $76.3$ & $37.1$ \\
\multicolumn{6}{@{}l}{\textit{Sparse lexical features of the same text}} \\
Char 3--5 grams  & $71.1 \pm 1.7$ & $69.8 \pm 7.4$ & $97.3 \pm 0.7$ & $\mathbf{79.4}$ & $41.9$ \\
Bag-of-words     & $70.8 \pm 2.7$ & $67.3 \pm 4.8$ & $99.3 \pm 0.2$ & $79.1$ & $\mathbf{45.0}$ \\
TF-IDF bigrams   & $59.5 \pm 2.1$ & $50.0 \pm 4.3$ & $97.2 \pm 1.2$ & $68.9$ & $37.2$ \\
\multicolumn{6}{@{}l}{\textit{Frozen LLM embeddings}} \\
\textsc{EmbedPlan} (Llama-3.3-70B) & $40.6 \pm 3.5$ & $53.7 \pm 9.8$ & $76.2 \pm 5.2$ & $56.8$ & $21.4$ \\
\textsc{EmbedPlan} (BGE-M3)        & $42.8 \pm 2.1$ & $51.7 \pm 3.9$ & $70.2 \pm 5.2$ & $54.9$ & $19.9$ \\
\midrule
Offset, lifted   & $34.3 \pm 0.8$ & $56.4 \pm 5.4$ & $50.7 \pm 1.7$ & $47.2$ & $5.5$ \\
Offset, grounded & $28.6 \pm 2.1$ & $53.1 \pm 2.3$ & $57.3 \pm 4.1$ & $46.3$ & $7.2$ \\
Identity         & $18.5 \pm 1.0$ & $49.8 \pm 3.4$ & $41.2 \pm 1.6$ & $36.5$ & $3.3$ \\
\midrule
Random vectors [null] & $4.4 \pm 0.5$ & $4.8 \pm 0.6$ & $5.2 \pm 0.5$ & $4.8$ & $0.7$ \\
Context-only [null]   & $0.0 \pm 0.0$ & $0.0 \pm 0.0$ & $0.0 \pm 0.0$ & $0.0$ & $0.0$ \\
\bottomrule
\end{tabular}
\end{table}

Because the head is independent of the encoder, the framework can attribute the transfer gap to its source. We train the same head on the same data, rank against the same pool, and vary only the state representation (Table~\ref{tab:baselines_extrap}), on three domains that span the observed extrapolation difficulty: Ferry, Logistics, and Goldminer.

\emph{Learning beats no change and lookup.} On unseen problems the learned transition improves on Identity by $20$~pp Hit@5 ($18$~pp Hit@1) and on the best Offset by $10$~pp ($14$~pp). A tabular one-hot model trained on identical data, which has no representation for states it has not seen, reaches $8.3\%$ Hit@1 on Logistics even within observed problems (against $96.3\%$ for \textsc{EmbedPlan}) and $0.2\%$ on unseen problems, below the $0.8\%$ chance level (Appendix~\ref{app:baselines}).
\emph{The head is not the main limit.} Fed character 3--5 grams of the same text, the identical head reaches $79.4\%$ Hit@5 ($41.9\%$ Hit@1), against $56.8\%$ ($21.4\%$) with Llama-3.3-70B embeddings, and given each state's literal decomposition, which our text interface does not require, lifted STRIPS induction recovers the dynamics from the same data exactly ($99.8\%$). Both advantages come from how the benchmark renders states. Every state is generated from a fixed template, so an action edits only a few facts of an otherwise identical text: \texttt{pick-up(C)} replaces ``arm empty'', ``C clear'', and ``C on table'' with ``arm holding C'' and leaves every other fact unchanged. Sparse features encode such an edit directly as a few changed dimensions, and a symbolic learner reads it off the literals, whereas a pooled LLM embedding of the whole prompt, most of which is shared problem and goal text, barely moves (cosine $0.997$--$0.999$ between consecutive states with BGE-M3). In free-form descriptions, where a fact can be phrased in many ways and no literal decomposition is given, neither shortcut is available, and that is the regime the text interface targets (Section~\ref{sec:limitations}). Adapting the encoder already helps: low-rank adaptation (LoRA) of BGE-M3 raises full-pool Hit@1 on unseen problems from $3.8\%$ to $9.3\%$ when the head is first trained with the encoder frozen ($6.2\%$ with a cold joint start, Appendix~\ref{app:finetune}), and larger frozen encoders extrapolate better (Table~\ref{tab:encoder}).

Two observations show what a better encoder must capture. First, transfer to a new problem means handling new objects: most held-out transitions use ground actions never seen in training ($57.6\%$ on Ferry, $65.5\%$ on Logistics), although no action schema is new (Appendix~\ref{app:grounding}), and the lifted STRIPS learner handles them by abstracting objects into schema parameters. Second, reordering a state's facts moves a BGE-M3 embedding $1.13\times$ further than applying an action (Appendix~\ref{app:factorder}), whereas bags of words, character n-grams, and literals are largely insensitive to fact order. Because the encoder is a plug-in, the results name a concrete target, an object-aware, order-invariant state encoder, which the framework can evaluate directly.

\section{Conclusion}

We introduced \textsc{EmbedPlan}, a transition model built on frozen text embeddings, and mapped its boundaries across 9 domains, 4 encoders, 6 protocols of increasing exposure, and reference methods from no-change floors to symbolic induction. Within observed problems, the component has the properties search needs: transition prediction is near-perfect ($99.7\%$ Hit@5), the true next state stays in the top five for most queries even when every observed state is a candidate, chaining predictions over several steps keeps $92$--$99\%$ of single-step accuracy, and \textsc{EmbedPlan} beats GPT-5.4 at picking the true next state from the same candidates while taking about $0.17$~ms per transition with cached state embeddings. On unseen problems it remains well above chance and the no-change floor ($54.6\%$ Hit@5), and on unseen domains it stays near chance. Because the same head can be trained over any representation, the framework also locates this boundary in the state representation. On our templated states, an action edits only a few facts, which sparse lexical and symbolic representations capture directly and pooled LLM embeddings barely register. State encoders that bind objects to roles and ignore fact order are therefore a concrete target for transfer to new problems. More broadly, separating a frozen encoder from a cheap learned transition makes text-based planning modular: any new text encoder can be dropped in and measured with the same head and protocols.

\subsection{Limitations and future work}
\label{sec:limitations}

\textsc{EmbedPlan} is a transition component for discrete, domain-specific settings, not a domain-general planner, and must be trained per domain because cross-domain transfer fails. Retrieval assumes a candidate pool that contains the true successor. Our domains are templated, so we have not yet tested the regime the text interface is meant for: states without a clean literal decomposition, where symbolic induction and lexical features would break. Several results also rest on narrower evidence: pool scaling, multi-step rollout, and LLM ranking are measured within observed problems, rollouts cover short test trajectories (mean $2.2$ steps), Extrapolation rests on one or two held-out problems per domain, and the reference comparison covers three domains, with two of three seeds sharing their held-out problems (Appendix~\ref{app:seed_variance}) and without ablating the problem and goal text that every prompt shares. Integrating \textsc{EmbedPlan} into beam or tree search over long horizons on held-out problems is the direct next step.

\section*{AI use statement}
We used AI assistants to edit author-written text and figures for clarity and structure, to help explore the relevant literature, and for coding assistance in implementing the experimental pipeline. The research questions, method, evaluation protocols, experiments, and results are the authors' own, and every reported number was produced by the experimental code and runs described in Section~\ref{sec:setup} and the appendix. The authors reviewed all AI-assisted text, figures, and code and take full responsibility for the content of this paper.

\subsection*{Ethics statement}

This work studies transition prediction on classical planning benchmarks rendered as natural language. It involves no human subjects, no personally identifiable information, and no sensitive data. All domains and problem instances are drawn from publicly available planning benchmarks, and the pre-trained encoders we use are publicly released models. We see no direct pathway from this study to harmful applications beyond those already associated with general-purpose language models. The reported failure modes under cross-domain transfer are stated explicitly so that the method is not over-claimed for safety-critical planning.

\subsection*{Reproducibility statement}
The experimental protocols, dataset construction, and splits are described in Section~\ref{sec:setup}, with full hyperparameters, encoder configurations, and per-domain results given in the appendix. All evaluation protocols are defined precisely enough to be re-implemented, and the complete numerical results underlying every figure and table in the main paper are reported in the appendix. Source code for \textsc{EmbedPlan} and the experiments is available at \url{https://github.com/embedplan/EmbedPlan}.

\appendix

\FloatBarrier
\section{Evaluation protocols}
\label{app:protocols}

These protocols form a diagnostic hierarchy for capability characterization. We deliberately isolate transition accuracy from search algorithms, candidate generation, and latency measurement to provide a clean assessment of what frozen embeddings support. End-to-end planning evaluation, while important for deployed systems, would conflate these factors.

We evaluate \textsc{EmbedPlan} under six protocols that measure progressively more demanding generalization capabilities. Let $\mathcal{D} = \{d_1, \ldots, d_9\}$ denote the set of nine planning domains. For each domain $d \in \mathcal{D}$, let $\mathcal{P}_d = \{p_1^d, \ldots, p_{n_d}^d\}$ denote the set of problem instances, where $n_d = |\mathcal{P}_d|$. Each problem $p \in \mathcal{P}_d$ yields a set of transitions $\mathcal{T}_p = \{(s_i, a_i, s'_i)\}_{i=1}^{m_p}$ extracted from optimal plan trajectories, where $m_p = |\mathcal{T}_p|$.

The complete transition set for domain $d$ is $\mathcal{T}_d = \bigcup_{p \in \mathcal{P}_d} \mathcal{T}_p$, and the global transition set across all domains is $\mathcal{T} = \bigcup_{d \in \mathcal{D}} \mathcal{T}_d$.

\subsection{Protocol 1: Interpolation split}
\label{app:proto_interp}

\paragraph{Definition}
Transitions are randomly partitioned regardless of problem or domain membership:
\begin{align}
    \mathcal{T}_d &= \mathcal{T}_d^{\text{train}} \cup \mathcal{T}_d^{\text{test}}, \quad \mathcal{T}_d^{\text{train}} \cap \mathcal{T}_d^{\text{test}} = \emptyset, \\
    |\mathcal{T}_d^{\text{train}}| &\approx 0.8 \, |\mathcal{T}_d|, \quad |\mathcal{T}_d^{\text{test}}| \approx 0.2 \, |\mathcal{T}_d|,
\end{align}
where each transition $(s, a, s') \in \mathcal{T}_d$ is assigned to train or test uniformly at random.

\paragraph{Key property}
For at least one problem $p$, transitions appear in both splits:
\begin{equation}
    \exists \, p \in \mathcal{P}_d : \; \mathcal{T}_p \cap \mathcal{T}_d^{\text{train}} \neq \emptyset \; \land \; \mathcal{T}_p \cap \mathcal{T}_d^{\text{test}} \neq \emptyset .
\end{equation}

\paragraph{Measures}
\emph{Interpolation} within known problem manifolds. Models can succeed by recognizing which problem instance a state belongs to and applying memorized problem-specific transition patterns.

\subsection{Protocol 2: Plan-Variant split}
\label{app:proto_planvariant}

\paragraph{Definition}
We evaluate full-plan execution on alternative optimal plans for problems seen during training. Let $\Pi_p = \{\pi_1, \pi_2, \ldots, \pi_k\}$ denote the set of optimal plans for problem $p$, where each plan $\pi_i$ is a sequence of actions leading from the initial state to the goal. Plans are partitioned:
\begin{align}
    \Pi_p &= \Pi_p^{\text{train}} \cup \Pi_p^{\text{test}}, \quad \Pi_p^{\text{train}} \cap \Pi_p^{\text{test}} = \emptyset, \\
    \mathcal{T}_p^{\text{train}} &= \bigcup_{\pi \in \Pi_p^{\text{train}}} \mathcal{T}_\pi, \quad \mathcal{T}_p^{\text{test}} = \bigcup_{\pi \in \Pi_p^{\text{test}}} \mathcal{T}_\pi,
\end{align}
where $\mathcal{T}_\pi$ denotes the transitions extracted from plan $\pi$. Optimal plans are obtained directly from the dataset. For problems admitting multiple optimal solutions, we partition plans such that train and test contain non-overlapping action sequences. Problems with unique optimal plans are excluded from this protocol.

\paragraph{Key property}
Train and test share the same underlying problem instances but differ in the action sequences used to reach the goal:
\begin{equation}
    \mathcal{P}^{\text{train}} = \mathcal{P}^{\text{test}} = \mathcal{P}_d, \quad \text{but} \quad \Pi_p^{\text{train}} \cap \Pi_p^{\text{test}} = \emptyset \;\; \forall \, p \in \mathcal{P}_d .
\end{equation}

\paragraph{Measures}
\emph{Generalization to unseen solution paths under fixed problem structure.} Models must learn transition dynamics that transfer across different valid action sequences within the same problem, rather than memorizing specific plan trajectories. This tests whether the model captures the underlying state-action mechanics or merely overfits to observed plans.

\subsection{Protocol 3: Extrapolation split}
\label{app:proto_extrap}

\paragraph{Definition}
Problems are partitioned, and all transitions from each problem go exclusively to train or test:
\begin{align}
    \mathcal{P}_d &= \mathcal{P}_d^{\text{train}} \cup \mathcal{P}_d^{\text{test}}, \quad \mathcal{P}_d^{\text{train}} \cap \mathcal{P}_d^{\text{test}} = \emptyset, \\
    \mathcal{T}_d^{\text{train}} &= \bigcup_{p \in \mathcal{P}_d^{\text{train}}} \mathcal{T}_p, \quad \mathcal{T}_d^{\text{test}} = \bigcup_{p \in \mathcal{P}_d^{\text{test}}} \mathcal{T}_p,
\end{align}
with $|\mathcal{P}_d^{\text{train}}| \approx 0.8 \, |\mathcal{P}_d|$.

\paragraph{Key property}
Train and test transitions come from disjoint problem sets:
\begin{equation}
    \forall \, p \in \mathcal{P}_d^{\text{test}} : \; \mathcal{T}_p \cap \mathcal{T}_d^{\text{train}} = \emptyset .
\end{equation}

\paragraph{Measures}
\emph{Extrapolation} to new problem configurations within a known domain. Models must learn transferable domain dynamics (e.g., ``pick-up removes a block from a surface'') rather than problem-specific patterns.

\subsection{Protocol 4: Multi-Domain learning (unified model)}
\label{app:proto_multi}

\paragraph{Definition}
We train a single model on all domains simultaneously, with Extrapolation splits within each domain:
\begin{align}
    \mathcal{T}^{\text{train}} &= \bigcup_{d \in \mathcal{D}} \mathcal{T}_d^{\text{train}}, \quad \text{where } \mathcal{T}_d^{\text{train}} = \bigcup_{p \in \mathcal{P}_d^{\text{train}}} \mathcal{T}_p, \\
    \mathcal{T}^{\text{test}} &= \bigcup_{d \in \mathcal{D}} \mathcal{T}_d^{\text{test}}, \quad \text{where } \mathcal{T}_d^{\text{test}} = \bigcup_{p \in \mathcal{P}_d^{\text{test}}} \mathcal{T}_p .
\end{align}

A single transition network $T_\theta^{\text{multi}}$ is trained on $\mathcal{T}^{\text{train}}$ and evaluated per domain:
\begin{equation}
    \text{Multi-Domain}(d) = \text{Hit@}k\big(T_\theta^{\text{multi}}, \mathcal{T}_d^{\text{test}}\big).
\end{equation}

\paragraph{Key property}
A single network $T_\theta^{\text{multi}} : \mathbb{R}^{128} \times \mathbb{R}^{128} \to \mathbb{R}^{128}$ must represent the transition functions of all nine domains in one shared parameter space.

\paragraph{Measures}
\emph{Capacity sharing across domains.} Tests whether a unified 128-dimensional latent space can encode transition dynamics for multiple domains simultaneously.

\subsection{Protocol 5: Cross-Domain transfer (zero-shot)}
\label{app:proto_cross}

\paragraph{Definition}
We train on one source domain and evaluate on a different target domain:
\begin{align}
    \mathcal{T}^{\text{train}} &= \mathcal{T}_{d_{\text{src}}}, \quad d_{\text{src}} \in \mathcal{D}, \\
    \mathcal{T}^{\text{test}} &= \mathcal{T}_{d_{\text{tgt}}}, \quad d_{\text{tgt}} \in \mathcal{D} \setminus \{d_{\text{src}}\} .
\end{align}

For comprehensive evaluation, we compute performance for all source-target pairs:
\begin{equation}
\text{Cross-Domain}(d_{\text{tgt}})
= \frac{1}{|\mathcal{D}|-1}
\sum_{d_{\text{src}} \neq d_{\text{tgt}}}
\text{Hit@}k\!\left(T_{\theta}^{d_{\text{src}}},\,\mathcal{T}_{d_{\text{tgt}}}\right),
\end{equation}
where $T_\theta^{d_{\text{src}}}$ denotes the transition network trained on domain $d_{\text{src}}$.

\paragraph{Key property}
Source and target share no domain, problem, or transition: $\mathcal{T}^{\text{train}} \cap \mathcal{T}^{\text{test}} = \emptyset$, and the object vocabularies and action schemas differ entirely.

\paragraph{Measures}
\emph{Zero-shot domain transfer.} Models must learn transition dynamics that generalize across fundamentally different planning structures (e.g., from block manipulation to logistics).

\subsection{Protocol 6: Leave-One-Out generalization}
\label{app:proto_loo}

\paragraph{Definition}
We train on $|\mathcal{D}| - 1$ domains and evaluate on the held-out domain:
\begin{align}
    \mathcal{T}^{\text{train}} &= \bigcup_{d \in \mathcal{D} \setminus \{d_{\text{held}}\}} \mathcal{T}_d, \\
    \mathcal{T}^{\text{test}} &= \mathcal{T}_{d_{\text{held}}} .
\end{align}

For comprehensive evaluation, we iterate over all held-out domains:
\begin{equation}
    \text{LOO}(d_{\text{held}}) = \text{Hit@}k\big(T_\theta^{\mathcal{D} \setminus \{d_{\text{held}}\}}, \mathcal{T}_{d_{\text{held}}}\big).
\end{equation}

\paragraph{Key property}
Unlike Cross-Domain, which uses a single source, LOO provides maximum training diversity:
\begin{equation}
    |\mathcal{T}^{\text{train}}| = \sum_{d \neq d_{\text{held}}} |\mathcal{T}_d| \; \gg \; |\mathcal{T}_{d_{\text{src}}}| .
\end{equation}

\paragraph{Measures}
\emph{Transfer from diverse training to an unseen domain.} Tests whether exposure to eight diverse domains enables generalization to an entirely novel ninth domain.

\subsection{Protocol hierarchy}

The six protocols form a hierarchy of increasing generalization difficulty.

\begin{table}[h]
\caption{Evaluation protocol hierarchy by generalization difficulty.}
\label{tab:protocol_hierarchy}
\begin{center}
\small
\begin{tabular}{@{}llcccc@{}}
\toprule
\textbf{Protocol} & \textbf{Measures} & \textbf{Shared} & \textbf{Shared} & \textbf{Shared} & \textbf{Difficulty} \\
 & & \textbf{domain} & \textbf{problems} & \textbf{plans} & \\
\midrule
Interpolation & Interpolation & Yes & Yes & Yes & Lowest \\
Plan-Variant & Plan generalization & Yes & Yes & No & Low \\
Extrapolation & Extrapolation & Yes & No & No & Medium \\
Multi-Domain & Capacity sharing & Yes & No & No & Medium \\
Cross-Domain & Zero-shot transfer & No & No & No & High \\
Leave-One-Out & Diverse transfer & No & No & No & Highest \\
\bottomrule
\end{tabular}
\end{center}
\end{table}

\paragraph{Design rationale}
This hierarchy isolates where frozen embeddings succeed versus fail:
\begin{itemize}
    \item \textbf{Interpolation} $\to$ \textbf{Plan-Variant}: quantifies overfitting to specific action sequences.
    \item \textbf{Plan-Variant} $\to$ \textbf{Extrapolation}: quantifies the interpolation-extrapolation gap.
    \item \textbf{Extrapolation} $\to$ \textbf{Multi-Domain}: tests capacity sharing.
    \item \textbf{Multi-Domain} $\to$ \textbf{Cross-Domain}/\textbf{LOO}: tests zero-shot transfer.
\end{itemize}

\FloatBarrier
\section{Experimental setup}
\label{app:experimental}

\subsection{Architecture}
\label{app:architecture}

Figure~\ref{fig:architecture_full} provides a complete illustration of the \textsc{EmbedPlan} architecture.

\begin{figure}[t]
\centering
\resizebox{\textwidth}{!}{%
\begin{tikzpicture}[
    box/.style={rectangle, draw, rounded corners=3pt, minimum height=1cm, minimum width=2.2cm, align=center, font=\small},
    frozen/.style={box, fill=blue!12, draw=blue!40, line width=0.8pt},
    learned/.style={box, fill=orange!15, draw=orange!50, line width=0.8pt},
    candidate/.style={box, fill=green!12, draw=green!40, line width=0.8pt},
    emb/.style={circle, draw, minimum size=0.8cm, font=\small, line width=0.6pt},
    arrow/.style={->, >=stealth, thick, line width=0.8pt},
    dashedarrow/.style={->, >=stealth, dashed, gray, line width=0.7pt},
    label/.style={font=\scriptsize, align=center}
]

\node[box, fill=gray!8] (state_text) at (0, 1) {State $s$\\[-1pt]{\scriptsize\texttt{"Block A is on B..."}}};
\node[box, fill=gray!8] (action_text) at (0, -1) {Action $a$\\[-1pt]{\scriptsize\texttt{"pick-up(C)"}}};

\node[frozen, minimum height=2.2cm] (encoder) at (3.5, 0) {Frozen LLM\\Encoder $E$};
\node[font=\footnotesize, blue!60] at (3.5, 1.5) {frozen};

\node[emb, fill=blue!5] (z_s) at (6, 0.7) {$\mathbf{z}_s$};
\node[emb, fill=blue!5] (z_a) at (6, -0.7) {$\mathbf{z}_a$};
\node[font=\scriptsize] at (6, 1.5) {768--8192d};

\node[learned, minimum width=1.2cm, minimum height=0.8cm] (proj_s) at (8, 0.7) {$\pi_s$};
\node[learned, minimum width=1.2cm, minimum height=0.8cm] (proj_a) at (8, -0.7) {$\pi_a$};
\node[font=\footnotesize, orange!60] at (8, 1.5) {learned};

\node[emb, fill=orange!8] (h_s) at (10, 0.7) {$\mathbf{h}_s$};
\node[emb, fill=orange!8] (h_a) at (10, -0.7) {$\mathbf{h}_a$};
\node[font=\scriptsize] at (10, 1.5) {128-d};

\node[learned, minimum width=2.5cm, minimum height=1.8cm] (transition) at (12.5, 0) {Transition\\Network $T_\theta$};

\node[emb, fill=orange!25, minimum size=0.9cm] (h_sp) at (15, 0) {$\hat{\mathbf{h}}_{s'}$};

\node[candidate, minimum height=2.4cm, minimum width=2cm] (candidates) at (17.5, 0) {Candidate\\Pool $\mathcal{C}$\\[-2pt]{\scriptsize $\mathbf{h}_{s'_1}, \mathbf{h}_{s'_2}, \ldots$}};

\node[candidate, fill=green!20, minimum width=1.8cm] (retrieved) at (20, 0) {Retrieved\\$\hat{s}'$};

\draw[arrow] (state_text.east) -- ++(0.8,0) |- (encoder.west);
\draw[arrow] (action_text.east) -- ++(0.8,0) |- (encoder.west);
\draw[arrow] (encoder.east) ++(0, 0.35) -- (z_s.west);
\draw[arrow] (encoder.east) ++(0, -0.35) -- (z_a.west);
\draw[arrow] (z_s) -- (proj_s);
\draw[arrow] (z_a) -- (proj_a);
\draw[arrow] (proj_s) -- (h_s);
\draw[arrow] (proj_a) -- (h_a);
\draw[arrow] (h_s.east) -- ++(0.5,0) |- ([yshift=3pt]transition.west);
\draw[arrow] (h_a.east) -- ++(0.5,0) |- ([yshift=-3pt]transition.west);
\draw[arrow] (transition.east) -- (h_sp.west);
\draw[arrow] (h_sp.east) -- (candidates.west) node[midway, above, font=\scriptsize] {sim};
\draw[arrow] (candidates.east) -- (retrieved.west) node[midway, above, font=\scriptsize] {argmax};
\draw[dashedarrow] (h_s.south) -- ++(0, -1.2) -| ([xshift=-5pt]transition.south);
\node[font=\scriptsize, gray] at (11.2, -1.5) {residual};

\node[align=center, font=\small] (loss) at (12.5, -3) {%
    \textbf{InfoNCE loss:} $\displaystyle\mathcal{L}_{\text{state}} = -\log \frac{\exp(\mathrm{sim}(\hat{\mathbf{h}}_{s'}, \mathbf{h}_{s'})/\tau)}{\sum_j \exp(\mathrm{sim}(\hat{\mathbf{h}}_{s'}, \mathbf{h}_{s'_j})/\tau)}$
};
\draw[dotted, gray, line width=0.6pt] (h_sp.south) -- ++(0, -0.8) -- (loss.north);

\end{tikzpicture}
}
\caption{Complete \textsc{EmbedPlan} architecture. State and action descriptions are encoded by a frozen LLM encoder $E$ into high-dimensional embeddings $\mathbf{z}_s, \mathbf{z}_a$. Learned projection heads $\pi_s, \pi_a$ reduce dimensionality to a shared 128-d space. The transition network $T_\theta$ (with a residual connection from $\mathbf{h}_s$) predicts the next-state embedding $\hat{\mathbf{h}}_{s'}$, trained via InfoNCE to maximize similarity to the ground-truth embedding. At inference, the model retrieves the most similar state from a candidate pool.}
\label{fig:architecture_full}
\end{figure}

\subsubsection{Frozen encoders}
\label{app:encoders}

States and actions are encoded using frozen pre-trained language models. We experiment with four encoders spanning different architectures and scales.

\paragraph{MPNet (all-mpnet-base-v2)}
A 110-million parameter sentence transformer~\citep{reimers2019sbert} trained on over 1 billion sentence pairs. This model produces 768-dimensional embeddings optimized for semantic similarity tasks. We use the \texttt{sentence-transformers} library with default mean pooling.

\paragraph{BGE-M3 (BAAI/bge-m3)}
A multilingual embedding model (568M parameters) designed for multi-granularity retrieval. We use the dense retrieval component, which extracts 1,024-dimensional embeddings from the \texttt{[CLS]} token of the final layer.

\paragraph{Qwen2.5-7B-Instruct}
A 7-billion parameter instruction-tuned autoregressive model. We extract sentence embeddings by applying mean pooling over the final layer's hidden states, producing 3,584-dimensional embeddings.

\paragraph{Llama-3.3-70B-Instruct}
A 70-billion parameter autoregressive language model. Since decoder-only models do not have a natural sentence embedding, we apply mean pooling over the final layer's hidden states across all input tokens:
\begin{equation}
    \mathbf{z} = \frac{1}{T} \sum_{t=1}^{T} \mathbf{h}_t^{(L)},
\end{equation}
where $\mathbf{h}_t^{(L)} \in \mathbb{R}^{8192}$ is the hidden state at position $t$ in the final layer $L$, and $T$ is the sequence length. Mean pooling over decoder-only hidden states is a common approximation when dedicated sentence embeddings are unavailable~\citep{muennighoff2022mteb}.

\paragraph{Rationale for freezing}
We freeze encoders for two reasons. First, it isolates our research question: we test whether \emph{existing} pre-trained representations support transition learning, without confounding this with task-specific fine-tuning that might introduce planning structure. Second, it enables efficient experimentation: embeddings are computed once and cached, reducing each training run from hours of LLM inference to minutes of lightweight optimization.

\paragraph{Pooling strategy rationale}
We adopt mean pooling uniformly across decoder-only models for consistency, acknowledging that this is one of several reasonable choices. Instruction-tuned sentence encoders or last-token pooling with explicit EOS prompting may yield embeddings better suited for semantic similarity, so our results represent a lower bound on what optimized embedding extraction could achieve. We prioritize architectural consistency across encoders over per-encoder optimization, and leave pooling sensitivity analysis to future work.

\subsubsection{Projection heads}
\label{app:projection}

High-dimensional encoder outputs present computational and statistical challenges. We introduce learnable projection heads that map to a lower-dimensional space where transition learning occurs.

Each projection head is a multi-layer perceptron:
\begin{equation}
    \pi(\mathbf{z}) = W_L \, \sigma\big(\mathrm{LN}(W_{L-1} \, \sigma(\mathrm{LN}(W_1 \mathbf{z})))\big),
\end{equation}
where $\sigma(\cdot)$ denotes the GELU activation and $\mathrm{LN}(\cdot)$ denotes layer normalization. We use separate projection heads for states ($\pi_s$) and actions ($\pi_a$), allowing each to learn modality-specific transformations.

\paragraph{Default configuration}
\begin{itemize}
    \item Input dimension: encoder-dependent (768 for MPNet, 1,024 for BGE-M3, 3,584 for Qwen2.5-7B, 8,192 for Llama-3.3-70B)
    \item Output dimension: 128
    \item Hidden layers: 4
    \item Activation: GELU
    \item Normalization: LayerNorm after each hidden layer
\end{itemize}

\subsubsection{Transition network}
\label{app:transition}

We investigate two architectures embodying different hypotheses about how actions transform states.

\paragraph{Residual MLP (primary)}
Our primary architecture processes the concatenated state-action representation through a feedforward network with a residual connection:
\begin{equation}
    \hat{\mathbf{h}}_{s'} = \mathrm{LN}\Big( f_\theta\big([\mathbf{h}_s, \mathbf{h}_a]\big) + W_{\text{res}} \, \mathbf{h}_s \Big).
\end{equation}

The feedforward network $f_\theta$ has architecture
\begin{equation}
    f_\theta(\mathbf{x}) = W_3 \, \sigma\big( W_2 \, \sigma(W_1 \mathbf{x}) \big),
\end{equation}
with dimensions $W_1 : \mathbb{R}^{256} \to \mathbb{R}^{128}$, $W_2 : \mathbb{R}^{128} \to \mathbb{R}^{128}$, and $W_3 : \mathbb{R}^{128} \to \mathbb{R}^{128}$.

The residual term $W_{\text{res}} \mathbf{h}_s$ encodes an inductive bias: actions produce \emph{incremental modifications} to states rather than wholesale replacements. Most predicates remain unchanged after a single action, and only a few flip.

\paragraph{Hypernetwork (alternative)}
An alternative hypothesis is that different actions require fundamentally different transformations. We implement this via a hypernetwork~\citep{ha2017hypernetworks} that generates action-conditioned modulation parameters:
\begin{equation}
    g_\phi(\mathbf{h}_a) = W_g^{(2)} \, \sigma\big( W_g^{(1)} \mathbf{h}_a \big) \in \mathbb{R}^{2 L \cdot d_{\text{adapt}}} .
\end{equation}
This output is split into $L$ pairs of scale and shift vectors $(\mathbf{A}_i, \mathbf{b}_i)$ for FiLM-style conditioning~\citep{perez2018film}.

\subsection{Training details}
\label{app:training}

\subsubsection{Loss formulations}
\label{app:loss_formulations}

We optimize the composite objective $\mathcal{L} = \mathcal{L}_{\text{state}} + \lambda \mathcal{L}_{\text{action}}$ with $\lambda = 2$.

The state prediction loss is InfoNCE~\citep{oord2018cpc} over batch elements:
\begin{equation}
    \mathcal{L}_{\text{state}} = -\frac{1}{B} \sum_{i=1}^{B} \log \frac{\exp\big(\mathrm{sim}(\hat{\mathbf{h}}_{s'_i}, \mathbf{h}_{s'_i}) / \tau\big)}{\sum_{j=1}^{B} \exp\big(\mathrm{sim}(\hat{\mathbf{h}}_{s'_i}, \mathbf{h}_{s'_j}) / \tau\big)},
\end{equation}
where $\mathrm{sim}(\mathbf{u}, \mathbf{v}) = \mathbf{u}^\top \mathbf{v} / (\|\mathbf{u}\| \|\mathbf{v}\|)$ is cosine similarity, $\tau = 0.07$ is the temperature, and $B = 128$ is the batch size.

The action disambiguation loss contrasts predictions from different actions applied to the same state:
\begin{equation}
\begin{aligned}
\mathcal{L}_{\text{action}} &= -\frac{1}{B}\sum_{i=1}^{B}\log \frac{\exp(z_{i,i}/\tau)}{\sum_{k=1}^{K}\exp(z_{i,k}/\tau)}, \\
z_{i,k} &:= \mathrm{sim}\!\big(\hat{\mathbf{h}}_{s'_i}^{(a_k)},\,\mathbf{h}_{s'_i}\big),
\end{aligned}
\end{equation}
where $\{\hat{\mathbf{h}}_{s'}^{(a_k)}\}_{k=1}^{K}$ are predictions from applying each of up to $K = 50$ grounded actions to state $s$.

\paragraph{In-batch negatives}
The denominator of $\mathcal{L}_{\text{state}}$ treats all $B$ states in the batch as candidates, with the $B{-}1$ non-matching states serving as negatives. This provides 127 negatives per positive without additional computation.

\paragraph{Why contrastive?}
Contrastive learning offers advantages over regression losses: (1) scale invariance via cosine similarity, (2) direct optimization of the retrieval objective, and (3) natural hard negative mining from same-domain states in the batch.

\subsubsection{Hyperparameter configuration}
\label{app:hyperparameters}

Table~\ref{tab:hyperparameters} lists the hyperparameter search space and final configuration.

\begin{table}[h]
\caption{Hyperparameter search space. Bold values indicate the final configuration.}
\label{tab:hyperparameters}
\begin{center}
\small
\begin{tabular}{@{}ll@{}}
\toprule
\textbf{Hyperparameter} & \textbf{Values explored} \\
\midrule
\multicolumn{2}{c}{\textit{Architecture}} \\
\midrule
Model type & \textbf{MLP}, hypernetwork \\
Hidden size & \textbf{128}, 256 \\
Number of layers & \textbf{2}, 4 \\
Dropout & \textbf{0.0}, 0.5 \\
Layer normalization & \textbf{Yes} \\
\midrule
\multicolumn{2}{c}{\textit{Loss function}} \\
\midrule
Action loss weight ($\lambda$) & 0, 0.5, 1, 1.5, \textbf{2}, 4 \\
\midrule
\multicolumn{2}{c}{\textit{Projection head}} \\
\midrule
Use projection & \textbf{Yes}, no \\
Projection dimension & \textbf{128} \\
Projection layers & 2, \textbf{4} \\
\midrule
\multicolumn{2}{c}{\textit{Optimization}} \\
\midrule
Learning rate & $2{\times}10^{-3}$, $1{\times}10^{-3}$, $\mathbf{4{\times}10^{-5}}$ \\
Batch size & 64, \textbf{128} \\
InfoNCE temperature ($\tau$) & \textbf{0.07} \\
Weight decay & $\mathbf{1{\times}10^{-2}}$ \\
\midrule
\multicolumn{2}{c}{\textit{Training}} \\
\midrule
Max epochs & \textbf{400} \\
Warmup epochs & \textbf{10} \\
Early stopping patience & \textbf{100} \\
\bottomrule
\end{tabular}
\end{center}
\end{table}

\subsection{Evaluation procedure}
\label{app:evaluation}

\paragraph{Hit@$k$ computation}
For each test transition $(s, a, s')$:
\begin{enumerate}
    \item Compute the predicted embedding $\hat{\mathbf{h}}_{s'} = T_\theta(\pi_s(\mathbf{z}_s), \pi_a(\mathbf{z}_a))$.
    \item Compute the similarity to all candidates, $\mathrm{sim}_j = \mathrm{sim}(\hat{\mathbf{h}}_{s'}, \mathbf{h}_{s'_j})$ for $s'_j \in \mathcal{C}$.
    \item Rank candidates by similarity in descending order.
    \item Record a hit if the correct $s'$ appears in the top $k$.
\end{enumerate}

The candidate pool $\mathcal{C}$ contains 128 states: the ground-truth next state plus 127 distractors sampled according to the evaluation protocol (uniformly from the domain for Interpolation, and from the same problem instance for Extrapolation).

\paragraph{Tie-breaking}
When multiple candidates have identical similarity scores, we select one of them uniformly at random. The reference-method comparison of Appendix~\ref{app:baselines} instead uses worst-case tie-breaking, which is required for its null controls to be meaningful (Appendix~\ref{app:ties}). Learned representations produce essentially no exact ties, so the convention does not affect their scores.

\paragraph{Action disambiguation}
Given a state $s$ and ground-truth next state $s'$, we apply every grounded action applicable in $s$, $\mathcal{A}(s)$, and compare each prediction with $s'$:
\begin{equation}
    \hat{a} = \argmax_{a \in \mathcal{A}(s)} \; \mathrm{sim}\big(T_\theta(\mathbf{h}_s, \mathbf{h}_a), \mathbf{h}_{s'}\big).
\end{equation}
Acc@$1$ measures how often $\hat{a}$ matches the true action. More generally, Acc@$k$ ranks all $a \in \mathcal{A}(s)$ by this similarity and measures how often the true action is among the top $k$.

\subsection{Candidate pool construction and size}
\label{app:pool_size}

Every Hit@$k$ in this paper is measured against a candidate pool, so the pool is part of
the metric. Two aspects matter independently: how distractors are \emph{drawn}, and how
\emph{many} there are.

\paragraph{Distractor difficulty}
The protocols of Appendix~\ref{app:protocols} already vary the first aspect. Under
Interpolation distractors are drawn uniformly from all domain states, so most share
neither objects nor goal with the query. Under Extrapolation they are drawn exclusively
from the query's own problem instance, so every candidate shares the object vocabulary
and differs from the target by a handful of predicates. Under the Plan-Variant protocol
the competing candidates are the successor states of alternative optimal plans, which are
reachable and often differ by a single literal. These three settings span easy, hard, and
adversarial distractors at a fixed pool size of 128.

\paragraph{Pool size}
The second aspect is held fixed at 128 throughout the main results, which is a controlled
measurement rather than a deployment setting. We therefore sweep the pool from 128 states
to every state observed in the domain on Ferry and Logistics, two domains whose observed
state sets differ by more than $3\times$ ($46{,}205$ and $13{,}373$ states). The split
is Interpolation and the encoder is Llama-3.3-70B. The transition model is unchanged and
only the retrieval pool grows.

\begin{table}[h]
\caption{Effect of candidate pool size under Interpolation, Llama-3.3-70B, mean $\pm$ SE
over 3 seeds. ``Full'' retrieves against every observed state in the domain. The
inflation column is the 128-way value divided by the full-pool value.}
\label{tab:app_pool_size}
\centering\small\setlength{\tabcolsep}{4pt}
\begin{tabular}{@{}llccccc|c@{}}
\toprule
\textbf{Domain} & \textbf{Metric} & \textbf{128} & \textbf{512} & \textbf{2{,}048} & \textbf{8{,}192} & \textbf{Full} & \textbf{Infl.} \\
\midrule
\multirow{3}{*}{\shortstack[l]{Ferry\\{\scriptsize 46{,}205 states}}}
 & Hit@1  & $98.96 \pm 0.09$ & $97.21 \pm 0.27$ & $90.90 \pm 0.55$ & $76.58 \pm 0.92$ & $35.23 \pm 1.14$ & $2.81\times$ \\
 & Hit@5  & $99.97 \pm 0.03$ & $99.85 \pm 0.08$ & $99.38 \pm 0.09$ & $97.97 \pm 0.25$ & $86.93 \pm 1.36$ & $1.15\times$ \\
 & Hit@10 & $99.99 \pm 0.01$ & $99.94 \pm 0.03$ & $99.73 \pm 0.08$ & $99.03 \pm 0.14$ & $94.96 \pm 0.63$ & $1.05\times$ \\
\midrule
\multirow{3}{*}{\shortstack[l]{Logistics\\{\scriptsize 13{,}373 states}}}
 & Hit@1  & $96.29 \pm 0.23$ & $88.21 \pm 0.65$ & $70.12 \pm 1.83$ & $47.20 \pm 1.60$ & $32.55 \pm 1.69$ & $2.96\times$ \\
 & Hit@5  & $99.85 \pm 0.08$ & $99.37 \pm 0.12$ & $96.73 \pm 0.31$ & $83.13 \pm 0.93$ & $74.89 \pm 1.71$ & $1.33\times$ \\
 & Hit@10 & $99.90 \pm 0.05$ & $99.72 \pm 0.09$ & $98.84 \pm 0.11$ & $92.99 \pm 0.50$ & $88.01 \pm 0.92$ & $1.14\times$ \\
\bottomrule
\end{tabular}
\end{table}

The premise holds: the 128-way pool inflates Hit@1 by $2.8\times$ on Ferry and $3.0\times$
on Logistics. Hit@$k$ for larger $k$ degrades far more gracefully: $1.15\times$ and
$1.33\times$ at $k{=}5$, and $1.05\times$ and $1.14\times$ at $k{=}10$. Against \emph{every}
observed state the model still places the correct successor in the top 5 for
$86.9\%$ of Ferry transitions and $74.9\%$ of Logistics transitions, and in the top 10
for $95.0\%$ and $88.0\%$. This is the empirical basis for emphasizing Hit@5 in
Section~\ref{sec:setup}: top-5 retrieval largely survives removing the 128-way pool,
whereas strict top-1 retrieval is substantially an artifact of it.

Logistics degrades faster in relative terms at intermediate sizes because $8{,}192$
candidates already cover $61\%$ of its state space, against $18\%$ of Ferry's. At the full pool the two domains converge to within three points of Hit@1 of each other.

\paragraph{When no pool is available}
The retrieval formulation presupposes an enumerable candidate set, which classical
planning benchmarks and discrete-observation model-based RL provide but open-ended
settings do not. The transition function itself does not depend on this: it maps
$(\mathbf{z}_s, \mathbf{z}_a)$ to a predicted next-state embedding and can be applied
recursively to its own output, yielding a latent rollout with no pool at any step.
Retrieval is then needed only where the trajectory must be read back out as text. We
report closed-loop rollouts with and without per-step snapping to a real state in
Appendix~\ref{app:closed_loop}. With snapping, rollout retains $92$--$99\%$ of teacher-forced accuracy. Without it, the latent drifts off the state manifold and error
compounds, which is the cost of dropping the pool entirely.

\subsection{Compute, latency, and cost analysis}
\label{app:compute}

All experiments were conducted on the following infrastructure:
\begin{itemize}
    \item \textbf{GPU}: NVIDIA A100 80\,GB
    \item \textbf{CPU}: AMD EPYC 7763 64-core processor
    \item \textbf{Memory}: 512\,GB RAM
    \item \textbf{Framework}: PyTorch 2.1, CUDA 12.1
\end{itemize}

\paragraph{Training time}
Training \textsc{EmbedPlan} is efficient because the heavy semantic processing is offloaded to the frozen encoder. Generating and caching embeddings represents the bulk of the preparation time, after which optimizing the lightweight transition network takes only minutes:
\begin{itemize}
    \item Embedding extraction (per domain, Llama-3.3-70B): $\sim$2 hours
    \item Embedding extraction (per domain, MPNet): $\sim$5 minutes
    \item Transition network training (per domain): $\sim$15 minutes
    \item Full experimental suite (all encoders, all protocols): $\sim$72 hours
\end{itemize}

\paragraph{Inference latency and efficiency}
A core motivation for \textsc{EmbedPlan} is bypassing the token-by-token generation bottleneck of standard LLM planning. To quantify this, we benchmarked the wall-clock latency of \textsc{EmbedPlan} against standard LLM API generation. As detailed in Table~\ref{tab:latency_cost}, even when accounting for the text-to-vector encoding overhead (full pipeline), our method is roughly 100$\times$ faster than standard LLM generation. When embeddings for the state-action space are pre-computed or batched, as is common in classical planning search algorithms, retrieval-based prediction drops to a fraction of a millisecond.

\begin{table}[h]
\caption{Wall-clock latency and computational cost comparison. The full pipeline includes text-to-vector encoding time with BGE-M3. \textsc{EmbedPlan} eliminates the sequential token bottleneck of autoregressive generation. $^\dagger$Projection heads, transition network, and scoring of 128 candidates with cached encoder outputs (Llama-3.3-70B dimensions). The full-pipeline latency additionally includes encoding the query texts.}
\label{tab:latency_cost}
\begin{center}
\small
\begin{tabular}{@{}lccc@{}}
\toprule
& \multicolumn{2}{c}{\textbf{\textsc{EmbedPlan}}} & \\
\cmidrule(lr){2-3}
\textbf{Metric} & \textbf{Full pipeline} & \textbf{Batched/cached} & \textbf{LLM API} \\
\midrule
Latency & 18.6~ms & 0.168~ms & 1{,}888.2~ms \\
Speedup vs.\ LLM & 101$\times$ & 11{,}215$\times$ & 1$\times$ \\
Compute per transition & $\sim$286M FLOPs$^\dagger$ & $\sim$286M FLOPs$^\dagger$ & 198 tokens \\
\bottomrule
\end{tabular}
\end{center}
\end{table}

\paragraph{Matched candidate-ranking comparison}
The generation baselines of Table~\ref{tab:llm_baselines} solve a harder task than
retrieval. To compare like with like, we give GPT-5.4 the identical ranking problem:
the Interpolation query $(s,a)$ together with the same 128 candidates \textsc{EmbedPlan}
ranks, from which it must select the successor. GPT-5.4 selects the correct successor for
$44\%$ of Ferry and $92\%$ of Logistics queries, against $99.0\%$ and $96.3\%$ Hit@1 for
\textsc{EmbedPlan} (Table~\ref{tab:app_pool_size}, with Logistics Hit@1 ranging over $91.2$--$96.3\%$ across our runs), at roughly four orders of magnitude lower
per-transition cost.

\paragraph{Training cost and break-even}
Training is a real tradeoff and is not amortized across domains, since cross-domain
transfer fails (Section~\ref{sec:results}). Per domain, \textsc{EmbedPlan} needs
$(s,a,s')$ transitions, a one-time embedding pass, and ${\sim}15$ minutes of training.
Against the measured $1{,}888$~ms autoregressive baseline it costs $0.168$~ms per
transition, so it breaks even after roughly $4{,}300$ transitions with Llama-3.3-70B as
the encoder, or $640$ with MPNet. The difference is the one-time embedding pass, which is
far costlier with a 70B encoder. A single search expands $10^3$ to $10^4$ successors, so
the method pays off for a domain that will be planned in repeatedly, not for one seen
once.

\paragraph{Carbon footprint}
Estimated total compute for training and evaluation is $\sim$200 GPU-hours on A100. Using a carbon intensity of 0.4~kg CO$_2$/kWh and an A100 TDP of 400W, the estimated emissions are $\sim$32~kg CO$_2$.

\FloatBarrier
\section{Dataset}
\label{app:dataset}

\subsection{Dataset statistics}
\label{app:dataset_stats}

We use nine classical PDDL domains from planning benchmarks~\citep{kokel2025acpbench}. Table~\ref{tab:dataset_stats} summarizes the dataset. Our transition datasets are derived from ACPBench domains but involve additional processing not included in the publicly released dataset. We will release our processed data upon acceptance.

\begin{table}[h]
\caption{Dataset statistics by domain.}
\label{tab:dataset_stats}
\begin{center}
\small
\begin{tabular}{@{}lrrrr@{}}
\toprule
\textbf{Domain} & \textbf{Problems} & \textbf{States} & \textbf{Transitions} & \textbf{Actions} \\
\midrule
Blocksworld & 5 & 43,551 & 43,065 & 4 \\
Depot & 7 & 5,795 & 13,256 & 5 \\
Ferry & 10 & 46,205 & 225,300 & 3 \\
Floortile & 6 & 33,608 & 166,565 & 6 \\
Goldminer & 7 & 12,237 & 52,023 & 7 \\
Grid & 5 & 8,671 & 664,346 & 5 \\
Logistics & 7 & 13,373 & 46,866 & 6 \\
Rovers & 10 & 46,783 & 509,457 & 9 \\
Satellite & 10 & 49,204 & 1,248,696 & 4 \\
\midrule
\textbf{Total} & \textbf{67} & \textbf{259,427} & \textbf{2,969,574} &  \\
\bottomrule
\end{tabular}
\end{center}
\end{table}

\subsection{State representation}
\label{app:state_repr}

States are rendered as natural language descriptions containing the current predicate values. An example from Blocksworld:

\begin{quote}
\small
\textit{``Block A is on the table. Block B is on Block A. Block C is clear. The robotic arm is empty.''}
\end{quote}

Actions are parameterized strings (e.g., \texttt{pick-up(BlockC)}, \texttt{stack(BlockA, BlockB)}). Note that the state count in Table~\ref{tab:dataset_stats} counts unique state \emph{descriptions}: the terminal state of each plan contributes a state but no transition, which is why some domains list more states than transitions.

\subsection{Domain descriptions}
\label{app:domains}

\paragraph{Blocksworld}
A robotic arm must rearrange colored blocks into a specified goal configuration. Only clear blocks (with nothing on top) can be moved. This domain is known for simple rules yet rich combinatorial complexity.

\paragraph{Depot}
Combines logistics and block stacking. Crates must be moved between depots using trucks for transportation and hoists for stacking and unstacking.

\paragraph{Ferry}
A ferry boat transports cars between locations. The ferry can carry only one car at a time, requiring optimization of loading and unloading sequences.

\paragraph{Floortile}
Robots paint the tiles of a grid-shaped floor according to a target pattern. A robot paints the tile above or below it, can switch between two colors, and cannot move onto a painted tile, so the painting order matters.

\paragraph{Goldminer}
A robot in a grid-shaped mine must reach and pick up the gold. A laser clears both hard and soft rock but destroys the gold if fired at it, whereas bombs clear only soft rock.

\paragraph{Grid}
An agent moves on a 2D grid to reach target locations, potentially with obstacles restricting movement.

\paragraph{Logistics}
Packages must be delivered within and across cities. Trucks handle intra-city transport, and airplanes handle inter-city deliveries.

\paragraph{Rovers}
Planetary exploration with multiple rovers collecting samples, taking images, and transmitting data. Rovers have specialized equipment and must communicate with a base station.

\paragraph{Satellite}
Satellites turn to point at targets and take images in given modes with onboard instruments, which must be switched on (one per satellite at a time) and calibrated first.

\subsection{Transition examples}
\label{app:examples}

Figures~\ref{fig:example_blocksworld}--\ref{fig:example_rovers} show representative transitions from four domains.

\begin{figure}[h]
\begin{center}
\begin{tcolorbox}[title=Blocksworld, left=2pt, right=2pt, fontupper=\small]
\textbf{State $s$:} Block block\_2 is on the table, Block block\_3 is located on the table, No blocks are placed on top of block\_1, The block block\_2 is currently situated under the block block\_1, The robotic arm is not holding anything, and Block block\_3 is clear. \\[4pt]
\textbf{Action $a$:} \texttt{(pick-up block\_3)} \\[4pt]
\textbf{Result $s'$:} Block block\_1 is clear, Block block\_2 is located on the table, The robotic arm is holding block\_3, and The block block\_2 is currently situated under the block block\_1.
\end{tcolorbox}
\end{center}
\caption{Representative transition from Blocksworld.}
\label{fig:example_blocksworld}
\end{figure}

\begin{figure}[h]
\centering
\begin{tcolorbox}[title=Ferry, left=2pt, right=2pt, fontupper=\small]
\textbf{State $s$:} Car c0 is at location l0, Car c2 is at location l1, The ferry is at l0, and Car c1 is on the ferry. \\[4pt]
\textbf{Action $a$:} \texttt{(debark c1 l0)} \\[4pt]
\textbf{Result $s'$:} Car c0 is at location l0, Car c1 is at location l0, The ferry is at l0, The ferry is empty, and Car c2 is at location l1.
\end{tcolorbox}
\caption{Representative transition from Ferry.}
\label{fig:example_ferry}
\end{figure}

\begin{figure}[h]
\centering
\begin{tcolorbox}[title=Logistics, left=2pt, right=2pt, fontupper=\small]
\textbf{State $s$:} p3 is in t1, a0 is at l0-0, t1 is at l1-0, p0 is at l1-0, p1 is in t1, t0 is at l0-0, and p2 is in a0. \\[4pt]
\textbf{Action $a$:} \texttt{(fly-airplane a0 l0-0 l1-0)} \\[4pt]
\textbf{Result $s'$:} t1 is at l1-0, p0 is at l1-0, p1 is in t1, p3 is in t1, t0 is at l0-0, a0 is at l1-0, and p2 is in a0.
\end{tcolorbox}
\caption{Representative transition from Logistics.}
\label{fig:example_logistics}
\end{figure}

\begin{figure}[h]
\begin{center}
\begin{tcolorbox}[title=Rovers, left=2pt, right=2pt, fontupper=\small]
\textbf{State $s$:} Store(s) store0 is empty, Channel general is free, Image objective1 was communicated in mode colour, Rover rover0 has image objective1 in mode colour, Rover rover1 has soil analyzed in waypoint waypoint0, Rover rover1 is available, Rover rover1 is at waypoint0, Rocks can be sampled at the following location(s): waypoint0, Store(s) store1 is empty, Rover rover0 is available, and Rover rover0 is at waypoint2. \\[4pt]
\textbf{Action $a$:} \texttt{(communicate\_soil\_data rover1 general waypoint0 waypoint0 waypoint1)} \\[4pt]
\textbf{Result $s'$:} Store(s) store0 is empty, Channel general is free, Rover rover0 has image objective1 in mode colour, Image objective1 was communicated in mode colour, Rover rover1 has soil analyzed in waypoint waypoint0, Rover rover1 is available, Rover rover1 is at waypoint0, Rocks can be sampled at the following location(s): waypoint0, Soil data was communicated from waypoint waypoint0, Store(s) store1 is empty, Rover rover0 is available, and Rover rover0 is at waypoint2.
\end{tcolorbox}
\end{center}
\caption{Representative transition from Rovers.}
\label{fig:example_rovers}
\end{figure}

\FloatBarrier
\section{Extended results}
\label{app:results}

\subsection{Per-domain results}
\label{app:per_domain}

Table~\ref{tab:per_domain_full} provides a complete per-domain breakdown across encoders and evaluation protocols.

\begin{table}[h]
\caption{Per-domain Hit@5 (\%) for the Interpolation and Extrapolation splits.}
\label{tab:per_domain_full}
\begin{center}
\small
\begin{tabular}{@{}lcccc@{}}
\toprule
& \multicolumn{2}{c}{\textbf{Qwen2.5-7B}} & \multicolumn{2}{c}{\textbf{Llama-3.3-70B}} \\
\cmidrule(lr){2-3} \cmidrule(lr){4-5}
\textbf{Domain} & \textbf{Interp.} & \textbf{Extrap.} & \textbf{Interp.} & \textbf{Extrap.} \\
\midrule
Blocksworld & 100.0 & 41.6{\scriptsize$\pm$8.7} & 100.0 & 49.1{\scriptsize$\pm$10} \\
Depot & 98.2 & 24.8{\scriptsize$\pm$5.2} & 98.8 & 25.9{\scriptsize$\pm$6} \\
Ferry & 99.9 & 36.7{\scriptsize$\pm$0.6} & 100.0 & 40.6{\scriptsize$\pm$3} \\
Floortile & 99.4 & 55.2{\scriptsize$\pm$11} & 99.6 & 68.8{\scriptsize$\pm$16} \\
Goldminer & 99.9 & 74.4{\scriptsize$\pm$5.8} & 100.0 & 76.2{\scriptsize$\pm$5} \\
Grid & 98.6 & 62.7{\scriptsize$\pm$5.2} & 99.8 & 74.9{\scriptsize$\pm$1} \\
Logistics & 99.6 & 44.7{\scriptsize$\pm$12} & 99.9 & 53.7{\scriptsize$\pm$10} \\
Rovers & 99.7 & 49.2{\scriptsize$\pm$1.7} & 99.0 & 54.4{\scriptsize$\pm$3} \\
Satellite & 99.9 & 40.0{\scriptsize$\pm$0.6} & 99.9 & 47.5{\scriptsize$\pm$6} \\
\midrule
\textbf{Mean} & 99.5 & 47.7{\scriptsize$\pm$4.9} & 99.7 & 54.6{\scriptsize$\pm$5.5} \\
\bottomrule
\end{tabular}
\end{center}
\end{table}

\subsection{Full performance metrics}
\label{app:full_metrics}

Table~\ref{tab:full_metrics} reports Hit@1/5/10 and action accuracy under the Extrapolation split.

\begin{table}[t]
\caption{Single-step transition prediction (Interpolation split, Llama-3.3-70B). \textit{Next state prediction}: whether the ground-truth next state is in the top-$k$ retrieved candidates. \textit{Action disambiguation}: whether, among all actions applied to $s$, the correct action $a$ produces the prediction closest to ground-truth $s'$. Mean Acc@5 is 85.3\%.}
\label{tab:single_step}
\centering
\small
\begin{tabular}{@{}lcccccc@{}}
\toprule
& \multicolumn{3}{c}{\textbf{Next state prediction (\%)}} & \multicolumn{3}{c}{\textbf{Action disambiguation (\%)}} \\
\cmidrule(lr){2-4} \cmidrule(lr){5-7}
\textbf{Domain} & \textbf{Hit@1} & \textbf{Hit@5} & \textbf{Hit@10} & \textbf{Acc@1} & \textbf{Acc@5} & \textbf{Acc@10} \\
\midrule
Blocksworld & $94.6 \pm 0.4$ & $100.0 \pm 0.0$ & $100.0 \pm 0.0$ & $25.9 \pm 0.9$ & $88.6 \pm 0.2$ & $99.5 \pm 0.1$ \\
Depot & $76.9 \pm 1.2$ & $98.8 \pm 0.1$ & $99.4 \pm 0.1$ & $29.8 \pm 2.3$ & $82.2 \pm 2.1$ & $93.3 \pm 1.4$ \\
Ferry & $98.5 \pm 0.2$ & $100.0 \pm 0.0$ & $100.0 \pm 0.0$ & $43.1 \pm 0.8$ & $96.8 \pm 0.1$ & $99.8 \pm 0.0$ \\
Floortile & $97.9 \pm 0.4$ & $99.6 \pm 0.0$ & $99.8 \pm 0.0$ & $37.1 \pm 2.8$ & $86.5 \pm 1.9$ & $97.3 \pm 0.8$ \\
Goldminer & $93.8 \pm 0.8$ & $100.0 \pm 0.0$ & $100.0 \pm 0.0$ & $17.0 \pm 0.5$ & $73.3 \pm 0.6$ & $97.4 \pm 0.2$ \\
Grid & $90.1 \pm 0.2$ & $99.8 \pm 0.1$ & $99.9 \pm 0.0$ & $28.6 \pm 0.2$ & $86.7 \pm 0.1$ & $96.4 \pm 0.3$ \\
Logistics & $93.4 \pm 0.6$ & $99.9 \pm 0.1$ & $99.9 \pm 0.0$ & $25.1 \pm 2.0$ & $81.3 \pm 3.8$ & $94.0 \pm 1.9$ \\
Rovers & $86.0 \pm 0.8$ & $99.0 \pm 0.1$ & $99.6 \pm 0.1$ & $26.1 \pm 1.5$ & $73.9 \pm 1.4$ & $86.3 \pm 0.7$ \\
Satellite & $97.6 \pm 0.2$ & $99.9 \pm 0.0$ & $100.0 \pm 0.0$ & $55.0 \pm 1.7$ & $98.5 \pm 0.3$ & $99.9 \pm 0.0$ \\
\midrule
\textbf{Mean} & $92.1 \pm 2.3$ & $99.7 \pm 0.1$ & $99.8 \pm 0.1$ & $32.0 \pm 3.8$ & $85.3 \pm 2.9$ & $96.0 \pm 1.4$ \\
\bottomrule
\end{tabular}
\end{table}

\begin{table}[t]
\caption{Full metrics (Llama-3.3-70B, Extrapolation split). Action accuracy measures whether the correct action is identified given $(s, s')$.}
\label{tab:full_metrics}
\begin{center}
\small
\begin{tabular}{@{}lcccccc@{}}
\toprule
& \multicolumn{3}{c}{\textbf{State prediction}} & \multicolumn{3}{c}{\textbf{Action accuracy}} \\
\cmidrule(lr){2-4} \cmidrule(lr){5-7}
\textbf{Domain} & \textbf{Hit@1} & \textbf{Hit@5} & \textbf{Hit@10} & \textbf{Acc@1} & \textbf{Acc@5} & \textbf{Acc@10} \\
\midrule
Blocksworld & 17.6{\scriptsize$\pm$5.2} & 49.1{\scriptsize$\pm$9.8} & 64.6{\scriptsize$\pm$7.5} & 0.7{\scriptsize$\pm$0.1} & 7.9{\scriptsize$\pm$1.2} & 24.0{\scriptsize$\pm$2.3} \\
Depot & 4.7{\scriptsize$\pm$1.2} & 25.9{\scriptsize$\pm$6.4} & 41.2{\scriptsize$\pm$8.1} & 0.7{\scriptsize$\pm$0.2} & 7.8{\scriptsize$\pm$1.7} & 21.8{\scriptsize$\pm$2.3} \\
Ferry & 12.0{\scriptsize$\pm$1.7} & 40.6{\scriptsize$\pm$3.5} & 58.1{\scriptsize$\pm$4.0} & 1.1{\scriptsize$\pm$0.4} & 10.3{\scriptsize$\pm$1.2} & 23.7{\scriptsize$\pm$2.3} \\
Floortile & 37.6{\scriptsize$\pm$13} & 68.8{\scriptsize$\pm$16} & 78.9{\scriptsize$\pm$13} & 3.2{\scriptsize$\pm$0.6} & 28.3{\scriptsize$\pm$8.7} & 52.3{\scriptsize$\pm$14} \\
Goldminer & 35.8{\scriptsize$\pm$6.4} & 76.2{\scriptsize$\pm$5.2} & 88.0{\scriptsize$\pm$1.7} & 3.0{\scriptsize$\pm$0.2} & 16.5{\scriptsize$\pm$1.2} & 45.7{\scriptsize$\pm$3.5} \\
Grid & 25.7{\scriptsize$\pm$1.2} & 74.9{\scriptsize$\pm$1.2} & 88.0{\scriptsize$\pm$1.2} & 8.0{\scriptsize$\pm$2.3} & 50.0{\scriptsize$\pm$4.6} & 73.9{\scriptsize$\pm$5.8} \\
Logistics & 16.5{\scriptsize$\pm$3.5} & 53.7{\scriptsize$\pm$9.8} & 70.8{\scriptsize$\pm$9.8} & 0.6{\scriptsize$\pm$0.3} & 8.4{\scriptsize$\pm$1.7} & 25.3{\scriptsize$\pm$1.7} \\
Rovers & 16.9{\scriptsize$\pm$1.2} & 54.4{\scriptsize$\pm$2.9} & 72.2{\scriptsize$\pm$2.3} & 0.9{\scriptsize$\pm$0.1} & 7.1{\scriptsize$\pm$0.2} & 16.5{\scriptsize$\pm$1.7} \\
Satellite & 14.4{\scriptsize$\pm$2.3} & 47.5{\scriptsize$\pm$5.8} & 66.9{\scriptsize$\pm$5.8} & 0.9{\scriptsize$\pm$0.2} & 9.1{\scriptsize$\pm$1.2} & 23.5{\scriptsize$\pm$4.0} \\
\midrule
\textbf{Mean} & 20.1{\scriptsize$\pm$3.6} & 54.6{\scriptsize$\pm$5.5} & 69.9{\scriptsize$\pm$4.9} & 2.1{\scriptsize$\pm$0.8} & 16.2{\scriptsize$\pm$4.8} & 34.1{\scriptsize$\pm$6.4} \\
\bottomrule
\end{tabular}
\end{center}
\end{table}

The gap between Hit@5 (54.6\%) and Acc@5 (16.2\%) reveals that models predict correct next states without fully capturing causal action structure.

\subsection{Cross-Domain transfer matrix}
\label{app:cross_domain}

Table~\ref{tab:cross_domain_matrix} shows the complete $9\times9$ transfer matrix.

\begin{table}[t]
\caption{Cross-Domain transfer (Llama-3.3-70B). Hit@5 (\%) training on the row domain and testing on the column domain. Chance: 3.9\% (5 of 128). Untrained model: 4.2\%.}
\label{tab:cross_domain_matrix}
\begin{center}
\small
\begin{tabular}{@{}lcccccccccc@{}}
\toprule
& \textbf{Block} & \textbf{Depot} & \textbf{Ferry} & \textbf{Floor} & \textbf{Gold} & \textbf{Grid} & \textbf{Logis} & \textbf{Rover} & \textbf{Satel} & \textbf{Mean} \\
\midrule
Blocksworld &  & 5.7 & 5.5 & 6.7 & 6.1 & 6.5 & 9.2 & 5.1 & 5.3 & 6.3 \\
Depot & 4.3 &  & 5.3 & 4.8 & 5.0 & 4.9 & 6.0 & 7.5 & 5.4 & 5.4 \\
Ferry & 5.3 & 8.7 &  & 9.3 & 5.2 & 9.0 & \textbf{22.3} & 6.6 & 5.6 & 9.0 \\
Floortile & 5.0 & 7.3 & 7.0 &  & 6.1 & 6.5 & 7.5 & 10.0 & 7.9 & 7.2 \\
Goldminer & 4.2 & 5.4 & 5.7 & 5.0 &  & 7.5 & 5.4 & 5.9 & 4.9 & 5.5 \\
Grid & 5.4 & 6.6 & 8.6 & 9.0 & 10.4 &  & 14.3 & 5.9 & 5.4 & 8.2 \\
Logistics & 4.3 & 5.5 & 7.7 & 4.9 & 7.0 & 6.3 &  & 5.1 & 4.5 & 5.7 \\
Rovers & 4.3 & 5.7 & 5.0 & 8.4 & 4.6 & 5.9 & 4.8 &  & 8.2 & 5.9 \\
Satellite & 4.5 & 7.5 & 6.3 & 7.8 & 6.4 & 5.2 & 10.1 & 5.9 &  & 6.7 \\
\midrule
\textbf{Mean} & 4.7 & 6.5 & 6.4 & 7.0 & 6.4 & 6.5 & 9.9 & 6.5 & 5.9 & \textbf{6.6} \\
\bottomrule
\end{tabular}
\end{center}
\end{table}

The only notable transfer is Ferry$\rightarrow$Logistics (22.3\%), which we attribute to shared transportation semantics in state descriptions.

\subsection{Leave-One-Out results}
\label{app:loo}

\begin{table}[h]
\caption{Leave-One-Out results (Llama-3.3-70B). Train on eight domains, test on the held-out domain.}
\label{tab:loo_results}
\begin{center}
\small
\begin{tabular}{@{}lccc@{}}
\toprule
\textbf{Held-out domain} & \textbf{Hit@1} & \textbf{Hit@5} & \textbf{Hit@10} \\
\midrule
Logistics & 3.3{\scriptsize$\pm$0.4} & 15.8{\scriptsize$\pm$2.2} & 27.6{\scriptsize$\pm$3.3} \\
Grid & 2.3{\scriptsize$\pm$0.2} & 12.3{\scriptsize$\pm$0.8} & 22.9{\scriptsize$\pm$1.4} \\
Rovers & 2.6{\scriptsize$\pm$0.3} & 12.6{\scriptsize$\pm$1.1} & 22.6{\scriptsize$\pm$1.6} \\
Ferry & 1.8{\scriptsize$\pm$0.1} & 9.0{\scriptsize$\pm$0.5} & 17.2{\scriptsize$\pm$0.8} \\
Satellite & 1.6{\scriptsize$\pm$0.1} & 8.4{\scriptsize$\pm$0.8} & 16.1{\scriptsize$\pm$1.5} \\
Floortile & 1.7{\scriptsize$\pm$0.3} & 7.9{\scriptsize$\pm$1.1} & 14.9{\scriptsize$\pm$2.0} \\
Goldminer & 1.3{\scriptsize$\pm$0.0} & 6.4{\scriptsize$\pm$0.2} & 12.5{\scriptsize$\pm$0.3} \\
Depot & 1.1{\scriptsize$\pm$0.1} & 5.6{\scriptsize$\pm$0.5} & 10.9{\scriptsize$\pm$0.9} \\
Blocksworld & 1.0{\scriptsize$\pm$0.0} & 5.2{\scriptsize$\pm$0.1} & 10.3{\scriptsize$\pm$0.3} \\
\midrule
\textbf{Mean} & 1.9{\scriptsize$\pm$0.3} & 9.2{\scriptsize$\pm$1.2} & 17.2{\scriptsize$\pm$2.0} \\
\bottomrule
\end{tabular}
\end{center}
\end{table}

Despite training on eight diverse domains, LOO performance (9.2\%) barely exceeds the untrained model (4.2\%).

\subsection{Multi-Domain results}
\label{app:multi_domain}

\begin{table}[h]
\caption{Multi-Domain unified model (Llama-3.3-70B, Extrapolation split).}
\label{tab:multi_domain}
\begin{center}
\small
\begin{tabular}{@{}lclc@{}}
\toprule
\textbf{Domain} & \textbf{Hit@5} & \textbf{Domain} & \textbf{Hit@5} \\
\midrule
Floortile & 52.0{\scriptsize$\pm$10} & Blocksworld & 36.9{\scriptsize$\pm$13} \\
Rovers & 51.9{\scriptsize$\pm$5} & Satellite & 34.3{\scriptsize$\pm$7} \\
Goldminer & 46.0{\scriptsize$\pm$8} & Grid & 32.6{\scriptsize$\pm$1} \\
Ferry & 37.7{\scriptsize$\pm$14} & Logistics & 24.8{\scriptsize$\pm$10} \\
& & Depot & 18.9{\scriptsize$\pm$12} \\
\midrule
\multicolumn{4}{c}{\textbf{Mean}: 37.2{\scriptsize$\pm$3.8} (vs.\ 54.6 single-domain)} \\
\bottomrule
\end{tabular}
\end{center}
\end{table}

\subsection{Untrained baseline}
\label{app:untrained}

To establish a performance floor, we evaluated the transition function with randomly initialized weights.

\begin{table}[h]
\caption{Untrained baseline (random weights, Llama-3.3-70B).}
\label{tab:untrained}
\begin{center}
\small
\begin{tabular}{@{}lccc@{}}
\toprule
\textbf{Domain} & \textbf{Hit@1} & \textbf{Hit@5} & \textbf{Hit@10} \\
\midrule
Blocksworld & 0.8{\scriptsize$\pm$0.1} & 4.0{\scriptsize$\pm$0.2} & 8.0{\scriptsize$\pm$0.4} \\
Depot & 0.9{\scriptsize$\pm$0.2} & 3.9{\scriptsize$\pm$0.0} & 8.0{\scriptsize$\pm$0.4} \\
Ferry & 0.7{\scriptsize$\pm$0.1} & 4.3{\scriptsize$\pm$0.6} & 8.4{\scriptsize$\pm$0.9} \\
Floortile & 0.8{\scriptsize$\pm$0.0} & 4.0{\scriptsize$\pm$0.1} & 8.0{\scriptsize$\pm$0.1} \\
Goldminer & 1.0{\scriptsize$\pm$0.4} & 4.8{\scriptsize$\pm$1.4} & 8.9{\scriptsize$\pm$2.0} \\
Grid & 0.9{\scriptsize$\pm$0.1} & 4.4{\scriptsize$\pm$0.3} & 8.0{\scriptsize$\pm$0.2} \\
Logistics & 0.9{\scriptsize$\pm$0.1} & 4.1{\scriptsize$\pm$0.2} & 8.6{\scriptsize$\pm$0.6} \\
Rovers & 0.7{\scriptsize$\pm$0.0} & 4.1{\scriptsize$\pm$0.2} & 8.4{\scriptsize$\pm$0.1} \\
Satellite & 0.7{\scriptsize$\pm$0.0} & 3.9{\scriptsize$\pm$0.1} & 7.8{\scriptsize$\pm$0.1} \\
\midrule
\textbf{Mean} & 0.8{\scriptsize$\pm$0.0} & 4.2{\scriptsize$\pm$0.1} & 8.2{\scriptsize$\pm$0.1} \\
\bottomrule
\end{tabular}
\end{center}
\end{table}

The untrained model (4.2\% Hit@5, close to the 3.9\% chance level of 5 in 128) confirms the retrieval task's inherent difficulty and that learned performance results from actual dynamics learning.

\subsection{Plan-level evaluation across encoders}
\label{app:plan_level}

We extend our evaluation to the trajectory level to assess whether high transition accuracy translates to reliable multi-step planning. We compare two splitting strategies:
\begin{itemize}
    \item \textbf{Plan-Variant}: test plans come from problem instances seen during training, but the specific plans are held out.
    \item \textbf{Extrapolation}: test plans come from entirely new problem instances never seen during training.
\end{itemize}

We report two metrics:
\begin{itemize}
    \item \textbf{Mean trajectory Hit@5}: the average Hit@5 across all steps in a trajectory.
    \item \textbf{Exact trajectory Hit@5}: the percentage of trajectories in which \textit{every} step is correctly retrieved.
\end{itemize}

Table~\ref{tab:plan_eval_llama} reports the full results for Llama-3.3-70B, and Table~\ref{tab:plan_eval_encoders} compares the three encoders. We observe a consistent trajectory generalization gap across all models. Under the Plan-Variant split, models achieve moderate reliability, though exact-match rates remain low due to error accumulation. Under Extrapolation, performance collapses. For Blocksworld, the exact trajectory rate falls to 0.0\% across all encoders. This indicates that the transition model's planning capability is largely confined to seen problem manifolds. When forced to extrapolate to new problems, the probability of executing a valid multi-step plan drops to near zero.

\begin{table}[t]
\caption{Plan-level evaluation (Llama-3.3-70B). Trajectory metrics under the Plan-Variant and Extrapolation splits.}
\label{tab:plan_eval_llama}
\begin{center}
\small
\begin{tabular}{@{}lcccc@{}}
\toprule
& \multicolumn{2}{c}{\textbf{Plan-Variant}} & \multicolumn{2}{c}{\textbf{Extrapolation}} \\
\cmidrule(lr){2-3} \cmidrule(lr){4-5}
\textbf{Domain} & \textbf{Mean Hit@5} & \textbf{Exact Hit@5} & \textbf{Mean Hit@5} & \textbf{Exact Hit@5} \\
\midrule
Blocksworld & $38.7 \pm 0.8$ & $1.6 \pm 0.2$ & $3.7 \pm 1.3$ & $0.0 \pm 0.0$ \\
Depot & $30.1 \pm 2.6$ & $11.2 \pm 1.5$ & $2.4 \pm 1.7$ & $1.0 \pm 1.0$ \\
Ferry & $71.1 \pm 1.1$ & $39.1 \pm 1.6$ & $2.0 \pm 0.7$ & $0.5 \pm 0.2$ \\
Floortile & $74.1 \pm 2.7$ & $33.6 \pm 3.3$ & $11.6 \pm 4.4$ & $2.8 \pm 1.2$ \\
Goldminer & $55.4 \pm 2.2$ & $23.8 \pm 1.6$ & $18.5 \pm 6.2$ & $5.5 \pm 1.8$ \\
Grid & $55.0 \pm 2.4$ & $37.4 \pm 2.2$ & $16.6 \pm 3.7$ & $8.2 \pm 2.1$ \\
Logistics & $52.8 \pm 0.6$ & $14.1 \pm 2.0$ & $22.7 \pm 8.6$ & $9.6 \pm 4.4$ \\
Rovers & $26.5 \pm 0.8$ & $10.0 \pm 0.7$ & $12.2 \pm 3.5$ & $1.9 \pm 0.8$ \\
Satellite & $56.8 \pm 1.2$ & $32.2 \pm 1.0$ & $1.2 \pm 0.4$ & $0.5 \pm 0.1$ \\
\midrule
\textbf{Mean} & $51.2 \pm 5.5$ & $22.6 \pm 4.6$ & $10.1 \pm 2.7$ & $3.3 \pm 1.2$ \\
\bottomrule
\end{tabular}
\end{center}
\end{table}

\begin{table}[t]
\caption{Plan-level evaluation across encoders. \emph{Top}: Plan-Variant trajectory metrics for BGE-M3 and MPNet (Llama-3.3-70B in Table~\ref{tab:plan_eval_llama}). \emph{Bottom}: Extrapolation, mean $\pm$ SE over the six domains with an independent run for every encoder (Blocksworld, Depot, Ferry, Floortile, Goldminer, Grid).}
\label{tab:plan_eval_encoders}
\begin{center}
\small
\begin{tabular}{@{}lcccc@{}}
\toprule
& \multicolumn{2}{c}{\textbf{BGE-M3}} & \multicolumn{2}{c}{\textbf{MPNet}} \\
\cmidrule(lr){2-3} \cmidrule(lr){4-5}
\textbf{Plan-Variant} & \textbf{Mean Hit@5} & \textbf{Exact Hit@5} & \textbf{Mean Hit@5} & \textbf{Exact Hit@5} \\
\midrule
Blocksworld & $23.7 \pm 1.5$ & $0.2 \pm 0.1$ & $16.7 \pm 0.3$ & $0.3 \pm 0.1$ \\
Depot & $45.5 \pm 1.0$ & $20.9 \pm 0.1$ & $31.1 \pm 1.6$ & $10.8 \pm 1.1$ \\
Ferry & $63.3 \pm 1.4$ & $27.8 \pm 2.1$ & $60.9 \pm 2.0$ & $29.1 \pm 2.3$ \\
Floortile & $49.2 \pm 1.0$ & $13.1 \pm 1.1$ & $9.2 \pm 0.5$ & $3.0 \pm 0.4$ \\
Goldminer & $67.9 \pm 2.6$ & $35.0 \pm 3.2$ & $26.8 \pm 0.8$ & $10.5 \pm 0.5$ \\
Grid & $51.9 \pm 1.5$ & $34.0 \pm 1.2$ & $32.7 \pm 2.7$ & $17.5 \pm 2.6$ \\
Logistics & $67.2 \pm 0.3$ & $25.6 \pm 1.6$ & $41.3 \pm 2.2$ & $9.9 \pm 1.9$ \\
Rovers & $60.5 \pm 0.2$ & $24.7 \pm 0.2$ & $0.1 \pm 0.0$ & $0.0 \pm 0.0$ \\
Satellite & $51.2 \pm 0.4$ & $27.5 \pm 0.6$ & $0.2 \pm 0.1$ & $0.1 \pm 0.0$ \\
\midrule
\textbf{Mean} & $53.4 \pm 4.6$ & $23.2 \pm 3.6$ & $24.3 \pm 6.7$ & $9.0 \pm 3.2$ \\
\bottomrule
\end{tabular}

\vspace{6pt}
\begin{tabular}{@{}lcc@{}}
\toprule
\textbf{Extrapolation (six domains)} & \textbf{Mean Hit@5} & \textbf{Exact Hit@5} \\
\midrule
Llama-3.3-70B & $9.1 \pm 3.0$ & $3.0 \pm 1.3$ \\
BGE-M3        & $8.2 \pm 3.4$ & $3.4 \pm 1.6$ \\
MPNet         & $4.6 \pm 2.7$ & $2.0 \pm 1.2$ \\
\bottomrule
\end{tabular}
\end{center}
\end{table}

\paragraph{Key observations}
\begin{itemize}
    \item \textbf{Plan-Variant performance is comparable for the two larger encoders}: Llama-3.3-70B and BGE-M3 achieve similar mean Hit@5 ($\sim$51--53\%), while MPNet lags substantially (24.3\%).
    \item \textbf{Extrapolation collapse is universal}: all encoders degrade sharply under Extrapolation, with exact Hit@5 at most $3.4\%$ on average.
    \item \textbf{Domain-specific patterns persist}: Goldminer and Grid extrapolate best (Table~\ref{tab:plan_eval_llama}), while Blocksworld fails for every encoder.
    \item \textbf{MPNet struggles even under Plan-Variant}: Rovers and Satellite show near-zero performance, suggesting these domains require higher-capacity embeddings.
\end{itemize}

\subsection{Closed-loop rollout}
\label{app:closed_loop}

The plan-level results above are teacher-forced: the true state is supplied at every
step, so errors cannot accumulate. Here we remove that correction. All three regimes
predict the next state from the current state and action and differ only in what is fed to the next step.
\begin{itemize}
    \item \textbf{Teacher-forced}: the true state $s_t$ (the setting of Appendix~\ref{app:plan_level}).
    \item \textbf{Closed loop, with snapping}: the model's own prediction, snapped to the
    candidate pool (its top-1 retrieved state) and fed back as the next input. If
    retrieval picks a wrong state, that wrong state is propagated, so errors compound.
    \item \textbf{Closed loop, without snapping}: the model's own prediction as a raw
    latent, never snapped. No candidate pool is consulted at any intermediate step.
\end{itemize}
The split is Interpolation and the retrieval pool has 1{,}000 candidates. Aggregate
step-wise Hit@1 appears in Table~\ref{tab:closed_loop}: with snapping, the rollout
retains $99\%$ (Ferry), $92\%$ (Logistics), and $98\%$ (Blocksworld) of teacher-forced
accuracy.

\begin{table}[h]
\caption{Closed-loop rollout by depth on Ferry. Step-wise Hit@1 in \% (mean $\pm$ SE), where $n$ is the number of test trajectories reaching that depth. Same models, trajectories, and pool as Table~\ref{tab:closed_loop}, and the depth rows weighted by $n$ reproduce the aggregate to
rounding.}
\label{tab:closed_loop_depth}
\centering\small
\begin{tabular}{@{}rrccc@{}}
\toprule
\textbf{Depth} & $\boldsymbol{n}$ & \textbf{True state at each step} & \textbf{Snapped} & \textbf{Not snapped} \\
\midrule
1 & 300 & $94.3 \pm 0.7$ & $94.3 \pm 0.7$ & $94.3 \pm 0.7$ \\
2 & 300 & $96.3 \pm 0.5$ & $94.4 \pm 0.7$ & $52.4 \pm 3.4$ \\
3 &  57 & $94.2 \pm 1.9$ & $92.4 \pm 0.5$ & $20.9 \pm 2.4$ \\
4 &   7 & $96.7 \pm 4.7$ & $93.3 \pm 9.4$ & $10.3 \pm 7.4$ \\
\bottomrule
\end{tabular}
\end{table}

\paragraph{Depth isolates the mechanism}
Table~\ref{tab:closed_loop_depth} breaks the Ferry rollout down by depth. The three regimes coincide at depth~1 by construction, before any feedback exists. From
depth~2 they separate: snapped rollout stays flat, while unsnapped rollout collapses
($52.4\%$, $20.9\%$, $10.3\%$). The collapse tracks a single measurable quantity, the drift
of the predicted latent off the manifold of real state embeddings: its cosine similarity
to the nearest real state is $1.00$ at the input, $0.83$ after one step, and ${\sim}0.55$
once unrolled further. Weighted over depths~2--4, snapped rollout retains $98\%$ of teacher-forced Hit@1
($94.1\%$ vs.\ $96.0\%$) and unsnapped rollout $49\%$ ($46.7\%$). Per-step snapping to
that manifold is therefore what keeps multi-step rollout accurate. Rollout without snapping is limited by transition-function fidelity, which is where further gains should
be sought.

\paragraph{Horizon}
Test plans in this benchmark are short: of the $300$ Ferry trajectories, $243$ have two
steps and $7$ reach four (mean $2.2$). Depth~3 ($n=57$) is the deepest row on which we
rest a claim, and there snapped rollout holds at $92.4\%$ against $20.9\%$ without
snapping. What is demonstrated is that snapping sustains rollout over these short test
trajectories, not over long horizons. Combining the snapped transition function with
beam or tree search over longer horizons is the natural next step.

\FloatBarrier
\section{Statistical analysis}
\label{app:statistics}

\subsection{Main generalization gap}
\label{app:gap_stats}

Table~\ref{tab:gap_stats} reports the interpolation-extrapolation gap.

\begin{table}[h]
\caption{Statistical analysis of the generalization gap (Qwen2.5-7B, paired across the nine domains).}
\label{tab:gap_stats}
\begin{center}
\small
\begin{tabular}{@{}ll@{}}
\toprule
\textbf{Statistic} & \textbf{Value} \\
\midrule
Interpolation (mean $\pm$ SE) & 99.5\% $\pm$ 0.2\% \\
Extrapolation (mean $\pm$ SE) & 47.7\% $\pm$ 4.9\% \\
Gap & 51.8~pp \\
Paired $t$-test & $t(8) = 10.58$ \\
$p$-value & $5.57 \times 10^{-6}$ \\
95\% CI & $[40.7, 62.9]$~pp \\
\bottomrule
\end{tabular}
\end{center}
\end{table}

\subsection{Effect sizes for key comparisons}
\label{app:effect_sizes}

\begin{table}[h]
\caption{Effect sizes for major findings. The encoder column is given explicitly because the comparisons are drawn from different encoder runs.}
\label{tab:effect_sizes}
\begin{center}
\small
\begin{tabular}{@{}llccc@{}}
\toprule
\textbf{Comparison} & \textbf{Encoder} & \textbf{$\Delta$} & \textbf{Cohen's $d$} & \textbf{$p$} \\
\midrule
Interpolation vs.\ Extrapolation & Qwen2.5-7B & $-$51.8~pp & 5.25 & $<10^{-5}$ \\
Extrapolation vs.\ Cross-Domain & Llama-3.3-70B & $-$48.0~pp & 4.32 & $<10^{-5}$ \\
Llama vs.\ MPNet (Extrap.) & both & $+$27.8~pp & 2.53 & $<0.001$ \\
Single vs.\ Multi (Extrap.) & Llama-3.3-70B & $+$17.4~pp & 1.42 & 0.028 \\
\bottomrule
\end{tabular}
\end{center}
\end{table}

\subsection{Per-domain statistical tests}
\label{app:per_domain_stats}

Table~\ref{tab:per_domain_stats} reports per-domain tests.

\begin{table}[h]
\caption{Independent $t$-tests comparing interpolation and extrapolation per domain (Qwen2.5-7B).}
\label{tab:per_domain_stats}
\begin{center}
\small
\begin{tabular}{@{}lccc@{}}
\toprule
\textbf{Domain} & \textbf{Gap (pp)} & \textbf{$t$-statistic} & \textbf{$p$-value} \\
\midrule
Depot & 73.4 & 12.87 & $1.17 \times 10^{-4}$ \\
Ferry & 63.3 & 65.49 & $3.24 \times 10^{-6}$ \\
Satellite & 59.9 & 153.39 & $4.11 \times 10^{-8}$ \\
Blocksworld & 58.4 & 5.95 & $2.70 \times 10^{-3}$ \\
Logistics & 55.0 & 4.28 & $8.28 \times 10^{-3}$ \\
Rovers & 50.5 & 20.22 & $8.90 \times 10^{-6}$ \\
Floortile & 44.2 & 3.68 & $1.57 \times 10^{-2}$ \\
Grid & 35.9 & 5.38 & $2.40 \times 10^{-3}$ \\
Goldminer & 25.5 & 3.54 & $1.13 \times 10^{-2}$ \\
\bottomrule
\end{tabular}
\end{center}
\end{table}

All comparisons remain significant after Bonferroni correction ($\alpha = 0.05/9 = 0.0056$) except Logistics, Floortile, and Goldminer, which are significant only at the uncorrected $\alpha = 0.05$.

\FloatBarrier
\section{Embedding analysis}
\label{app:embedding}

\subsection{PCA visualization}
\label{app:pca}

\begin{figure}[h]
\begin{center}
    \begin{minipage}{0.48\textwidth}
        \centering
        \includegraphics[width=\linewidth]{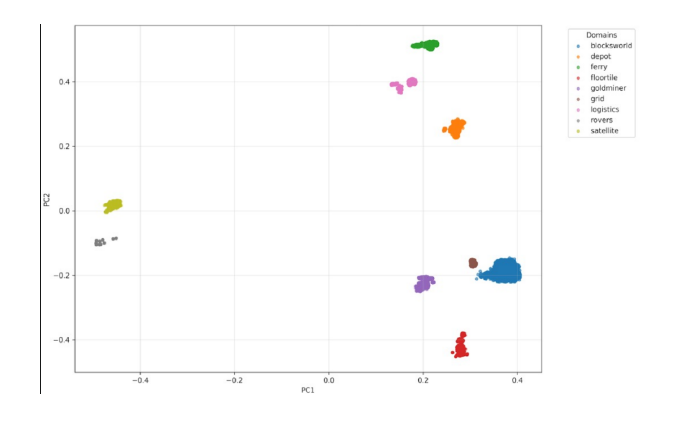}
        \centerline{\small (A) MPNet (110M)}
    \end{minipage}
    \hfill
    \begin{minipage}{0.48\textwidth}
        \centering
        \includegraphics[width=\linewidth]{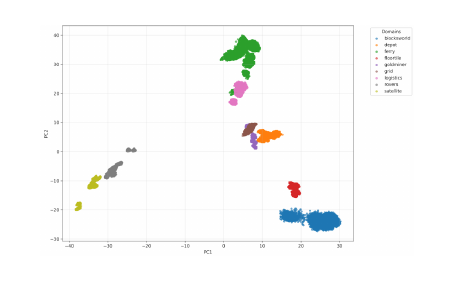}
        \centerline{\small (B) Llama-3.3-70B}
    \end{minipage}
\end{center}
\caption{Embedding space fragmentation across scales. PCA of frozen state embeddings, colored by domain. \textbf{(A)} MPNet places each domain in its own tight, isolated cluster. \textbf{(B)} Llama-3.3-70B, despite being roughly 640$\times$ larger, shows the same fragmentation. Pre-trained embeddings thus group states by domain-specific surface features rather than by abstract planning roles, regardless of model scale.}
\label{fig:pca_comparison}
\end{figure}

Figure~\ref{fig:pca_comparison} compares embedding geometry across model scales.

Both MPNet and Llama place each domain in an isolated cluster, often split further into sub-clusters. While Llama shows slightly more spread within clusters, the critical structural limitation remains: the manifolds of different domains are disjoint. Scaling up model parameters does not automatically induce abstract, problem-invariant representations.

\subsection{Domain complexity analysis}
\label{app:complexity}

\begin{table}[h]
\caption{Domain complexity metrics and generalization gaps (Qwen2.5-7B). Action counts are action schemas as defined in the PDDL domain file, matching Table~\ref{tab:dataset_stats}.}
\label{tab:complexity}
\begin{center}
\small
\begin{tabular}{@{}lccc@{}}
\toprule
\textbf{Domain} & \textbf{Actions} & \textbf{Predicates} & \textbf{Gap} \\
\midrule
Depot & 5 & 8 & 73.4 \\
Logistics & 6 & 6 & 55.0 \\
Satellite & 4 & 8 & 59.9 \\
Blocksworld & 4 & 5 & 58.4 \\
Ferry & 3 & 5 & 63.3 \\
Rovers & 9 & 26 & 50.5 \\
Floortile & 6 & 10 & 44.2 \\
Grid & 5 & 9 & 35.9 \\
Goldminer & 7 & 7 & 25.5 \\
\bottomrule
\end{tabular}
\end{center}
\end{table}

We find no clean relationship between the surface complexity of a domain and its generalization gap. Rovers has by far the largest action and predicate counts yet a mid-range gap (50.5~pp), while Ferry is among the simplest domains and shows one of the largest gaps (63.3~pp). This is consistent with our main claim: what limits extrapolation is how the embedding space organizes states, not how large or intricate the domain is.

\FloatBarrier
\section{Ablations}
\label{app:ablations}

\subsection{Architecture and encoder comparison}
\label{app:arch_encoder_compare}

\begin{table}[h]
\caption{Architecture comparison for the two largest encoders (Extrapolation split, Hit@5 \%).}
\label{tab:arch_encoder_compare}
\begin{center}
\small
\begin{tabular}{@{}lccc@{}}
\toprule
\textbf{Encoder} & \textbf{Parameters} & \textbf{MLP} & \textbf{Hypernetwork} \\
\midrule
Llama-3.3-70B & 70B  & 54.6$\pm$5.5 & 53.8$\pm$5.5 \\
Qwen2.5-7B    & 7B   & 47.7$\pm$4.9 & 46.2$\pm$5.0 \\
\bottomrule
\end{tabular}
\end{center}
\end{table}

Architecture choice has minimal impact on performance: the residual MLP matches or slightly exceeds the hypernetwork for both encoders, by at most $1.5$~pp. The generalization gap is consistent across architectures, which points to the embedding rather than the transition network design as the limitation.

Larger encoders yield better extrapolation performance, but the improvement is sublinear: a roughly 640$\times$ increase in parameters (MPNet to Llama) yields only a 2.0$\times$ improvement in Hit@5 (Table~\ref{tab:encoder}). This suggests that scale alone does not resolve the underlying structural limitation.

\subsection{Effect of the action disambiguation loss}
\label{app:ablation_action_loss}

We ablate the contribution of the action disambiguation loss $\mathcal{L}_{\text{action}}$ by comparing models trained with the full composite objective ($\lambda = 2$) against models trained with the state prediction loss only ($\lambda = 0$).

\begin{table}[h]
\caption{Ablation of the action disambiguation loss. Comparison of models trained without ($\lambda{=}0$) and with ($\lambda{=}2$) action disambiguation under Extrapolation evaluation (Llama-3.3-70B). The action loss yields gains in both state prediction ($+$19.3~pp) and action accuracy ($3.4\times$).}
\label{tab:ablation_action_loss}
\begin{center}
\small
\begin{tabular}{@{}lcccc@{}}
\toprule
& \multicolumn{2}{c}{\textbf{Hit@5 (\%)}} & \multicolumn{2}{c}{\textbf{Action Acc@5 (\%)}} \\
\cmidrule(lr){2-3} \cmidrule(lr){4-5}
\textbf{Domain} & $\lambda{=}0$ & $\lambda{=}2$ & $\lambda{=}0$ & $\lambda{=}2$ \\
\midrule
Blocksworld & 30.2{\scriptsize$\pm$8} & 49.1{\scriptsize$\pm$10} & 1.8{\scriptsize$\pm$0.4} & 7.9{\scriptsize$\pm$2} \\
Depot & 13.4{\scriptsize$\pm$4} & 25.9{\scriptsize$\pm$6} & 1.6{\scriptsize$\pm$0.3} & 7.8{\scriptsize$\pm$3} \\
Ferry & 24.8{\scriptsize$\pm$3} & 40.6{\scriptsize$\pm$3} & 2.5{\scriptsize$\pm$0.5} & 10.3{\scriptsize$\pm$2} \\
Floortile & 45.3{\scriptsize$\pm$14} & 68.8{\scriptsize$\pm$16} & 8.2{\scriptsize$\pm$3} & 28.3{\scriptsize$\pm$15} \\
Goldminer & 55.7{\scriptsize$\pm$6} & 76.2{\scriptsize$\pm$5} & 4.9{\scriptsize$\pm$1.2} & 16.5{\scriptsize$\pm$2} \\
Grid & 52.1{\scriptsize$\pm$2} & 74.9{\scriptsize$\pm$1} & 18.7{\scriptsize$\pm$5} & 50.0{\scriptsize$\pm$8} \\
Logistics & 31.5{\scriptsize$\pm$9} & 53.7{\scriptsize$\pm$10} & 1.9{\scriptsize$\pm$0.6} & 8.4{\scriptsize$\pm$3} \\
Rovers & 35.2{\scriptsize$\pm$4} & 54.4{\scriptsize$\pm$3} & 1.7{\scriptsize$\pm$0.3} & 7.1{\scriptsize$\pm$0.4} \\
Satellite & 29.8{\scriptsize$\pm$5} & 47.5{\scriptsize$\pm$6} & 2.2{\scriptsize$\pm$0.5} & 9.1{\scriptsize$\pm$2} \\
\midrule
\textbf{Mean} & 35.3{\scriptsize$\pm$4.5} & 54.6{\scriptsize$\pm$5.5} & 4.8{\scriptsize$\pm$1.9} & 16.2{\scriptsize$\pm$4.8} \\
\midrule
\textbf{$\Delta$} & \multicolumn{2}{c}{$+$19.3~pp ($+$55\%)} & \multicolumn{2}{c}{$+$11.4~pp ($3.4\times$)} \\
\bottomrule
\end{tabular}
\end{center}
\end{table}

The action disambiguation loss yields improvements across both metrics and all nine domains. The direct target, Action Acc@5, shows the largest gain, improving $3.4\times$ from 4.8\% to 16.2\%. Without explicit supervision on action effects, models fail to distinguish between actions with similar but distinct consequences: Action Acc@5 under $\lambda = 0$ barely exceeds the untrained baseline in most domains, indicating that the state prediction loss alone provides little signal for learning action semantics.

Hit@5 also improves by 19.3~pp ($+$55\% relative), despite $\mathcal{L}_{\text{action}}$ not directly optimizing this metric. We suggest two mechanisms for this indirect benefit: (1) without action-contrastive training, the model may exploit spurious correlations between surface-level state-action features and outcomes, which fail to transfer to unseen problems, and (2) the action loss pushes the network to encode transformation patterns that separate an action's effect from its object binding, for example distinguishing \texttt{pick-up(A)} from \texttt{pick-up(B)}.

The improvement is consistent across domains but varies in magnitude. Grid shows the largest absolute gain in Action Acc@5 ($+$31.3~pp), likely because its spatial action semantics (movement in cardinal directions) are highly distinctive when explicitly supervised. Depot and Logistics show the largest relative Hit@5 gains ($+$93\% and $+$70\%), suggesting that domains with complex multi-object interactions benefit most from learning precise action effects.

We set $\lambda = 2$ to emphasize action disambiguation, reflecting that distinguishing among $K$ domain actions applied to the same state requires finer-grained representations than distinguishing among $B$ random states in a batch. Without it, \textsc{EmbedPlan}'s extrapolation performance drops by roughly one-third and action accuracy is nearly absent. The full ablation across $\lambda \in \{0, 0.5, 1, 1.5, 2, 4\}$ confirms $\lambda = 2$ as the best setting, and performance degrades slightly at $\lambda = 4$ due to over-emphasis on action discrimination at the expense of state prediction.

\FloatBarrier
\section{Preliminary studies}
\label{app:preliminary}

\subsection{Latent distance alignment}
\label{app:lda}

Before learning transition functions, we tested whether pre-trained embeddings already encode planning-relevant structure. If embedding geometry reflects plan costs, one could use simple distance-based heuristics for search without any additional learning.

\paragraph{Hypothesis}
We define \emph{latent distance alignment} (LDA) as the property that embedding distance between a state and a goal correlates with the number of actions required to reach the goal:
\begin{equation}
\mathrm{corr}\big(d(\mathbf{e}_s, \mathbf{e}_g), \; \mathrm{cost}(s, g)\big) > 0,
\end{equation}
where $\mathbf{e}_s$ and $\mathbf{e}_g$ are the embeddings of state $s$ and goal $g$, $d(\cdot, \cdot)$ is cosine distance, and $\mathrm{cost}(s, g)$ is the optimal plan length from $s$ to $g$.

\paragraph{Motivation}
This hypothesis draws from successes in other domains. CLIP embeddings align images and text such that semantic similarity corresponds to embedding proximity~\citep{radford2021clip}. Sentence embeddings place entailed sentences closer than contradictions~\citep{reimers2019sbert}. We test whether similar alignment emerges for planning cost.

\paragraph{Method}
For 21,003 state-goal pairs across the nine domains, we computed embedding distances using all four encoders and correlated them with ground-truth plan costs from A$^*$ search.

\paragraph{Result}
After controlling for prompt length as a confound, correlations collapsed to near zero across all encoders. Pre-trained embeddings do not encode planning cost through geometric distance.

\paragraph{Implication}
This negative result motivated our transition learning approach: rather than relying on inherent geometry, we explicitly learn how actions transform states in embedding space.

\FloatBarrier
\section{Reproducibility}
\label{app:reproducibility}

\subsection{Code and data availability}
\label{app:code}

Code is available at \url{https://github.com/embedplan/EmbedPlan}, and the processed transition datasets will be released upon acceptance. All base encoders (Llama-3.3-70B, Qwen2.5-7B, MPNet, BGE-M3) are publicly available.

\subsection{Experimental reproducibility}

\begin{itemize}
    \item \textbf{Random seeds}: all experiments run with seeds $\{42, 123, 456\}$. Throughout the paper, $\pm$ denotes the standard error of the mean, over seeds for per-domain entries and over the nine domains for cross-domain means.
    \item \textbf{Hardware}: NVIDIA A100 80\,GB GPU.
    \item \textbf{Software}: PyTorch 2.1, CUDA 12.1, Python 3.10.
\end{itemize}

\FloatBarrier
\section{Error analysis}
\label{app:error_analysis}

We qualitatively analyzed Hit@5 errors under Extrapolation evaluation to characterize failure modes. We sampled 50 incorrect predictions per domain, where the ground-truth next state ranked outside the top 5, and manually inspected the retrieved candidates.

\paragraph{Methodology}
For each error, we compared the top-1 retrieved state against the ground-truth next state, counting the number of differing predicates.

\paragraph{Findings}
Across domains, the median predicate difference between the top-1 retrieved state and the ground truth was 2 (IQR: 1--3). Table~\ref{tab:error_examples} shows representative examples, with the differing predicates set in italics.

\begin{table}[h]
\caption{Representative Hit@5 errors. Retrieved states differ from the ground truth by one or two predicates (in italics), typically involving object locations or holdings within the same problem instance.}
\label{tab:error_examples}
\begin{center}
\small
\begin{tabular}{@{}p{1.6cm}p{4.2cm}p{4.2cm}@{}}
\toprule
\textbf{Domain} & \textbf{Ground truth $s'$} & \textbf{Retrieved $\hat{s}'$} \\
\midrule
Blocksworld & Block A is clear, \textit{arm holds B}, \textit{C is on table} & Block A is clear, \textit{arm holds C}, \textit{B is on table} \\
\addlinespace
Ferry & Car c1 at l0, ferry empty, \textit{c2 at l1} & Car c1 at l0, ferry empty, \textit{c2 at l0} \\
\addlinespace
Logistics & \textit{Package p1 in truck t0}, t0 at l1-0 & \textit{Package p1 at l1-0}, t0 at l1-0 \\
\bottomrule
\end{tabular}
\end{center}
\end{table}

These errors suggest the model captures coarse transition structure (correct problem context, approximate state region) but struggles to resolve fine-grained predicate changes, particularly when multiple objects undergo similar transformations.

\FloatBarrier
\section{Complete results tables}
\label{app:complete_tables}

Tables~\ref{tab:complete_qwen} and~\ref{tab:complete_llama} report complete metrics for the two largest encoders.

\begin{table}[h]
\caption{Complete metrics for Qwen2.5-7B across all domains and splits.}
\label{tab:complete_qwen}
\begin{center}
\small
\begin{tabular}{@{}llccc@{}}
\toprule
\textbf{Domain} & \textbf{Split} & \textbf{Hit@1} & \textbf{Hit@5} & \textbf{Hit@10} \\
\midrule
\multirow{2}{*}{Blocksworld} & Interpolation & 94.3$\pm$0.2 & 100.0$\pm$0.0 & 100.0$\pm$0.0 \\
 & Extrapolation & 14.2$\pm$4.0 & 41.6$\pm$8.8 & 56.6$\pm$8.3 \\
\midrule
\multirow{2}{*}{Depot} & Interpolation & 76.7$\pm$1.0 & 98.2$\pm$0.1 & 99.0$\pm$0.0 \\
 & Extrapolation & 4.9$\pm$1.2 & 24.8$\pm$4.9 & 42.2$\pm$7.8 \\
\midrule
\multirow{2}{*}{Ferry} & Interpolation & 98.7$\pm$0.1 & 99.9$\pm$0.0 & 100.0$\pm$0.0 \\
 & Extrapolation & 10.7$\pm$0.8 & 36.7$\pm$0.7 & 52.8$\pm$1.1 \\
\midrule
\multirow{2}{*}{Floortile} & Interpolation & 96.2$\pm$0.3 & 99.4$\pm$0.0 & 99.6$\pm$0.0 \\
 & Extrapolation & 24.0$\pm$5.7 & 55.2$\pm$11.0 & 69.0$\pm$11.5 \\
\midrule
\multirow{2}{*}{Goldminer} & Interpolation & 94.0$\pm$0.6 & 99.9$\pm$0.0 & 100.0$\pm$0.0 \\
 & Extrapolation & 34.5$\pm$6.8 & 74.4$\pm$5.7 & 86.7$\pm$2.2 \\
\midrule
\multirow{2}{*}{Grid} & Interpolation & 80.9$\pm$1.2 & 98.6$\pm$0.1 & 99.6$\pm$0.1 \\
 & Extrapolation & 19.4$\pm$3.2 & 62.7$\pm$5.2 & 76.9$\pm$5.7 \\
\midrule
\multirow{2}{*}{Logistics} & Interpolation & 88.9$\pm$1.1 & 99.6$\pm$0.1 & 99.8$\pm$0.0 \\
 & Extrapolation & 14.9$\pm$4.6 & 44.7$\pm$11.3 & 60.7$\pm$12.3 \\
\midrule
\multirow{2}{*}{Rovers} & Interpolation & 97.4$\pm$0.1 & 99.7$\pm$0.0 & 99.8$\pm$0.0 \\
 & Extrapolation & 18.5$\pm$1.0 & 49.2$\pm$1.8 & 65.0$\pm$1.8 \\
\midrule
\multirow{2}{*}{Satellite} & Interpolation & 98.1$\pm$0.1 & 99.9$\pm$0.0 & 100.0$\pm$0.0 \\
 & Extrapolation & 14.3$\pm$0.4 & 40.0$\pm$0.6 & 54.8$\pm$1.0 \\
\bottomrule
\end{tabular}
\end{center}
\end{table}

\begin{table}[h]
\caption{Complete metrics for Llama-3.3-70B across all domains and splits.}
\label{tab:complete_llama}
\begin{center}
\small
\begin{tabular}{@{}llccc@{}}
\toprule
\textbf{Domain} & \textbf{Split} & \textbf{Hit@1} & \textbf{Hit@5} & \textbf{Hit@10} \\
\midrule
\multirow{2}{*}{Blocksworld} & Interpolation & 96.2$\pm$0.2 & 100.0$\pm$0.0 & 100.0$\pm$0.0 \\
 & Extrapolation & 17.6$\pm$5.2 & 49.1$\pm$9.8 & 64.6$\pm$7.5 \\
\midrule
\multirow{2}{*}{Depot} & Interpolation & 79.5$\pm$0.7 & 98.8$\pm$0.1 & 99.2$\pm$0.1 \\
 & Extrapolation & 4.7$\pm$1.2 & 25.9$\pm$6.4 & 41.2$\pm$8.1 \\
\midrule
\multirow{2}{*}{Ferry} & Interpolation & 99.1$\pm$0.1 & 100.0$\pm$0.0 & 100.0$\pm$0.0 \\
 & Extrapolation & 12.0$\pm$1.7 & 40.6$\pm$3.5 & 58.1$\pm$4.0 \\
\midrule
\multirow{2}{*}{Floortile} & Interpolation & 97.4$\pm$0.2 & 99.6$\pm$0.0 & 99.8$\pm$0.0 \\
 & Extrapolation & 37.6$\pm$12.7 & 68.8$\pm$15.6 & 78.9$\pm$12.7 \\
\midrule
\multirow{2}{*}{Goldminer} & Interpolation & 95.8$\pm$0.3 & 100.0$\pm$0.0 & 100.0$\pm$0.0 \\
 & Extrapolation & 35.8$\pm$6.4 & 76.2$\pm$5.2 & 88.0$\pm$1.7 \\
\midrule
\multirow{2}{*}{Grid} & Interpolation & 84.3$\pm$0.8 & 99.8$\pm$0.0 & 99.9$\pm$0.0 \\
 & Extrapolation & 25.7$\pm$1.2 & 74.9$\pm$1.2 & 88.0$\pm$1.2 \\
\midrule
\multirow{2}{*}{Logistics} & Interpolation & 91.2$\pm$0.6 & 99.9$\pm$0.0 & 100.0$\pm$0.0 \\
 & Extrapolation & 16.5$\pm$3.5 & 53.7$\pm$9.8 & 70.8$\pm$9.8 \\
\midrule
\multirow{2}{*}{Rovers} & Interpolation & 97.8$\pm$0.1 & 99.0$\pm$0.1 & 99.5$\pm$0.1 \\
 & Extrapolation & 16.9$\pm$1.2 & 54.4$\pm$2.9 & 72.2$\pm$2.3 \\
\midrule
\multirow{2}{*}{Satellite} & Interpolation & 98.5$\pm$0.1 & 99.9$\pm$0.0 & 100.0$\pm$0.0 \\
 & Extrapolation & 14.4$\pm$2.3 & 47.5$\pm$5.8 & 66.9$\pm$5.8 \\
\bottomrule
\end{tabular}
\end{center}
\end{table}

\FloatBarrier
\section{Reference methods: specifications and full results}
\label{app:baselines}

All reference methods use the protocol of Section~\ref{sec:setup}: identical splits,
identically budgeted heads, a 128-candidate pool, and
best-epoch selection. Three domains spanning the observed range of extrapolation
difficulty, three seeds.

\subsection{Specifications}

\textbf{Non-learned floors.} \emph{Identity} predicts $\hat{s}'=E(s)$. \emph{Offset} adds
a mean training displacement keyed by ground action (\emph{grounded}) or action schema
(\emph{lifted}), with unseen keys falling back to the global mean. Both operate in the
encoder's own space.

\textbf{Non-LLM encoders.} Each replaces $E$ with a sparse featurization of the same text
through an identical head: \emph{TF-IDF} (word 1--2 grams, sublinear tf, smooth idf),
\emph{bag-of-words} (raw counts), \emph{char} (character 3--5 grams), and
\emph{bag-of-literals} (binary bag over the symbolic literal vocabulary). Vocabularies are fitted on training states only. Their widths range from 234 to 2{,}498 dimensions, and no
test state received an all-zero representation.

\textbf{Lifted STRIPS induction (oracle).} For each training transition $(s,a,s')$ with
$a=(\textit{schema}\ o_1\ldots o_n)$ we take $\textit{add}=s'\setminus s$,
$\textit{del}=s\setminus s'$, and replace each $o_i$ by a positional marker $?p_i$, and application substitutes back. One transition per schema suffices under STRIPS, and every
further transition is a consistency check. Disagreements with the first pair seen for a
schema number zero on Ferry and Logistics and 52 on Goldminer, whose
\texttt{fire-laser} and \texttt{detonate-bomb} have cell-dependent effects. An unseen
schema yields no prediction and is scored as a miss, which is why the method scores zero
under Cross-Domain and Leave-One-Out.

\textbf{Tabular one-hot.} Represents each state by a one-hot lookup row instead of its
frozen text embedding and is trained on identical data. It isolates what the frozen
embedding contributes beyond lookup: every test state that never occurred in training
has an untrained row. On Logistics under
Interpolation it reaches $8.3\%$ Hit@1 against $96.3\%$ for \textsc{EmbedPlan}, and on
held-out problems $0.2\%$, below the $0.8\%$ chance level, since every held-out state maps to an untrained row.
The near-total failure of lookup on identical data indicates that the generalization
reported in the main text comes from the geometry of the frozen embedding space.

\textbf{Null controls.} \emph{Context-only} deletes the \texttt{CURRENT STATE} block,
retaining problem and goal text, and therefore carries no state information.
\emph{Random} assigns a fixed Gaussian vector per state.

\subsection{Tie-breaking and the null controls}
\label{app:ties}

Ranks use worst-case tie-breaking: a candidate scoring exactly equal to the ground truth
counts against it. Any representation that collapses same-problem states produces ties
among all candidates, and deleting the state block does exactly this. On Ferry,
22{,}075 distinct states within one held-out problem map to a single vector. Under
best-case tie-breaking such a representation would attain $100.0\%$ Hit@5 while carrying no state information, whereas under the worst-case convention it attains $0.0\%$. The random
control, equally uninformative but with distinct vectors, scores near chance under both.
The floor of the Extrapolation protocol is therefore $0.0$, and the convention is
load-bearing.

\subsection{Full results}

\begin{table}[h]
\caption{Extrapolation, Hit@5 (\%), mean $\pm$ SE over 3 seeds. The Llama-3.3-70B row is
reproduced from Table~\ref{tab:per_domain_full}, and all other rows were computed for this
comparison under the identical protocol.}
\label{tab:app_extrap_h5}
\centering\small
\begin{tabular}{@{}lcccc@{}}
\toprule
\textbf{Method} & \textbf{Ferry} & \textbf{Logistics} & \textbf{Goldminer} & \textbf{Mean} \\
\midrule
\textit{Lifted STRIPS induction (oracle)} & $100.0 \pm 0.0$ & $100.0 \pm 0.0$ & $99.3 \pm 0.3$ & $99.8$ \\
\midrule
Char 3--5 grams & $71.1 \pm 1.7$ & $69.8 \pm 7.4$ & $97.3 \pm 0.7$ & $79.4$ \\
Bag-of-words & $70.8 \pm 2.7$ & $67.3 \pm 4.8$ & $99.3 \pm 0.2$ & $79.1$ \\
Bag-of-literals & $51.9 \pm 5.7$ & $81.4 \pm 0.7$ & $95.7 \pm 2.0$ & $76.3$ \\
TF-IDF bigrams & $59.5 \pm 2.1$ & $50.0 \pm 4.3$ & $97.2 \pm 1.2$ & $68.9$ \\
\midrule
\textsc{EmbedPlan} (Llama-3.3-70B) & $40.6 \pm 3.5$ & $53.7 \pm 9.8$ & $76.2 \pm 5.2$ & $56.8$ \\
\textsc{EmbedPlan} (BGE-M3) & $42.8 \pm 2.1$ & $51.7 \pm 3.9$ & $70.2 \pm 5.2$ & $54.9$ \\
\midrule
Offset, lifted & $34.3 \pm 0.8$ & $56.4 \pm 5.4$ & $50.7 \pm 1.7$ & $47.2$ \\
Offset, grounded & $28.6 \pm 2.1$ & $53.1 \pm 2.3$ & $57.3 \pm 4.1$ & $46.3$ \\
Identity & $18.5 \pm 1.0$ & $49.8 \pm 3.4$ & $41.2 \pm 1.6$ & $36.5$ \\
\midrule
Random vectors [null] & $4.4 \pm 0.5$ & $4.8 \pm 0.6$ & $5.2 \pm 0.5$ & $4.8$ \\
Context-only [null] & $0.0 \pm 0.0$ & $0.0 \pm 0.0$ & $0.0 \pm 0.0$ & $0.0$ \\
\bottomrule
\end{tabular}
\end{table}

\begin{table}[h]
\caption{Extrapolation, Hit@1 (\%), mean $\pm$ SE over 3 seeds. The separation between
\textsc{EmbedPlan} and the non-learned floors is wider here than at Hit@5.}
\label{tab:app_extrap_h1}
\centering\small
\begin{tabular}{@{}lcccc@{}}
\toprule
\textbf{Method} & \textbf{Ferry} & \textbf{Logistics} & \textbf{Goldminer} & \textbf{Mean} \\
\midrule
\textit{Lifted STRIPS induction (oracle)} & $100.0 \pm 0.0$ & $100.0 \pm 0.0$ & $99.3 \pm 0.3$ & $99.8$ \\
\midrule
Bag-of-words & $31.5 \pm 3.3$ & $25.4 \pm 4.2$ & $78.0 \pm 1.9$ & $45.0$ \\
Char 3--5 grams & $31.0 \pm 2.4$ & $24.2 \pm 2.4$ & $70.6 \pm 3.5$ & $41.9$ \\
TF-IDF bigrams & $24.7 \pm 1.5$ & $14.2 \pm 2.9$ & $72.8 \pm 2.5$ & $37.2$ \\
Bag-of-literals & $14.2 \pm 0.9$ & $30.7 \pm 0.8$ & $66.5 \pm 8.7$ & $37.1$ \\
\midrule
\textsc{EmbedPlan} (Llama-3.3-70B) & $12.0 \pm 1.7$ & $16.5 \pm 3.5$ & $35.8 \pm 6.4$ & $21.4$ \\
\textsc{EmbedPlan} (BGE-M3) & $12.9 \pm 0.6$ & $15.5 \pm 1.7$ & $31.3 \pm 5.3$ & $19.9$ \\
\midrule
Offset, grounded & $6.9 \pm 2.1$ & $5.0 \pm 0.4$ & $9.7 \pm 2.5$ & $7.2$ \\
Offset, lifted & $5.9 \pm 0.9$ & $4.9 \pm 0.9$ & $5.8 \pm 2.3$ & $5.5$ \\
Identity & $2.2 \pm 0.1$ & $4.4 \pm 0.8$ & $3.4 \pm 0.1$ & $3.3$ \\
\midrule
Random vectors [null] & $0.9 \pm 0.1$ & $0.6 \pm 0.2$ & $0.6 \pm 0.1$ & $0.7$ \\
Context-only [null] & $0.0 \pm 0.0$ & $0.0 \pm 0.0$ & $0.0 \pm 0.0$ & $0.0$ \\
\bottomrule
\end{tabular}
\end{table}

\begin{table}[h]
\caption{Interpolation, Hit@5 and Hit@1 (\%), mean $\pm$ SE over 3 seeds. Six methods lie
within $0.2$ points of one another on Hit@5.}
\label{tab:app_interp}
\centering\small
\begin{tabular}{@{}lcccc@{}}
\toprule
\textbf{Method} & \textbf{Ferry} & \textbf{Logistics} & \textbf{Goldminer} & \textbf{Mean} \\
\midrule
\multicolumn{5}{@{}l}{\textit{Hit@5}} \\
Bag-of-literals & $100.0 \pm 0.0$ & $100.0 \pm 0.0$ & $100.0 \pm 0.1$ & $100.0$ \\
Bag-of-words & $99.9 \pm 0.1$ & $100.0 \pm 0.0$ & $100.0 \pm 0.0$ & $100.0$ \\
\textsc{EmbedPlan} (Llama-3.3-70B) & $100.0 \pm 0.0$ & $99.9 \pm 0.0$ & $100.0 \pm 0.0$ & $100.0$ \\
TF-IDF bigrams & $99.9 \pm 0.0$ & $100.0 \pm 0.0$ & $100.0 \pm 0.0$ & $100.0$ \\
Char 3--5 grams & $99.9 \pm 0.0$ & $99.9 \pm 0.1$ & $100.0 \pm 0.0$ & $100.0$ \\
\textsc{EmbedPlan} (BGE-M3) & $100.0 \pm 0.0$ & $99.8 \pm 0.1$ & $99.9 \pm 0.0$ & $99.9$ \\
\textit{Lifted STRIPS induction (oracle)} & $100.0 \pm 0.0$ & $100.0 \pm 0.0$ & $99.5 \pm 0.1$ & $99.8$ \\
Offset, grounded & $90.1 \pm 0.0$ & $99.1 \pm 0.1$ & $96.1 \pm 0.3$ & $95.1$ \\
Offset, lifted & $79.1 \pm 0.2$ & $98.3 \pm 0.1$ & $89.6 \pm 0.3$ & $89.0$ \\
Identity & $49.6 \pm 0.2$ & $98.2 \pm 0.2$ & $73.2 \pm 0.2$ & $73.7$ \\
Random vectors [null] & $8.2 \pm 0.3$ & $6.2 \pm 0.2$ & $7.8 \pm 0.5$ & $7.4$ \\
Context-only [null] & $5.7 \pm 0.3$ & $6.2 \pm 0.5$ & $4.3 \pm 0.1$ & $5.4$ \\
\midrule
\multicolumn{5}{@{}l}{\textit{Hit@1}} \\
\textit{Lifted STRIPS induction (oracle)} & $100.0 \pm 0.0$ & $100.0 \pm 0.0$ & $99.5 \pm 0.1$ & $99.8$ \\
Bag-of-literals & $98.7 \pm 0.1$ & $97.9 \pm 0.2$ & $95.7 \pm 0.1$ & $97.4$ \\
Char 3--5 grams & $95.4 \pm 0.2$ & $97.1 \pm 0.0$ & $97.6 \pm 0.1$ & $96.7$ \\
Bag-of-words & $95.5 \pm 0.1$ & $96.4 \pm 1.0$ & $97.8 \pm 0.2$ & $96.6$ \\
TF-IDF bigrams & $94.3 \pm 0.2$ & $96.5 \pm 0.5$ & $98.0 \pm 0.1$ & $96.2$ \\
\textsc{EmbedPlan} (Llama-3.3-70B) & $99.1 \pm 0.1$ & $91.2 \pm 0.6$ & $95.8 \pm 0.3$ & $95.4$ \\
\textsc{EmbedPlan} (BGE-M3) & $97.8 \pm 0.2$ & $93.6 \pm 0.9$ & $94.3 \pm 0.8$ & $95.2$ \\
Offset, grounded & $72.6 \pm 0.1$ & $86.1 \pm 0.4$ & $69.5 \pm 0.6$ & $76.1$ \\
Offset, lifted & $51.7 \pm 0.6$ & $78.8 \pm 0.1$ & $54.6 \pm 0.4$ & $61.7$ \\
Identity & $24.6 \pm 0.2$ & $76.6 \pm 0.1$ & $39.5 \pm 0.3$ & $46.9$ \\
Random vectors [null] & $1.9 \pm 0.1$ & $1.2 \pm 0.1$ & $1.4 \pm 0.1$ & $1.5$ \\
Context-only [null] & $1.0 \pm 0.1$ & $0.0 \pm 0.0$ & $0.1 \pm 0.1$ & $0.4$ \\
\bottomrule
\end{tabular}
\end{table}

Under Interpolation every text representation exceeds $99.8\%$ Hit@5, so the protocol cannot separate them, and Hit@1 separates them only slightly ($95.2$--$97.4$). The main-text
comparison is therefore reported under Extrapolation.

\subsection{The extrapolation gap is partly a grounding effect}
\label{app:grounding}

A held-out problem introduces new objects, so many test transitions invoke a ground
action absent from training ($57.6\%$ on Ferry, $65.5\%$ on Logistics, $11.7\%$ on
Goldminer), yet \emph{no} test action schema is unseen. Keying the offset reference on
schemas rather than ground actions gains $5.7$ points on Ferry and $3.3$ on Logistics,
while costing $6.6$ on Goldminer where most groundings are already observed. The gap
reflects a representation keyed on object identity rather than an inability to acquire
domain structure.

\subsection{Seed variance}
\label{app:seed_variance}

The Extrapolation split shuffles the problem list and holds out the last $20\%$. With
$5$--$10$ problems per domain, distinct seeds need not yield distinct partitions: seeds
$42$ and $456$ select identical held-out problems in all three domains, and every
non-learned reference differs by exactly $0.00$ between them. The seed $42$--$456$
difference therefore isolates training stochasticity ($1.5$ points for
\textsc{EmbedPlan}), while the difference against seed $123$ reflects the choice of
held-out problems ($11.0$ points). Split choice dominates training noise by
$4$--$7\times$, so a three-seed spread over two distinct partitions
understates the true variance.

\FloatBarrier
\section{Preliminary study: unfreezing the encoder}
\label{app:finetune}

To bound what the frozen constraint costs, we adapt BGE-M3 with LoRA (rank $16$,
$\alpha{=}32$, on \texttt{query}, \texttt{key}, \texttt{value}, \texttt{dense}) jointly
with the heads and transition network, for two epochs at batch size $16$ with encoder
learning rate $10^{-5}$. Both arms then re-encode the pool with their own encoder and fit
a fresh, identically budgeted head, so only the embedding space differs. A \textbf{cold
start} unfreezes the encoder immediately, whereas a \textbf{warm start} first trains the head to
convergence in the unadapted space. Since LoRA is initialized to a zero delta, the
adapted encoder is bit-identical to the frozen one at that point and the warmed head
transfers exactly.

\begin{table}[h]
\caption{Full-pool Hit@1 (\%), mean over three domains and three seeds. The full pool
contains every state in the domain ($12{,}237$--$46{,}205$) rather than 128 candidates.}
\label{tab:app_finetune_main}
\centering\small
\begin{tabular}{@{}lcc@{}}
\toprule
\textbf{Encoder treatment} & \textbf{Extrapolation} & \textbf{Interpolation} \\
\midrule
Frozen           & $3.8$ & $25.6$ \\
LoRA, cold start & $6.2$ & $39.6$ \\
LoRA, warm start & $\mathbf{9.3}$ & $\mathbf{41.7}$ \\
\bottomrule
\end{tabular}
\end{table}

\begin{table}[h]
\caption{Effect of the warm start, in points of full-pool Hit@1 relative to the cold
start. The sign is consistent per domain across two independent splits.}
\label{tab:app_finetune_schedule}
\centering\small
\begin{tabular}{@{}lcc@{}}
\toprule
\textbf{Domain} & \textbf{Extrapolation} & \textbf{Interpolation} \\
\midrule
Goldminer & $+13.3$ & $+10.5$ \\
Ferry     & $-0.1$  & $+1.6$  \\
Logistics & $-3.9$  & $-5.9$  \\
\bottomrule
\end{tabular}
\end{table}

Adaptation is worth $1.6\times$ (Interpolation) to $2.4\times$ (Extrapolation) over the frozen encoder, and the
schedule matters in a domain-dependent but reproducible way. On Goldminer, cold-start
joint training is \emph{worse than not fine-tuning at all}: the adapter is optimized
against a transition function whose loss has not converged ($\approx 2.05$ during the
joint stage, against $0.59$ for the same head once the space stops moving) and which is
then discarded. Warming the head removes this. Across all nine domain--seed cells the
warm-started model is never worse than frozen, whereas the cold-started model is worse in
two. We report per-domain effects rather than a pooled test, since the effect changes
sign across domains.

Encoder capacity helps as well: scaling the frozen encoder from $568$M to $70$B
parameters raises Extrapolation Hit@5 from $36.3\%$ to $54.6\%$ averaged over the nine
domains (Table~\ref{tab:encoder}). On our templated states the sparse lexical references
of Appendix~\ref{app:baselines} remain ahead ($79.4\%$ against $56.8\%$ for the largest
encoder on the three domains compared there), because an action edits only a few facts of
an otherwise identical text, which sparse features encode directly while a pooled
embedding of the whole prompt barely moves (Section~\ref{sec:why}). The gap therefore
reflects the form of the representation rather than its size.

\FloatBarrier
\section{Sensitivity to fact ordering}
\label{app:factorder}

State descriptions render a \emph{set} of literals as a comma-separated clause list, so
the ordering is arbitrary and semantically null. We shuffle the clauses within the
\texttt{CURRENT STATE} block, leaving problem and goal text untouched, and compare the
resulting displacement against that of a real transition, over $256$ states per domain
and four reorderings each.

\begin{table}[h]
\caption{Embedding displacement from a semantically null reordering
($d_{\text{order}} = 1-\cos(s,\sigma(s))$, with $\sigma$ a random clause order) against that from a real transition
($d_{\text{action}} = 1-\cos(s,s')$), BGE-M3.}
\label{tab:app_factorder}
\centering\small
\begin{tabular}{@{}lccc@{}}
\toprule
\textbf{Domain} & $\cos(s,s')$ & $\cos(s,\sigma(s))$ & $d_{\text{order}}/d_{\text{action}}$ \\
\midrule
Ferry     & $0.99672$ & $0.99654$ & $1.06$ \\
Goldminer & $0.99881$ & $0.99862$ & $1.16$ \\
Logistics & $0.99928$ & $0.99915$ & $1.18$ \\
\midrule
\textbf{Mean} & & & $\mathbf{1.13}$ \\
\bottomrule
\end{tabular}
\end{table}

Reordering displaces the embedding slightly \emph{further} than applying an action does,
in every domain. This is consistent with the ordering in Appendix~\ref{app:baselines}:
the representations performing best under Extrapolation are bags of words, character
n-grams, or literals and are largely insensitive to clause order. Order augmentation during encoding is a concrete direction for
future work.
\FloatBarrier
\section{Terms, common questions and released resources}
\label{app:peo}

This appendix gathers the paper's terms in one place, answers the questions a reader is likely to
bring to it, each with a pointer to the full account, and lists what is released.

\paragraph{Keywords}
\textsc{EmbedPlan}, LLM planning, transition model, world model, latent transitions, frozen text
embeddings, classical planning, PDDL, \textsc{ACPBench}, contrastive learning, nearest-neighbor
retrieval, generalization.

\subsection{Terms}

\begin{description}
\item[\textsc{EmbedPlan}.] A transition model built on frozen text embeddings: it embeds natural
language descriptions of the state and the action with a frozen LLM, predicts the embedding of the
next state with a lightweight learned network, and returns the closest real state
(Section~\ref{sec:setup}).
\item[Transition model.] A model that predicts how each action changes the current state, which
planning requires (Section~\ref{sec:introduction}).
\item[Frozen encoder $E$.] A frozen LLM encoder mapping text to $\mathbb{R}^d$: MPNet
(\mbox{all-mpnet-base-v2}), \mbox{BGE-M3}, \mbox{Qwen2.5-7B} or \mbox{Llama-3.3-70B} (Section~\ref{sec:setup},
Appendix~\ref{app:encoders}).
\item[Head.] The learned state and action projection heads $\pi_s, \pi_a$, which map to
$d' = 128$ dimensions, together with the transition network $T_\theta$ that predicts the next-state
embedding. Only the head is trained (Section~\ref{sec:setup}).
\item[Candidate pool.] The set $\mathcal{C}$ of encoded states from which the nearest neighbor, by
cosine similarity, is returned as the next state (Section~\ref{sec:setup}).
\item[Domain, problem, grounded and lifted actions.] A domain fixes the action schemas, and a
problem fixes the objects, the initial state, and the goal. Each action is a grounded instance, such
as \texttt{pick-up(C)}, of a lifted action schema, \texttt{pick-up(?x)} (Section~\ref{sec:setup}).
\item[Hit@$k$.] The fraction of queries whose true next state ranks in the top $k$ of a candidate
pool. With the default pool of 128 states, chance is $3.9\%$ for Hit@5 and $0.8\%$ for Hit@1
(Section~\ref{sec:setup}).
\item[Acc@$k$.] The fraction of queries whose true action ranks in the top $k$ when every action
applicable in the state is applied and ranked by how close its prediction comes to the true next
state (Section~\ref{sec:within}).
\item[Snapping.] In multi-step rollout, replacing each state prediction by the nearest real state
the model retrieves, right or wrong, so that this state becomes the next input
(Section~\ref{sec:within}).
\item[Lifted STRIPS induction.] A reference method that infers each action schema's add and delete
effects from the symbolic facts of each state and serves as an oracle upper bound
(Section~\ref{sec:setup}).
\end{description}

\subsection{Common questions}

\paragraph{What is \textsc{EmbedPlan}?}
\textsc{EmbedPlan} is a transition model for planning that runs on frozen text embeddings. A frozen
LLM embeds a natural language description of the state and of the action, a lightweight learned
network predicts the embedding of the next state, and nearest-neighbor retrieval returns the closest
real state, which can be fed back for multi-step rollout. The transition network has fewer than 500K
parameters. \textsc{EmbedPlan} is a transition component, not a complete planner: action selection
and search remain external (Section~\ref{sec:setup}).

\paragraph{Why not use an LLM directly as the transition model?}
Generating each next state with an LLM makes search slow and expensive. When an LLM serves as the
world model of a planner, every next state is generated token by token, with a full forward pass per
token, which makes multi-step lookahead and rollout-based search prohibitively expensive. Prompting
methods \citep{wei2022cot, yao2023tot, yao2023react} still query the LLM at every step, and
compilation to PDDL \citep{liu2023llmp, guan2023leveraging} requires a domain that can be expressed
as a symbolic model. \textsc{EmbedPlan} instead turns each planning step into a cheap vector
operation over frozen embeddings (Section~\ref{sec:introduction}).

\paragraph{Which datasets, domains and models does the paper use?}
The paper uses nine classical PDDL domains from \textsc{ACPBench} \citep{kokel2025acpbench}:
Blocksworld, Depot, Ferry, Floortile, Goldminer, Grid, Logistics, Rovers and Satellite, with states
rendered as natural language. It extracts 2.97M state-action-next-state transitions over 67 problems
(259K unique states). It compares four frozen encoders, MPNet (\mbox{all-mpnet-base-v2}), \mbox{BGE-M3},
\mbox{Qwen2.5-7B} and \mbox{Llama-3.3-70B}, from 768 to 8{,}192 dimensions, and reports \mbox{Llama-3.3-70B} unless
another encoder is named (Section~\ref{sec:setup}, Appendix~\ref{app:dataset}).

\paragraph{How is transition accuracy measured?}
Transition accuracy is measured by Hit@$k$, the fraction of queries whose true next state ranks in
the top $k$ of a pool of 128 states, the true successor and 127 distractors, so chance is $3.9\%$ for
Hit@5. Six protocols form a ladder of exposure. Interpolation and Plan-Variant test observed
problems, Extrapolation and Multi-Domain test unseen problems of an observed domain, and
Cross-Domain and Leave-One-Out test unseen domains (Section~\ref{sec:setup},
Table~\ref{tab:protocols}).

\paragraph{How accurate is \textsc{EmbedPlan} on problems seen during training?}
Within observed problems, \textsc{EmbedPlan} is near-perfect: Interpolation Hit@5 is $99.7\%$ with
\mbox{Llama-3.3-70B}, and mean Hit@1 is $92.1\%$. When the pool grows to every observed state, the true
successor stays in the top five for $86.9\%$ of Ferry and $74.9\%$ of Logistics queries, although
Hit@1 falls about $3\times$. With snapping, multi-step rollout keeps $92$ to $99\%$ of the accuracy
obtained with the true state at each step, over short test trajectories (mean $2.2$ steps on Ferry)
(Section~\ref{sec:within}, Tables~\ref{tab:pool_main} and~\ref{tab:closed_loop}).

\paragraph{How well does \textsc{EmbedPlan} generalize to unseen problems and unseen domains?}
Accuracy falls with each step down the exposure ladder. On unseen problems of an observed domain
(Extrapolation), Hit@5 is $54.6\%$, $14\times$ chance, and ranges by domain from $26$ to $76\%$.
One model trained on all nine domains reaches $37.2\%$. On unseen domains, Cross-Domain reaches
$6.6\%$, only $2.7$ points above chance, and Leave-One-Out reaches $9.2\%$ despite eight training
domains. The one notable transfer is Ferry to Logistics ($22.3\%$), two domains that both transport
objects between locations (Section~\ref{sec:boundary}, Tables~\ref{tab:hierarchy} and~\ref{tab:cross_domain_matrix}).

\paragraph{What limits transfer: the learned transition or the text representation?}
The frozen state representation is the main limit, not the learned transition. With the head, data
and pool fixed, the same head fed character 3 to 5 grams of the same text reaches $79.4\%$ Hit@5 on
unseen problems of Ferry, Logistics and Goldminer, against $56.8\%$ with \mbox{Llama-3.3-70B} embeddings,
and lifted STRIPS induction from the symbolic facts reaches $99.8\%$. On these templated states an
action edits only a few facts, so a pooled LLM embedding of the whole prompt barely moves (cosine
$0.997$ to $0.999$ between consecutive states with \mbox{BGE-M3}) (Section~\ref{sec:why},
Table~\ref{tab:baselines_extrap}).

\paragraph{How does \textsc{EmbedPlan} compare with an LLM that generates the next state?}
Within observed problems, generation is the harder route. Asked to write the next state, five LLMs
from \mbox{Llama-3.1-8B} to \mbox{GPT-5.4} all do well on Logistics, but none matches the true Ferry state more
than $56\%$ of the time, while \textsc{EmbedPlan}'s retrieval Hit@5 on Ferry is $100.0\%$. The
comparison is matched in task, not in information: \textsc{EmbedPlan} has observed other transitions
of the same problems, whereas the LLMs are zero-shot, and on unseen problems its Ferry Hit@1 falls to
$12.0\%$ (Section~\ref{sec:within}, Tables~\ref{tab:llm_baselines} and~\ref{tab:full_metrics}).

\paragraph{What are the limitations?}
\textsc{EmbedPlan} is a transition component for discrete, domain-specific settings, not a
domain-general planner, and it must be trained per domain because cross-domain transfer fails.
Retrieval assumes a candidate pool that contains the true successor. The domains are templated, so
states without a clean literal decomposition, the regime the text interface is meant for, are not
yet tested. Pool scaling, multi-step rollout and LLM ranking are measured within observed problems
only, and Extrapolation rests on one or two held-out problems per domain
(Section~\ref{sec:limitations}).

\paragraph{Where are the code and the data?}
The code, with the scripts that generate the transition datasets, is at
\url{https://github.com/embedplan/EmbedPlan}. The processed transition datasets are derived from
\textsc{ACPBench} \citep{kokel2025acpbench} with additional processing that its public release does
not include, and they will be released upon acceptance. All four base encoders, \mbox{Llama-3.3-70B},
\mbox{Qwen2.5-7B}, MPNet and \mbox{BGE-M3}, are publicly available (Appendix~\ref{app:code},
Appendix~\ref{app:dataset_stats}).

\subsection{Where this work sits}

\textsc{EmbedPlan} sits where LLM planning meets learned world models. It starts from
prompting-based planning and reasoning \citep{wei2022cot, yao2023tot, yao2023react, hao2023rap},
from compiling natural language into PDDL for classical planners \citep{liu2023llmp,
guan2023leveraging, oswald-et-al-icaps2024, tantakoun-etal-2025-llms, zuo-etal-2025-planetarium},
and from the finding that LLM outputs in either role can be hallucinated or unexecutable
\citep{kambhampati2024position, katz2024thought}. It borrows its form from latent world models and
model-based reinforcement learning \citep{asai2018latplan, ha2018worldmodels, hafner2019planet,
hafner2020dreamer, schrittwieser2020muzero}, and closest to it, from contrastive latent transitions
\citep{kipf2020cswm} and dynamics over frozen pretrained visual features \citep{zhou2025dinowm}. It
adds a learned latent state to text world models, which keep the state in text
\citep{wang2024textworldsim, li2026bridgingagentworldgaptext}. Its structured counterparts, each
presupposing a symbolic decomposition of the state, are object-centric architectures
\citep{battaglia2018relational}, generalized planners over lifted PDDL \citep{toyer2018asnets,
shen2020stripshgn, stahlberg2022gnn} and action-model learning \citep{yang2007arms,
cresswell2013locm, aineto2019fama, juba2021sam}. It builds on sentence embeddings and contrastive
representation learning \citep{reimers2019sbert, gao2021simcse, oord2018cpc}, evaluates on classical
planning benchmarks \citep{mcdermott1998pddl, kokel2025acpbench}, and its transfer results echo
compositional generalization benchmarks, where neural models often memorize training patterns
\citep{lake2018scan, kim2020cogs}.


\begin{thebibliography}{51}
\providecommand{\natexlab}[1]{#1}
\providecommand{\url}[1]{\texttt{#1}}
\expandafter\ifx\csname urlstyle\endcsname\relax
  \providecommand{\doi}[1]{doi: #1}\else
  \providecommand{\doi}{doi: \begingroup \urlstyle{rm}\Url}\fi

\bibitem[Adhikari et~al.(2020)Adhikari, Yuan, Côté, Zelinka, Rondeau,
  Laroche, Poupart, Tang, Trischler, and Hamilton]{adhikari2020gata}
Ashutosh Adhikari, Xingdi Yuan, Marc-Alexandre Côté, Mikuláš Zelinka,
  Marc-Antoine Rondeau, Romain Laroche, Pascal Poupart, Jian Tang, Adam
  Trischler, and Will Hamilton.
\newblock Learning {Dynamic} {Belief} {Graphs} to {Generalize} on
  {Text}-{Based} {Games}.
\newblock In \emph{Advances in {Neural} {Information} {Processing} {Systems}},
  volume~33, pp.\  3045--3057. Curran Associates, Inc., 2020.
\newblock URL
  \url{https://proceedings.neurips.cc/paper_files/paper/2020/hash/1fc30b9d4319760b04fab735fbfed9a9-Abstract.html}.

\bibitem[Aineto et~al.(2019)Aineto, Jiménez~Celorrio, and
  Onaindia]{aineto2019fama}
Diego Aineto, Sergio Jiménez~Celorrio, and Eva Onaindia.
\newblock Learning action models with minimal observability.
\newblock \emph{Artificial Intelligence}, 275:\penalty0 104--137, October 2019.
\newblock ISSN 0004-3702.
\newblock \doi{10.1016/j.artint.2019.05.003}.
\newblock URL
  \url{https://www.sciencedirect.com/science/article/pii/S0004370218304259}.

\bibitem[Asai \& Fukunaga(2018)Asai and Fukunaga]{asai2018latplan}
Masataro Asai and Alex Fukunaga.
\newblock Classical planning in deep latent space: Bridging the
  subsymbolic-symbolic boundary.
\newblock In \emph{Proceedings of the aaai conference on artificial
  intelligence}, volume~32, 2018.

\bibitem[Assran et~al.(2023)Assran, Duval, Misra, Bojanowski, Vincent, Rabbat,
  LeCun, and Ballas]{assran2023self}
Mahmoud Assran, Quentin Duval, Ishan Misra, Piotr Bojanowski, Pascal Vincent,
  Michael Rabbat, Yann LeCun, and Nicolas Ballas.
\newblock Self-supervised learning from images with a joint-embedding
  predictive architecture.
\newblock In \emph{Proceedings of the IEEE/CVF Conference on Computer Vision
  and Pattern Recognition}, pp.\  15619--15629, 2023.

\bibitem[Battaglia et~al.(2018)Battaglia, Hamrick, Bapst, Sanchez-Gonzalez,
  Zambaldi, Malinowski, Tacchetti, Raposo, Santoro, Faulkner,
  et~al.]{battaglia2018relational}
Peter~W Battaglia, Jessica~B Hamrick, Victor Bapst, Alvaro Sanchez-Gonzalez,
  Vinicius Zambaldi, Mateusz Malinowski, Andrea Tacchetti, David Raposo, Adam
  Santoro, Ryan Faulkner, et~al.
\newblock Relational inductive biases, deep learning, and graph networks.
\newblock \emph{arXiv preprint arXiv:1806.01261}, 2018.

\bibitem[Cresswell et~al.(2013)Cresswell, McCluskey, and
  West]{cresswell2013locm}
Stephen~N. Cresswell, Thomas~L. McCluskey, and Margaret~M. West.
\newblock Acquiring planning domain models using {LOCM}.
\newblock \emph{The Knowledge Engineering Review}, 28\penalty0 (2):\penalty0
  195--213, June 2013.
\newblock ISSN 0269-8889, 1469-8005.
\newblock \doi{10.1017/S0269888912000422}.
\newblock URL
  \url{https://www.cambridge.org/core/journals/knowledge-engineering-review/article/abs/acquiring-planning-domain-models-using-locm/139F7A0986CD481F88A778B08891DC49}.

\bibitem[Gao et~al.(2021)Gao, Yao, and Chen]{gao2021simcse}
Tianyu Gao, Xingcheng Yao, and Danqi Chen.
\newblock Simcse: Simple contrastive learning of sentence embeddings.
\newblock In \emph{Proceedings of the 2021 Conference on Empirical Methods in
  Natural Language Processing}, pp.\  6894--6910, 2021.

\bibitem[Gieselmann \& Pokorny(2022)Gieselmann and
  Pokorny]{gieselmann2022latent}
Robert Gieselmann and Florian~T Pokorny.
\newblock Latent planning via expansive tree search.
\newblock \emph{Advances in Neural Information Processing Systems},
  35:\penalty0 16821--16835, 2022.

\bibitem[Guan et~al.(2023)Guan, Valmeekam, Sreedharan, and
  Kambhampati]{guan2023leveraging}
Lin Guan, Karthik Valmeekam, Sarath Sreedharan, and Subbarao Kambhampati.
\newblock Leveraging pre-trained large language models to construct and utilize
  world models for model-based task planning.
\newblock \emph{Advances in Neural Information Processing Systems},
  36:\penalty0 79081--79094, 2023.

\bibitem[Ha \& Schmidhuber(2018)Ha and Schmidhuber]{ha2018worldmodels}
David Ha and J{\"u}rgen Schmidhuber.
\newblock Recurrent world models facilitate policy evolution.
\newblock \emph{Advances in neural information processing systems}, 31, 2018.

\bibitem[Ha et~al.(2017)Ha, Dai, and Le]{ha2017hypernetworks}
David Ha, Andrew~M. Dai, and Quoc~V. Le.
\newblock Hypernetworks.
\newblock In \emph{International Conference on Learning Representations}, 2017.
\newblock URL \url{https://openreview.net/forum?id=rkpACe1lx}.

\bibitem[Hafner et~al.(2019)Hafner, Lillicrap, Fischer, Villegas, Ha, Lee, and
  Davidson]{hafner2019planet}
Danijar Hafner, Timothy Lillicrap, Ian Fischer, Ruben Villegas, David Ha,
  Honglak Lee, and James Davidson.
\newblock Learning latent dynamics for planning from pixels.
\newblock In \emph{International conference on machine learning}, pp.\
  2555--2565. PMLR, 2019.

\bibitem[Hafner et~al.(2020)Hafner, Lillicrap, Ba, and
  Norouzi]{hafner2020dreamer}
Danijar Hafner, Timothy Lillicrap, Jimmy Ba, and Mohammad Norouzi.
\newblock Dream to control: Learning behaviors by latent imagination.
\newblock In \emph{International Conference on Learning Representations}, 2020.

\bibitem[Hao et~al.(2023)Hao, Gu, Ma, Hong, Wang, Wang, and Hu]{hao2023rap}
Shibo Hao, Yi~Gu, Haodi Ma, Joshua Hong, Zhen Wang, Daisy Wang, and Zhiting Hu.
\newblock Reasoning with language model is planning with world model.
\newblock In \emph{Proceedings of the 2023 Conference on Empirical Methods in
  Natural Language Processing}, pp.\  8154--8173, 2023.

\bibitem[Huang et~al.(2026)Huang, Chen, Yao, Wang, Yang, Du, Zhao, Zhao, Lin,
  Rajmohan, and Zhang]{huang2026stateconsistencybehaviorconsistency}
Youling Huang, Guanqiao Chen, Junchi Yao, Lu~Wang, Fangkai Yang, Chao Du,
  ChenZhuo Zhao, Pu~Zhao, Qingwei Lin, Saravan Rajmohan, and Dongmei Zhang.
\newblock Beyond state consistency: Behavior consistency in text-based world
  models, 2026.
\newblock URL \url{https://arxiv.org/abs/2604.13824}.

\bibitem[Humphreys et~al.(2022)Humphreys, Guez, Tieleman, Sifre, Weber, and
  Lillicrap]{humphreys2022large}
Peter Humphreys, Arthur Guez, Olivier Tieleman, Laurent Sifre, Th{\'e}ophane
  Weber, and Timothy Lillicrap.
\newblock Large-scale retrieval for reinforcement learning.
\newblock \emph{Advances in Neural Information Processing Systems},
  35:\penalty0 20092--20104, 2022.

\bibitem[Juba et~al.(2021)Juba, Le, and Stern]{juba2021sam}
Brendan Juba, Hai~S. Le, and Roni Stern.
\newblock Safe {Learning} of {Lifted} {Action} {Models}, July 2021.
\newblock URL \url{http://arxiv.org/abs/2107.04169}.
\newblock arXiv:2107.04169 [cs.AI].

\bibitem[Kambhampati et~al.(2024)Kambhampati, Valmeekam, Guan, Verma, Stechly,
  Bhambri, Saldyt, and Murthy]{kambhampati2024position}
Subbarao Kambhampati, Karthik Valmeekam, Lin Guan, Mudit Verma, Kaya Stechly,
  Siddhant Bhambri, Lucas~Paul Saldyt, and Anil~B Murthy.
\newblock Position: Llms can’t plan, but can help planning in llm-modulo
  frameworks.
\newblock In \emph{Forty-first International Conference on Machine Learning},
  2024.

\bibitem[Katz et~al.(2024)Katz, Kokel, Srinivas, and
  Sohrabi~Araghi]{katz2024thought}
Michael Katz, Harsha Kokel, Kavitha Srinivas, and Shirin Sohrabi~Araghi.
\newblock Thought of search: Planning with language models through the lens of
  efficiency.
\newblock \emph{Advances in Neural Information Processing Systems},
  37:\penalty0 138491--138568, 2024.

\bibitem[Khandelwal et~al.(2020)Khandelwal, Levy, Jurafsky, Zettlemoyer, and
  Lewis]{khandelwalgeneralization}
Urvashi Khandelwal, Omer Levy, Dan Jurafsky, Luke Zettlemoyer, and Mike Lewis.
\newblock Generalization through memorization: Nearest neighbor language
  models.
\newblock In \emph{International Conference on Learning Representations}, 2020.

\bibitem[Kim \& Linzen(2020)Kim and Linzen]{kim2020cogs}
Najoung Kim and Tal Linzen.
\newblock Cogs: A compositional generalization challenge based on semantic
  interpretation.
\newblock In \emph{Empirical Methods in Natural Language Processing}, 2020.

\bibitem[Kipf et~al.(2020)Kipf, van~der Pol, and Welling]{kipf2020cswm}
Thomas Kipf, Elise van~der Pol, and Max Welling.
\newblock Contrastive learning of structured world models.
\newblock In \emph{International Conference on Learning Representations}, 2020.
\newblock URL \url{https://openreview.net/forum?id=H1gax6VtDB}.

\bibitem[Kokel et~al.(2025)Kokel, Katz, Srinivas, and
  Sohrabi]{kokel2025acpbench}
Harsha Kokel, Michael Katz, Kavitha Srinivas, and Shirin Sohrabi.
\newblock Acpbench: Reasoning about action, change, and planning.
\newblock In \emph{Proceedings of the AAAI Conference on Artificial
  Intelligence}, volume~39, pp.\  26559--26568, 2025.

\bibitem[Lake \& Baroni(2018)Lake and Baroni]{lake2018scan}
Brenden Lake and Marco Baroni.
\newblock Generalization without systematicity: On the compositional skills of
  sequence-to-sequence recurrent networks.
\newblock In \emph{International conference on machine learning}, pp.\
  2873--2882. PMLR, 2018.

\bibitem[Laskin et~al.(2020)Laskin, Srinivas, and Abbeel]{srinivas2020curl}
Michael Laskin, Aravind Srinivas, and Pieter Abbeel.
\newblock Curl: Contrastive unsupervised representations for reinforcement
  learning.
\newblock In \emph{International conference on machine learning}, pp.\
  5639--5650. PMLR, 2020.

\bibitem[Li et~al.(2026)Li, Wang, Lai, Ruan, Zhu, Zhu, Zhao, Hu, Chen, Yang,
  Li, Pan, Pan, Chen, Liu, and Li]{li2026bridgingagentworldgaptext}
Yixia Li, Hongru Wang, Peng Lai, Zhiwen Ruan, He~Zhu, Youxin Zhu, Ganlong Zhao,
  Minda Hu, Yun Chen, Sibei Yang, Peng Li, Jeff~Z. Pan, Jia Pan, Guanhua Chen,
  Yang Liu, and Guanbin Li.
\newblock Bridging the agent-world gap: Text world models for llm-based agents,
  2026.
\newblock URL \url{https://arxiv.org/abs/2606.09032}.

\bibitem[Lin et~al.(2024)Lin, Du, Watkins, Hafner, Abbeel, Klein, and
  Dragan]{lin2024dynalang}
Jessy Lin, Yuqing Du, Olivia Watkins, Danijar Hafner, Pieter Abbeel, Dan Klein,
  and Anca Dragan.
\newblock Learning to {Model} the {World} with {Language}, May 2024.
\newblock URL \url{http://arxiv.org/abs/2308.01399}.
\newblock arXiv:2308.01399 [cs.CL].

\bibitem[Liu et~al.(2023)Liu, Jiang, Zhang, Liu, Zhang, Biswas, and
  Stone]{liu2023llmp}
Bo~Liu, Yuqian Jiang, Xiaohan Zhang, Qiang Liu, Shiqi Zhang, Joydeep Biswas,
  and Peter Stone.
\newblock Llm+ p: Empowering large language models with optimal planning
  proficiency.
\newblock \emph{arXiv preprint arXiv:2304.11477}, 2023.

\bibitem[McDermott et~al.(1998)McDermott, Ghallab, Howe, Knoblock, Ram, Veloso,
  Weld, and Wilkins]{mcdermott1998pddl}
Drew McDermott, Malik Ghallab, Adele Howe, Craig Knoblock, Ashwin Ram, Manuela
  Veloso, Daniel Weld, and David Wilkins.
\newblock {PDDL} -- {The} {Planning} {Domain} {Definition} {Language} --
  {Version} 1.2.
\newblock Technical Report CVC TR-98-003/DCS TR-1165, Yale Center for
  Computational Vision and Control, 1998.

\bibitem[Micheli et~al.(2023)Micheli, Alonso, and Fleuret]{micheli2023iris}
Vincent Micheli, Eloi Alonso, and Fran{\c{c}}ois Fleuret.
\newblock Transformers are sample-efficient world models.
\newblock In \emph{International Conference on Learning Representations}, 2023.

\bibitem[Muennighoff et~al.(2023)Muennighoff, Tazi, Magne, and
  Reimers]{muennighoff2022mteb}
Niklas Muennighoff, Nouamane Tazi, Lo{\"\i}c Magne, and Nils Reimers.
\newblock Mteb: Massive text embedding benchmark.
\newblock In \emph{Proceedings of the 17th Conference of the European Chapter
  of the Association for Computational Linguistics}, pp.\  2014--2037, 2023.

\bibitem[Oord et~al.(2018)Oord, Li, and Vinyals]{oord2018cpc}
Aaron van~den Oord, Yazhe Li, and Oriol Vinyals.
\newblock Representation learning with contrastive predictive coding.
\newblock \emph{arXiv preprint arXiv:1807.03748}, 2018.

\bibitem[Oswald et~al.(2024)Oswald, Srinivas, Kokel, Lee, Katz, and
  Sohrabi]{oswald-et-al-icaps2024}
James Oswald, Kavitha Srinivas, Harsha Kokel, Junkyu Lee, Michael Katz, and
  Shirin Sohrabi.
\newblock Large language models as planning domain generators.
\newblock In \emph{Proceedings of the Thirty-Fourth International Conference on
  Automated Planning and Scheduling (ICAPS 2024)}, pp.\  423--431. AAAI Press,
  2024.

\bibitem[Oved et~al.(2025)Oved, Shlomov, Zeltyn, Mashkif, and
  Yaeli]{oved2025snap}
Alon Oved, Segev Shlomov, Sergey Zeltyn, Nir Mashkif, and Avi Yaeli.
\newblock Snap: semantic stories for next activity prediction.
\newblock In \emph{Proceedings of the AAAI Conference on Artificial
  Intelligence}, volume~39, pp.\  28871--28877, 2025.

\bibitem[Perez et~al.(2018)Perez, Strub, De~Vries, Dumoulin, and
  Courville]{perez2018film}
Ethan Perez, Florian Strub, Harm De~Vries, Vincent Dumoulin, and Aaron
  Courville.
\newblock Film: Visual reasoning with a general conditioning layer.
\newblock In \emph{Proceedings of the AAAI conference on artificial
  intelligence}, volume~32, 2018.

\bibitem[Radford et~al.(2021)Radford, Kim, Hallacy, Ramesh, Goh, Agarwal,
  Sastry, Askell, Mishkin, Clark, et~al.]{radford2021clip}
Alec Radford, Jong~Wook Kim, Chris Hallacy, Aditya Ramesh, Gabriel Goh,
  Sandhini Agarwal, Girish Sastry, Amanda Askell, Pamela Mishkin, Jack Clark,
  et~al.
\newblock Learning transferable visual models from natural language
  supervision.
\newblock In \emph{International conference on machine learning}, pp.\
  8748--8763. PmLR, 2021.

\bibitem[Reimers \& Gurevych(2019)Reimers and Gurevych]{reimers2019sbert}
Nils Reimers and Iryna Gurevych.
\newblock Sentence-bert: Sentence embeddings using siamese bert-networks.
\newblock In \emph{Proceedings of the 2019 Conference on Empirical Methods in
  Natural Language Processing and the 9th International Joint Conference on
  Natural Language Processing (EMNLP-IJCNLP)}, pp.\  3982--3992, 2019.

\bibitem[Schrittwieser et~al.(2020)Schrittwieser, Antonoglou, Hubert, Simonyan,
  Sifre, Schmitt, Guez, Lockhart, Hassabis, Graepel,
  et~al.]{schrittwieser2020muzero}
Julian Schrittwieser, Ioannis Antonoglou, Thomas Hubert, Karen Simonyan,
  Laurent Sifre, Simon Schmitt, Arthur Guez, Edward Lockhart, Demis Hassabis,
  Thore Graepel, et~al.
\newblock Mastering atari, go, chess and shogi by planning with a learned
  model.
\newblock \emph{Nature}, 588\penalty0 (7839):\penalty0 604--609, 2020.

\bibitem[Schwarzer et~al.(2021)Schwarzer, Anand, Goel, Hjelm, Courville, and
  Bachman]{schwarzer2021spr}
Max Schwarzer, Ankesh Anand, Rishab Goel, R~Devon Hjelm, Aaron Courville, and
  Philip Bachman.
\newblock Data-efficient reinforcement learning with self-predictive
  representations.
\newblock In \emph{International Conference on Learning Representations}, 2021.

\bibitem[Shen et~al.(2020)Shen, Trevizan, and Thiébaux]{shen2020stripshgn}
William Shen, Felipe Trevizan, and Sylvie Thiébaux.
\newblock Learning {Domain}-{Independent} {Planning} {Heuristics} with
  {Hypergraph} {Networks}.
\newblock \emph{Proceedings of the International Conference on Automated
  Planning and Scheduling}, 30:\penalty0 574--584, June 2020.
\newblock ISSN 2334-0843.
\newblock \doi{10.1609/icaps.v30i1.6754}.
\newblock URL \url{https://ojs.aaai.org/index.php/ICAPS/article/view/6754}.

\bibitem[Shinn et~al.(2023)Shinn, Cassano, Gopinath, Narasimhan, and
  Yao]{shinn2023reflexion}
Noah Shinn, Federico Cassano, Ashwin Gopinath, Karthik Narasimhan, and Shunyu
  Yao.
\newblock Reflexion: Language agents with verbal reinforcement learning.
\newblock \emph{Advances in Neural Information Processing Systems},
  36:\penalty0 8634--8652, 2023.

\bibitem[Ståhlberg et~al.(2022)Ståhlberg, Bonet, and
  Geffner]{stahlberg2022gnn}
Simon Ståhlberg, Blai Bonet, and Hector Geffner.
\newblock Learning {General} {Optimal} {Policies} with {Graph} {Neural}
  {Networks}: {Expressive} {Power}, {Transparency}, and {Limits}.
\newblock \emph{Proceedings of the International Conference on Automated
  Planning and Scheduling}, 32:\penalty0 629--637, June 2022.
\newblock ISSN 2334-0843.
\newblock \doi{10.1609/icaps.v32i1.19851}.
\newblock URL \url{https://ojs.aaai.org/index.php/ICAPS/article/view/19851}.

\bibitem[Tantakoun et~al.(2025)Tantakoun, Muise, and
  Zhu]{tantakoun-etal-2025-llms}
Marcus Tantakoun, Christian Muise, and Xiaodan Zhu.
\newblock {LLM}s as planning formalizers: A survey for leveraging large
  language models to construct automated planning models.
\newblock In Wanxiang Che, Joyce Nabende, Ekaterina Shutova, and Mohammad~Taher
  Pilehvar (eds.), \emph{Findings of the Association for Computational
  Linguistics: ACL 2025}, pp.\  25167--25188, Vienna, Austria, July 2025.
  Association for Computational Linguistics.
\newblock ISBN 979-8-89176-256-5.
\newblock \doi{10.18653/v1/2025.findings-acl.1291}.
\newblock URL \url{https://aclanthology.org/2025.findings-acl.1291/}.

\bibitem[Toyer et~al.(2018)Toyer, Trevizan, Thiébaux, and
  Xie]{toyer2018asnets}
Sam Toyer, Felipe Trevizan, Sylvie Thiébaux, and Lexing Xie.
\newblock Action {Schema} {Networks}: {Generalised} {Policies} {With} {Deep}
  {Learning}.
\newblock \emph{Proceedings of the AAAI Conference on Artificial Intelligence},
  32\penalty0 (1), April 2018.
\newblock ISSN 2374-3468.
\newblock \doi{10.1609/aaai.v32i1.12089}.
\newblock URL \url{https://ojs.aaai.org/index.php/AAAI/article/view/12089}.

\bibitem[Wang et~al.(2024)Wang, Todd, Xiao, Yuan, Côté, Clark, and
  Jansen]{wang2024textworldsim}
Ruoyao Wang, Graham Todd, Ziang Xiao, Xingdi Yuan, Marc-Alexandre Côté, Peter
  Clark, and Peter Jansen.
\newblock Can language models serve as text-based world simulators?
\newblock In \emph{Proceedings of the 62nd {Annual} {Meeting} of the
  {Association} for {Computational} {Linguistics} ({Volume} 2: {Short}
  {Papers})}, pp.\  1--17, 2024.

\bibitem[Wei et~al.(2022)Wei, Wang, Schuurmans, Bosma, Xia, Chi, Le, Zhou,
  et~al.]{wei2022cot}
Jason Wei, Xuezhi Wang, Dale Schuurmans, Maarten Bosma, Fei Xia, Ed~Chi, Quoc~V
  Le, Denny Zhou, et~al.
\newblock Chain-of-thought prompting elicits reasoning in large language
  models.
\newblock \emph{Advances in neural information processing systems},
  35:\penalty0 24824--24837, 2022.

\bibitem[Yang et~al.(2007)Yang, Wu, and Jiang]{yang2007arms}
Qiang Yang, Kangheng Wu, and Yunfei Jiang.
\newblock Learning action models from plan examples using weighted {MAX}-{SAT}.
\newblock \emph{Artificial Intelligence}, 171\penalty0 (2):\penalty0 107--143,
  February 2007.
\newblock ISSN 0004-3702.
\newblock \doi{10.1016/j.artint.2006.11.005}.
\newblock URL
  \url{https://www.sciencedirect.com/science/article/pii/S0004370206001408}.

\bibitem[Yao et~al.(2023{\natexlab{a}})Yao, Yu, Zhao, Shafran, Griffiths, Cao,
  and Narasimhan]{yao2023tot}
Shunyu Yao, Dian Yu, Jeffrey Zhao, Izhak Shafran, Tom Griffiths, Yuan Cao, and
  Karthik Narasimhan.
\newblock Tree of thoughts: Deliberate problem solving with large language
  models.
\newblock \emph{Advances in neural information processing systems},
  36:\penalty0 11809--11822, 2023{\natexlab{a}}.

\bibitem[Yao et~al.(2023{\natexlab{b}})Yao, Zhao, Yu, Du, Shafran, Narasimhan,
  and Cao]{yao2023react}
Shunyu Yao, Jeffrey Zhao, Dian Yu, Nan Du, Izhak Shafran, Karthik~R Narasimhan,
  and Yuan Cao.
\newblock React: Synergizing reasoning and acting in language models.
\newblock In \emph{The Eleventh International Conference on Learning
  Representations}, 2023{\natexlab{b}}.
\newblock URL \url{https://openreview.net/forum?id=WE_vluYUL-X}.

\bibitem[Zhou et~al.(2025)Zhou, Pan, Lecun, and Pinto]{zhou2025dinowm}
Gaoyue Zhou, Hengkai Pan, Yann Lecun, and Lerrel Pinto.
\newblock {DINO}-{WM}: World models on pre-trained visual features enable
  zero-shot planning.
\newblock In Aarti Singh, Maryam Fazel, Daniel Hsu, Simon Lacoste-Julien, Felix
  Berkenkamp, Tegan Maharaj, Kiri Wagstaff, and Jerry Zhu (eds.),
  \emph{Proceedings of the 42nd International Conference on Machine Learning},
  volume 267 of \emph{Proceedings of Machine Learning Research}, pp.\
  79115--79135. PMLR, 13--19 Jul 2025.
\newblock URL \url{https://proceedings.mlr.press/v267/zhou25t.html}.

\bibitem[Zuo et~al.(2025)Zuo, Velez, Li, Littman, and
  Bach]{zuo-etal-2025-planetarium}
Max Zuo, Francisco~Piedrahita Velez, Xiaochen Li, Michael Littman, and Stephen
  Bach.
\newblock Planetarium: A rigorous benchmark for translating text to structured
  planning languages.
\newblock In Luis Chiruzzo, Alan Ritter, and Lu~Wang (eds.), \emph{Proceedings
  of the 2025 Conference of the Nations of the Americas Chapter of the
  Association for Computational Linguistics: Human Language Technologies
  (Volume 1: Long Papers)}, pp.\  11223--11240, Albuquerque, New Mexico, April
  2025. Association for Computational Linguistics.
\newblock ISBN 979-8-89176-189-6.
\newblock \doi{10.18653/v1/2025.naacl-long.560}.
\newblock URL \url{https://aclanthology.org/2025.naacl-long.560/}.

\end{thebibliography}
\end{document}